\pdfoutput=1
\documentclass{article} %
\usepackage{iclr2027_conference,times}
\usepackage[T1]{fontenc}

\usepackage{amsmath,amsfonts,bm}

\def\eqref#1{equation~\ref{#1}}
\def\1{\bm{1}}

\def\ve{{\bm{e}}}

\def\vn{{\bm{n}}}

\def\vu{{\bm{u}}}

\def\vx{{\bm{x}}}

\def\mI{{\bm{I}}}

\DeclareMathAlphabet{\mathsfit}{\encodingdefault}{\sfdefault}{m}{sl}
\SetMathAlphabet{\mathsfit}{bold}{\encodingdefault}{\sfdefault}{bx}{n}

\newcommand{\R}{\mathbb{R}}

\usepackage{hyperref}
\usepackage{url}
\usepackage{graphicx}
\usepackage{adjustbox}
\usepackage{amsmath}
\usepackage{booktabs}
\usepackage{multirow}
\usepackage{tabularx}
\usepackage{array}
\usepackage{xcolor}
\usepackage{float}
\usepackage{placeins}

\newcolumntype{Y}{>{\raggedright\arraybackslash}X}
\newcolumntype{L}[1]{>{\raggedright\arraybackslash}p{#1}}
\newcommand{\campaignrule}{\specialrule{0.18pt}{2pt}{2pt}}

\graphicspath{{figs/}}

\title{PoreML: A Data-Driven Framework for Learning Multiphase Flow in Porous Media}

\author{Chunyang Wang$^{1}$,\; Mingrui Zhang$^{1}$,\; Yuyan Zhang$^{1}$,\; Linqi Zhu$^{1}$,\; Xin Ju$^{2,3}$, \\
\textbf{Edo Sicco Boek$^{4}$,\; Martin J. Blunt$^{1}$,\; Gege Wen$^{1,3}$} \\[0.6ex]
\normalfont $^{1}$Imperial College London \quad $^{2}$Stanford University \quad $^{3}$EarthFlow AI, Inc. \\
\normalfont $^{4}$Queen Mary University of London
}

\iclrfinalcopy
\begin{document}

\maketitle
\lhead{Preprint.}

\begin{abstract} 
Multiphase flow in porous microstructures is central to CO$_2$ storage, fuel-cell operation, and flip-chip packaging. Predicting these flows remains challenging because wettability and complex pore geometry govern the nonlinear evolution of fluid interfaces. Machine learning holds substantial promise for advancing the field, but progress is constrained by scarce time-resolved 3D datasets and a lack of unified workflow for training and evaluating models. To fill this critical gap, we introduce \textbf{PoreML}, an open-source framework unifying data generation, model training, and evaluation grounded in pore-scale physics. The framework comprises three core components. \textbf{(a)} A modern GPU-native lattice Boltzmann solver\footnote{\url{https://github.com/PoreML/bob}}, validated against analytical solutions and published experiments, enables reproducible data generation. \textbf{(b)} A \textbf{3.3\,TB dataset}\footnote{\url{https://huggingface.co/datasets/PoreML/PoreML_data}} contains \textbf{560 simulation runs} and \textbf{158{,}546 stored time steps} across four application-driven scenarios. These trajectories span synthetic structures and geometries derived from micro-CT scans of real materials, covering diverse wetting conditions and viscosity ratios. \textbf{(c)} A unified learning framework\footnote{\url{https://github.com/PoreML/poreml}} evaluates one-step prediction and autoregressive rollouts. Its domain-specific evaluation protocols assess predictive accuracy and physical consistency. We evaluate five models of diverse architecture under these protocols. Two complementary challenges assess transfer to larger domains and from synthetic to micro-CT-derived structures. PoreML provides a shared foundation for machine-learning research on multiphase flow in porous media, with the aim of empowering the community to develop reliable predictive models and advance the field.
\end{abstract}

\begin{figure}[H]
  \centering
  \includegraphics[width=0.8\linewidth]{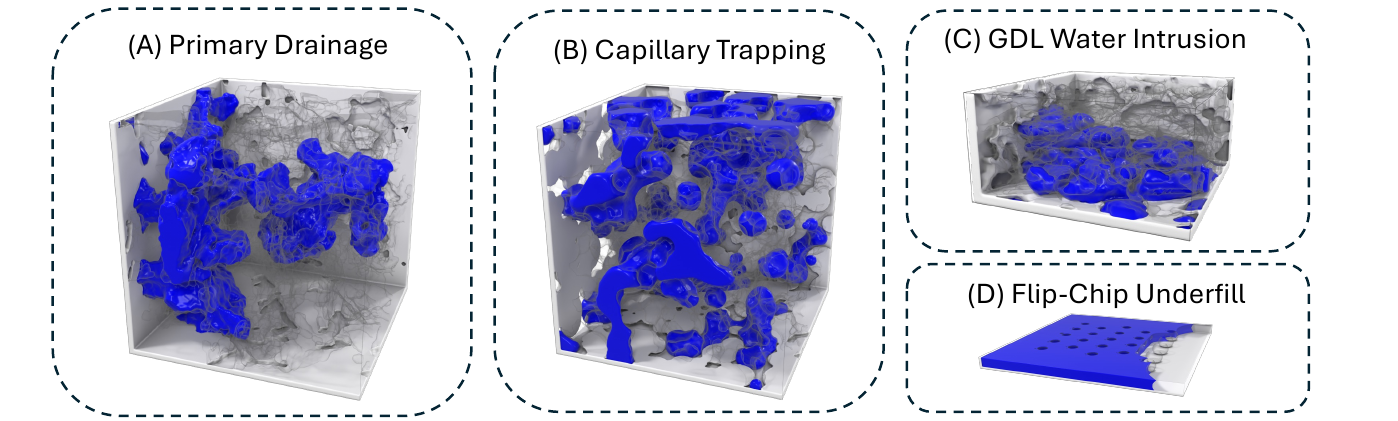}
  \caption{\textbf{PoreML simulation campaigns.} \textbf{(A)}~Primary
  drainage and \textbf{(B)}~capillary trapping for CO$_2$ sequestration and
  geological energy storage, \textbf{(C)}~GDL water intrusion for fuel-cell
  design, and \textbf{(D)}~flip-chip underfill for chip packaging. Only the fluid phase of interest is shown for clarity; translucent gray meshes depict the solid.}
  \label{fig:teaser}
\end{figure}

\section{Introduction}
\label{sec:intro}

\begin{figure}[htbp]
  \centering
  \includegraphics[width=0.9\linewidth]{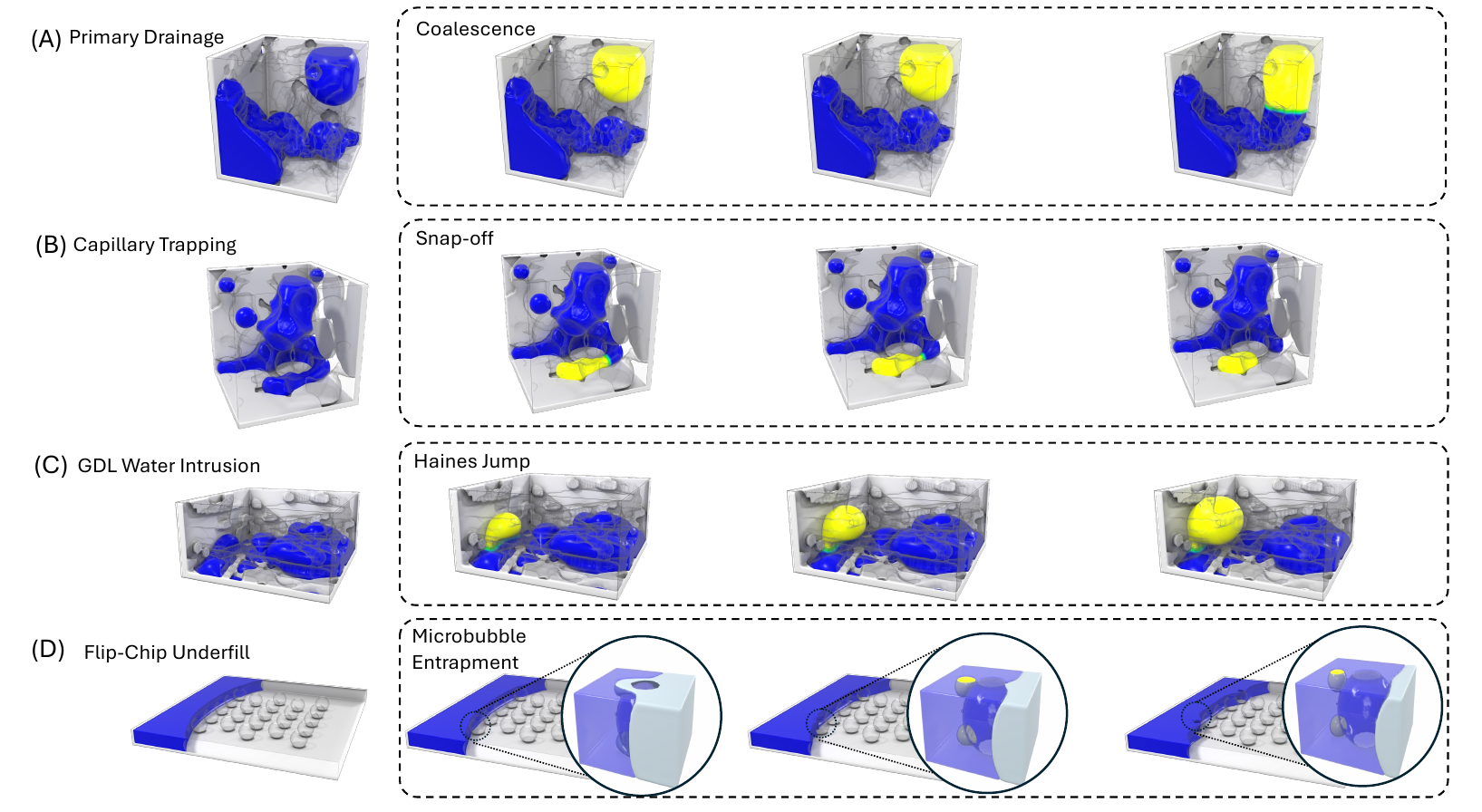}
  \caption{\textbf{Representative pore-scale dynamics.} Localized events are
  highlighted in yellow; these mechanisms are not exclusive to the displayed
  campaign, and events such as snap-off can also occur during drainage.
  \textbf{(A)~Primary drainage}: ganglion coalescence, where interfacial
  tension retracts menisci and connects neighboring non-wetting bodies.
  \textbf{(B)~Capillary trapping}: snap-off, where wetting films invade a
  constricted throat and pinch a non-wetting thread into an isolated ganglion.
  \textbf{(C)~GDL intrusion}: Haines jump, an abrupt meniscus depinning and
  pore-filling event in a fibrous throat network. \textbf{(D)~Underfill}:
  microbubble or void entrapment, where advancing encapsulant fronts around
  solder balls leave trapped air pockets.}
  \label{fig:ganglion}
\end{figure}

Some of the fundamental technologies of the twenty-first century, from
geological CO$_2$ sequestration to efficient fuel cells, rely on porous
materials \citep{farber2025porous}. Their performance is decided inside pores
only micrometers wide, where a small shift of a fluid interface can reroute
the entire flow. At low flow speeds, capillarity and wettability dominate.
An interface pins at a narrow throat, then abruptly floods the adjacent
pores, an event known as a Haines jump
\citep[Figure~\ref{fig:ganglion}C;][]{berg2013realtime,tekseth2024multiscale}.
The same forces break and remake fluid connections: in
\emph{snap-off}, a wetting film swells in a throat until it pinches off the
non-wetting thread and strands an isolated ganglion
\citep[Figure~\ref{fig:ganglion}B;][]{roof1970snapoff,singh2017dynamics}; in
\emph{coalescence}, separated ganglia meet and merge, reopening a pathway
\citep[Figure~\ref{fig:ganglion}A;][]{rucker2015connected,reynolds2017dynamic}.
These pore-scale events shape behavior at far larger scales: how much fluid
is trapped, how easily each phase flows, and why displacement is hysteretic
\citep{armstrong2016beyond,schluter2016porescale,bultreys2024microvelocimetry}.
Yet they can occur at fixed saturation, hidden from bulk descriptors
\citep{armstrong2016beyond}, and how they add up to continuum-scale flow is
still not fully understood \citep{singh2019capillary,berg2026interface}.
Predicting them is nonetheless essential. For example, in
geological CO$_2$ storage, returning brine fragments connected CO$_2$ into
trapped ganglia, helping prevent buoyant escape
\citep{krevor2015capillary,andrew2014porescale,krevor2023subsurface}.
In polymer-electrolyte fuel cells, product water must leave through the
fibrous gas-diffusion layer (GDL) before it blocks the pathways that supply
reactant gas \citep{weber2014critical,wang2023large,sarkeziselsky2022lattice}.
In flip-chip packaging, capillary forces draw encapsulant through
solder-bump arrays
\citep{ng2022analytical,wang2016underfill,lee2024numerical,wang2026lbm}, and
the cured underfill protects the chip by shielding its solder joints from
thermal stress \citep{zhang2004recent}.
 
\begin{figure}[H]
  \centering
  \includegraphics[width=\linewidth]{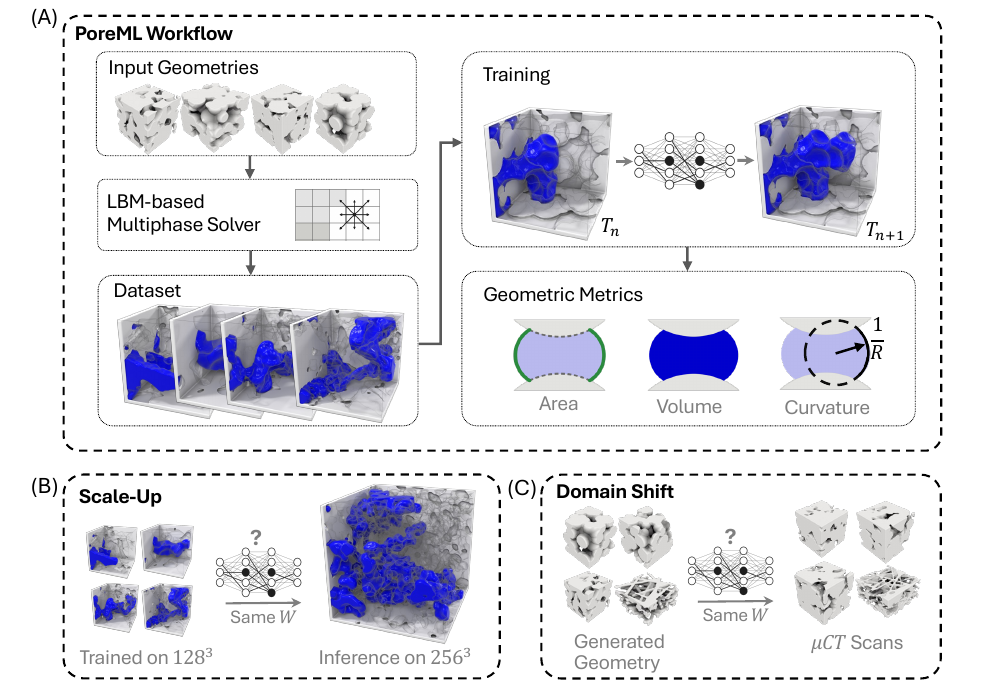}
  \caption{\textbf{PoreML framework.} \textbf{(A)}~A 3D LBM solver turns
  generated and $\mu$CT geometries under controlled flow conditions into
  time-resolved phase, pressure, and velocity fields; models predict the next
  frame and are scored with field errors and geometry-aware metrics.
  \textbf{(B)~Scale-Up}: train on small domains, test on larger ones at the
  same resolution. \textbf{(C)~Geo-Shift}: train on generated geometries,
  test on $\mu$CT scans.}
  \label{fig:workflow}
\end{figure}

Machine learning offers new avenues for understanding and modeling these
dynamics \citep{kochkov2021machine,santos2020poreflow,shodiev2021machine},
but progress is constrained by the scarcity of time-resolved,
three-dimensional experimental and simulation data
\citep{emeliana2026integrating}. Experiments demand carefully
prepared samples, specialized facilities, and expert operators;
core-analysis campaigns can take weeks to months
\citep{mcphee2015coreanalysis,speight2020reservoirs}, while capturing rapid
pore-scale events in 3D can require synchrotron imaging
\citep{berg2013realtime}. Direct simulation on $\mu$CT scans provides a controlled
alternative, but it inherits the quality of its input mesh: the pore space must be
imaged finely and segmented with care, since both change the computed flow
\citep{saxena2018imaging,garfi2020sensitivity,carrillo2022impact}. It is
also bound by resolution: the narrowest throats must span several voxels
before the simulated interface behaves physically
\citep{li2020discretization}, and at that resolution the cost climbs steeply
as three-dimensional domains grow \citep{armstrong2026lattice}. Statistically
meaningful results need a representative elementary volume (REV), the
smallest sample whose averages no longer depend on its size
\citep{bear1972dynamics}, which
for multiphase flow far exceeds that for porosity
\citep{georgiadis2013porescale,jackson2020representative}; simulations at
that scale have relied on leadership-class supercomputers
\citep{wang2023large}, beyond the reach of most groups.
Consequently, dense ground truth is most accessible on small synthetic
samples, whereas practical predictions are needed on representative domains
and real materials \citep{blunt2013porescale,berg2026interface}. This
mismatch poses two distinct challenges. \emph{Scale}: at fixed resolution, a
domain $n$ voxels wide costs $\mathcal{O}(n^3)$ per time step, so
large-domain ground truth stays scarce while larger samples bring more pores
and longer connectivity paths. \emph{Geometry}: moving from synthetic
structures to scanned materials means generalizing to a different
distribution of pore geometries. The central challenge is to turn data we can
afford to generate into reliable predictions at the scales and in the
materials where they matter, and to judge them with metrics grounded in the
physics rather than in voxel error alone.

We introduce \textbf{PoreML}, a simulation-to-learning framework for 3D
multiphase flow in porous media (Figure~\ref{fig:workflow}). We first develop
a GPU-native, JAX-based color-gradient lattice Boltzmann method (LBM) solver
with geometric wetting \citep{leclaire2017generalized,akai2018wetting},
validated against analytical solutions and published pore-scale experiments.
Using this engine, we invest \textbf{over 3,000 GPU-hours} to generate
\textbf{560 time-resolved 3D simulations} with \textbf{158{,}546 time
snapshots} across four applications: rock drainage, capillary trapping, GDL water
intrusion, and flip-chip underfill. The dataset
pairs generated structures with real geometries obtained from micro-computed
tomography ($\mu$CT) scans, and spans multiple domain sizes, wettabilities,
and viscosity contrasts. Two challenges test application-relevant
generalization: \textbf{Geo-Shift} transfers models from generated structures
to real rock and carbon-paper geometries from $\mu$CT scans, while
\textbf{Scale-Up} transfers them to domains with doubled spatial dimensions
at fixed resolution.

Our contributions are: \textbf{(1) An open-source solver}, validated on an
eight-case suite of analytical solutions and published experiments
(Appendix~\ref{app:validation}) and sharded across GPUs for large domains,
with which the community can generate data for new geometries, physical
conditions, and applications. \textbf{(2) An open dataset}, to our knowledge
the first, and the largest, to provide time-resolved, full 3D fluid
distributions for multiphase flow in porous media, spanning generated and
$\mu$CT-scanned geometries, four applications, two domain sizes, and a grid
of wettabilities and viscosity ratios. \textbf{(3) An integrated learning
benchmark}: standardized one-step training and rollout evaluation of five
architectures, with metrics assessing field accuracy, geometry, and topology.
~It also includes,
to our knowledge, the first joint evaluation of geometry transfer and
domain-size extrapolation for learned 3D multiphase porous-media dynamics.
By making all three publicly available through an extensible, unified framework, we aim to help the community develop and compare reliable machine-learning models for multiphase flow in porous media and advance the field..

\section{Related Work}

\paragraph{Learning-based models and data for pore-scale flow.}
Existing models and public datasets offer limited coverage of three-dimensional multiphase dynamics at the pore scale. Learned 3D
pore-scale models predict steady single-phase flow \citep{santos2020poreflow},
while evolving rock-fluid surrogates remain two-dimensional
\citep{pinheiro2026surrogate}. Public 3D resources provide single-phase
transport \citep{santos2022dataset}, static wettability states
\citep{gorenkov2026porescale}, or scalar time-series summaries rather than
full fluid configurations \citep{armstrong2026lattice}. General PDE
collections contain no multiphase flow through porous microstructure
\citep{takamoto2022pdebench, ohana2024well}, while BubbleML contains
multiphase dynamics without a confining porous solid
\citep{hassan2023bubbleml}. This leaves a need for time-resolved 3D multiphase data spanning diverse
microgeometries, wettabilities, and viscosity ratios for training and
evaluating machine-learning models.

\label{sec:related}
\paragraph{Lattice Boltzmann solvers for porous media.}
LBM is widely used for direct simulation of multiphase flow in porous
media, combining efficient computation with the ability to handle complex
pore geometries directly from segmented $\mu$CT voxel data
\citep{pan2004lattice,ahrenholz2008prediction}, and the same strengths
have made it a common choice for capillary underfill flow in flip-chip
packaging \citep{wang2016underfill,lee2024numerical,wang2026lbm}. Among its
multiphase formulations, the color-gradient lineage
\citep{gunstensen1991lattice} is especially well suited to narrow throats:
unlike diffusive pseudopotential interfaces under viscosity contrast, it
sustains thin interfaces \citep{yang2013comparison}, although accuracy still
deteriorates when a throat spans only a few voxels
\citep{li2020discretization}. Wetting models likewise progressed from
indirect wall coloring \citep{latvakokko2005static} to geometric
contact-angle conditions \citep{leclaire2017generalized, akai2018wetting}; a
recent comparison separates their strengths, favoring Akai for static angles
and Leclaire for dynamic Washburn flow \citep{sedahmed2024}. Palabos and LBPM
brought this physics to production simulation
\citep{leclaire2017generalized, mcclure2021lbpm}, and JAX-LaB recently brought
a pseudopotential model to JAX \citep{pradhan2026jaxlab}. Our data engine
brings the color-gradient model with geometric wetting into the learning loop
through an accelerator-native implementation
(Section~\ref{sec:dataset:solver}).

\section{PoreML: Learning Multiphase Flow in Porous Media}

\subsection{A JAX-Native Color-Gradient LBM Solver}
\label{sec:dataset:solver}

We implement a modern, GPU-native lattice Boltzmann solver for immiscible
two-phase flow in 3D porous media and use it to generate every PoreML
trajectory. Its distinctive contribution is to unite thin color-gradient interfaces
under viscosity contrast \citep{yang2013comparison}, direct contact-angle
control on complex solids \citep{akai2018wetting}, and multi-GPU JAX
execution in a well-validated, open, extensible engine for generating 3D
pore-scale training data.
It supports common
boundary conditions, including Zou--He pressure boundaries
\citep{zou1997pressure} and velocity and piston inlets, and is validated
against analytical solutions and published experiments
(Appendix~\ref{app:validation}); Appendices~\ref{app:model}
and~\ref{app:solver} detail the model and its implementation. We hope this
open-source solver will help the community generate far more diverse data
and foster reproducible, collaborative research.

\subsection{Dataset Overview}
\label{sec:dataset}

\begin{figure}[t]
  \centering
  \includegraphics[width=\linewidth]{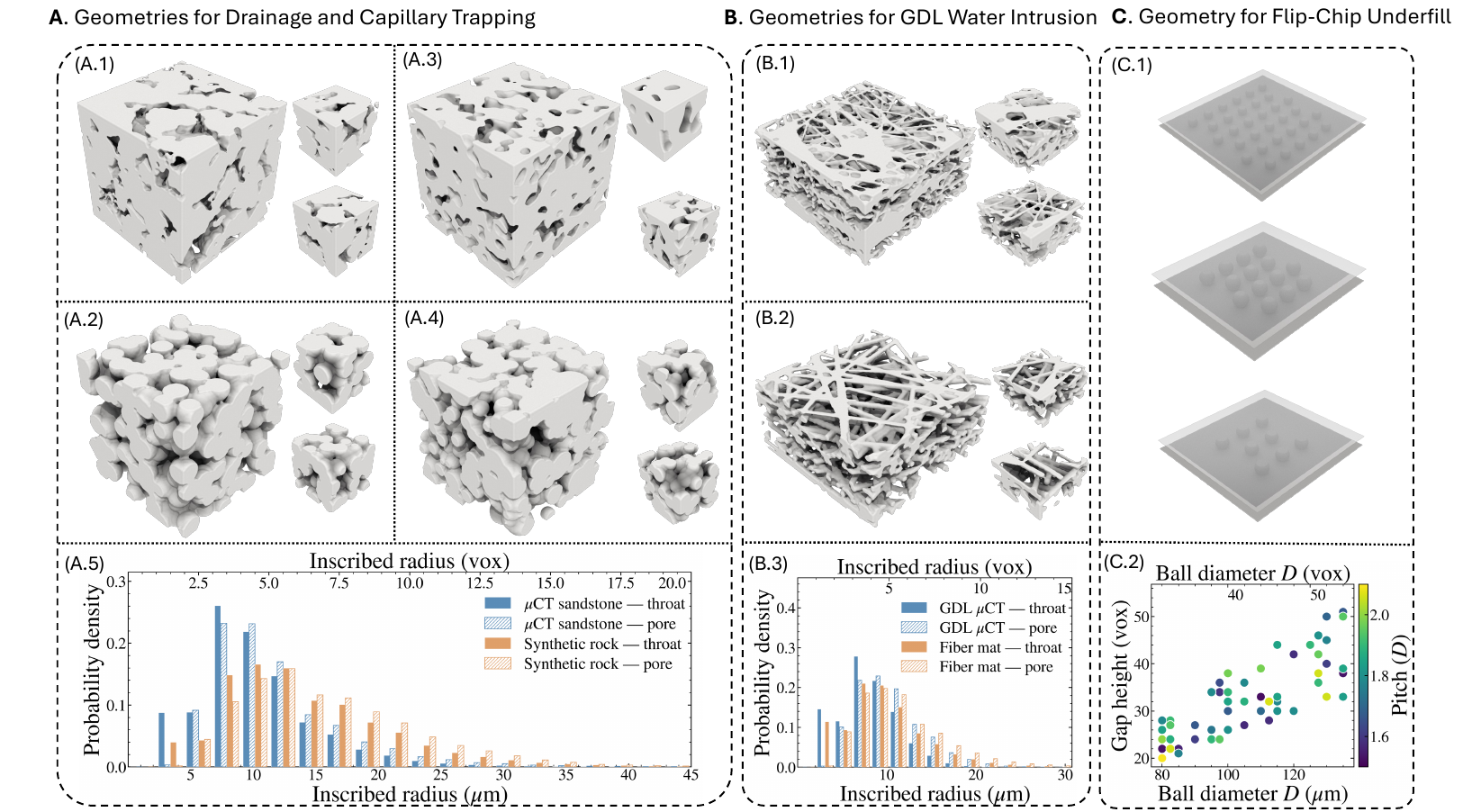}
  \caption{\textbf{Geometry families used by PoreML.}
  \textbf{(A)~Rock-like media}, each shown as one large sample ($256^3$
  voxels) beside two small ones ($128^3$): \textbf{(A.1)}~sandstone from a
  $\mu$CT scan and generated \textbf{(A.2)}~sphere packs,
  \textbf{(A.3)}~blob media, and \textbf{(A.4)}~polydisperse spheres, with
  \textbf{(A.5)}~pore and throat radii (maximal inscribed spheres).
  \textbf{(B)~Gas-diffusion layers} at $256{\times}256{\times}128$ and
  $128{\times}128{\times}64$ voxels: \textbf{(B.1)}~$\mu$CT scan,
  \textbf{(B.2)}~generated fiber mat, and \textbf{(B.3)}~pore and throat
  radii. \textbf{(C)~Underfill}: flip-chip gaps set by solder-ball diameter,
  gap height, and pitch. More samples: Appendix~\ref{app:geo}.}
  \label{fig:geo}
\end{figure}

\textbf{Geometries.} PoreML pairs real microstructures with generated ones,
\label{sec:dataset:geometry}and provides both at two domain sizes (Figure~\ref{fig:geo}). The real
samples are crops of segmented $\mu$CT
scans: Bentheimer, CastleGate, and Buff Berea sandstones
\citep{neumann2021high}, and carbon-paper gas-diffusion layers (GDLs) coated
with 5\%, 20\%, and 40\% PTFE \citep{shojaei2022minimal}. The
generated families overlap them in local pore size but have controlled
statistics: sphere packs (well sorted, wide throats), Gaussian-random-field
blobs (irregular, tight), polydisperse overlapping grains (poorly sorted
sandstone analogs), and in-plane fiber mats. Underfill uses parametric
flip-chip gaps, the thin space between a chip and its substrate that an array
of solder balls bridges; three numbers describe one: ball diameter, pitch
(center-to-center ball spacing), and gap height (the chip-to-substrate
standoff). The two sizes of the rock and GDL families share one resolution,
so a larger domain holds more pores, not finer ones.

\textbf{Simulations.} \label{sec:dataset:geometry}
We build the dataset by simulating four practical flow scenarios in these
geometries (Table~\ref{tab:campaigns}). In \textbf{drainage}, a
non-wetting phase invades a rock saturated with the wetting phase, as in
hydrocarbon recovery \citep{blunt2017multiphase} and the underground
CO$_2$ and hydrogen storage behind net-zero pathways
\citep{krevor2023subsurface}; in \textbf{trapping}, the wetting phase returns and
pinches the non-wetting phase into disconnected ganglia that can no longer
move, keeping injected CO$_2$ in place
\citep{krevor2015capillary}. \textbf{GDL intrusion} pushes liquid water
through a hydrophobic fiber layer, the path product water takes out of a
hydrogen fuel cell before it blocks the gas supply and cuts clean power
\citep{weber2014critical,wang2023large}. \textbf{Underfill} has no pump:
capillary suction draws encapsulant under a chip to shield its solder joints
from thermal stress \citep{zhang2004recent}, and fronts that pass the solder
balls at different speeds close around air pockets, the voids that threaten
package reliability \citep{ng2022analytical}. Each campaign
varies the contact angle $\theta$, which says which fluid the solid prefers
\citep{aljaberi2026spatially}, and the viscosity ratio $M$; together they
decide whether a meniscus pins, jumps, or is bypassed. The three porous-media
campaigns run at capillary number $\mathrm{Ca}=10^{-5}$ (the dimensionless
ratio of viscous to capillary forces; \citealp{lenormand1988numerical}),
where capillarity dominates. Appendix~\ref{app:setup} gives the domains,
boundary conditions, parameters, and termination conditions. Every stored
time step records the 3D phase distribution, velocity components, and
pressure, and each run keeps its control and simulation parameters as
metadata.

\begin{table}[t]
  \caption{\textbf{PoreML dataset summary and key parameter description.}
  $M=\mu_{\mathrm{nw}}/\mu_{\mathrm{w}}$ for drainage, trapping, and GDL;
  underfill instead follows the application convention
  $M=\mu_{\mathrm{encapsulant}}/\mu_{\mathrm{air}}$. Contact angle $\theta$ is
  measured through the non-wetting phase for the porous-media campaigns and
  through the wetting encapsulant for underfill.}
  \label{tab:campaigns}
  \centering
  \scriptsize
  \setlength{\tabcolsep}{4.5pt}
  \renewcommand{\arraystretch}{1.15}

  \begin{tabularx}{\textwidth}{@{}
    L{0.115\textwidth}
    L{0.095\textwidth}
    L{0.115\textwidth}
    L{0.10\textwidth}
    L{0.075\textwidth}
    L{0.165\textwidth}
    Y@{}}
    \toprule
    \parbox[t]{\linewidth}{\raggedright Dataset\\name} &
    \parbox[t]{\linewidth}{\raggedright Cases} &
    \parbox[t]{\linewidth}{\raggedright Domain size\\(voxels)} &
    \parbox[t]{\linewidth}{\raggedright Contact angle\\$\theta$ ($^\circ$)} &
    \parbox[t]{\linewidth}{\raggedright Viscosity\\ratio $M$} &
    \parbox[t]{\linewidth}{\raggedright Domain\\geometry} &
    \parbox[t]{\linewidth}{\raggedright Simulation\\protocol} \\
    \midrule

    \rule{0pt}{5ex}\texttt{drainage\_128} &
    \shortstack[l]{80 generated;\\80 $\mu$CT} &
    $128^3$ &
    $\left\{\substack{120,130,\\140,150}\right\}$ &
    \multirow{2}{=}{\raggedright
      $\left\{\substack{0.05,0.1,\\0.2,1}\right\}$} &
    \multirow{2}{=}{\raggedright
      Generated blobs, polydisperse spheres, and sphere packs; $\mu$CT sandstones} &
    \multirow{2}{=}{\raggedright
      Piston-driven non-wetting injection at $\mathrm{Ca}=10^{-5}$;
      water-wet porous outlet plate} \\
    \cmidrule(lr){1-4}
    \rule{0pt}{5ex}\texttt{drainage\_256} &
    \shortstack[l]{8 generated;\\8 $\mu$CT} &
    $256^3$ &
    $\{120,150\}$ & & & \\

    \campaignrule

    \rule{0pt}{5ex}\texttt{trapping\_128} &
    \shortstack[l]{80 generated;\\80 $\mu$CT} &
    $128^3$ &
    $\left\{\substack{120,130,\\140,150}\right\}$ &
    \multirow{2}{=}{\raggedright
      $\left\{\substack{0.05,0.1,\\0.2,1}\right\}$} &
    \multirow{2}{=}{\raggedright
      Generated and $\mu$CT rock-like media} &
    \multirow{2}{=}{\raggedright
      Piston-off equilibration, then wetting-water flood at
      $\mathrm{Ca}=10^{-5}$; open, pressure-anchored outlet} \\
    \cmidrule(lr){1-4}
    \rule{0pt}{5ex}\texttt{trapping\_256} &
    \shortstack[l]{8 generated;\\8 $\mu$CT} &
    $256^3$ &
    $\{120,150\}$ & & & \\

    \campaignrule

    \rule{0pt}{5ex}\texttt{gdl\_128} &
    \shortstack[l]{80 generated;\\80 $\mu$CT} &
    $128{\times}128{\times}64$ &
    $\left\{\substack{120,130,\\140,150}\right\}$ &
    \multirow{2}{=}{\raggedright
      $\left\{\substack{1,5,\\10,20}\right\}$} &
    \multirow{2}{=}{\raggedright
      Generated fiber mats; $\mu$CT carbon paper at
      5\%, 20\%, and 40\% PTFE loading} &
    \multirow{2}{=}{\raggedright
      Piston-driven through-plane water injection at
      $\mathrm{Ca}=10^{-5}$; open outlet; no plate} \\
    \cmidrule(lr){1-4}
    \rule{0pt}{5ex}\texttt{gdl\_256} &
    \shortstack[l]{8 generated;\\8 $\mu$CT} &
    $256{\times}256{\times}128$ &
    $\{120,150\}$ & & & \\

    \campaignrule

    \texttt{underfill} &
    32 flip-chip &
    \shortstack[l]{$(22\text{--}48){\times}482$\\${\times}(432\text{--}476)$} &
    $\{30,40,50\}$ &
    $\left\{\substack{5,10,\\20,30}\right\}$ &
    Patterned solder-ball gaps &
    Spontaneous imbibition between
    equal-pressure inlet and outlet \\

    \bottomrule
  \end{tabularx}
\end{table}

\subsection{Machine Learning Framework}
\label{sec:dataset:framework}

\textbf{Notation and task.} Let $\Omega\subset\mathbb{R}^3$ denote a physical
sample and $\Omega_h$ its uniform voxelization. The static field
$g:\Omega_h\rightarrow\{0,1\}$ is both the solid mask and the geometry input:
$g(\mathbf r)=1$ marks solid and $g(\mathbf r)=0$ the pore space. The geometry
families of Section~\ref{sec:dataset:geometry} define two distributions,
generated structures $g\sim P_{\mathrm{gen}}$ and segmented $\mu$CT scans
$g\sim P_{\mu\mathrm{CT}}$. At saved time $t$ the state is
\begin{equation}
\label{eq:state}
\mathbf{x}_t(\mathbf{r}) =
\bigl(\phi_t(\mathbf{r}),\,\mathbf{u}_t(\mathbf{r}),\,p_t(\mathbf{r})\bigr)
\in\mathbb{R}^{5},\qquad \mathbf{r}\in\Omega_h,
\end{equation}
where $\phi$ identifies the two phases, $\mathbf{u}$ is velocity, and $p$ is
pressure. A run is further specified by its physical conditions
$\boldsymbol{\lambda}=(\mathrm{Ca},\theta,M)$ and by the boundary protocol
$\mathcal{B}$ of its campaign on $\partial\Omega_h$. The simulator advances
the state through a discrete evolution operator,
\begin{equation}
\label{eq:evolution}
\mathcal{G}_{\Delta t}:
(\mathbf{x}_{t-h+1:t},g,\boldsymbol{\lambda},\mathcal{B})
\mapsto \mathbf{x}_{t+\Delta t},
\end{equation}
where $h$ is the history length and $\Delta t$ the interval between saved
frames. The learning task is to approximate this transition with a network
$\widehat{\mathcal{G}}_{W}$ that is conditioned explicitly on geometry,
physical parameters, and boundary protocol; $\widehat{\mathbf{x}}$ denotes
its prediction.

\textbf{Geometry-aware metrics.}
\label{sec:dataset:metrics}
We adopt metrics grounded in pore-scale physics, the four Minkowski
functionals \citep{armstrong2019minkowski}, which carry richer diagnostic
information than MAE or other field errors: they report not how many voxels
are wrong, but what is wrong with the fluid configuration. In
capillary-dominated flow the shape of the interface is part of the physical
state. At local equilibrium the Young--Laplace equation,
$p_c=p_{\mathrm{nw}}-p_{\mathrm{w}}=2\sigma H$, ties the capillary pressure
$p_c$ to the interfacial tension $\sigma$ and the mean curvature
$H=(\kappa_1+\kappa_2)/2$, where $\kappa_1$ and $\kappa_2$ are the principal
curvatures of the interface. A prediction with a small $\phi$ error can
therefore still carry the wrong curvature, and hence the wrong local
capillary pressure, misplace interfacial area, or miss a snap-off that
disconnects a ganglion while changing only a few voxels. For a phase body
$X_t$ with smooth boundary $\partial X_t$, the functionals are:

\par\noindent
\makebox[\linewidth][s]{%
\begin{minipage}[c]{0.38\linewidth}
\begin{equation}
\label{eq:minkowski-m0}
M_0(X_t)=\int_{X_t} dV,
\end{equation}
\end{minipage}\hfill
\begin{minipage}[c]{0.60\linewidth}
\begin{equation}
\label{eq:minkowski-m1}
M_1(X_t)=\int_{\partial X_t} dS,
\end{equation}
\end{minipage}%
}\par
\noindent
\makebox[\linewidth][s]{%
\begin{minipage}[c]{0.38\linewidth}
\begin{equation}
\label{eq:minkowski-m2}
M_2(X_t)=\int_{\partial X'_t} H\,dS,
\end{equation}
\end{minipage}\hfill
\begin{minipage}[c]{0.60\linewidth}
\begin{equation}
\label{eq:minkowski-m3}
M_3(X_t)=\int_{\partial X_t}\kappa_1\kappa_2\,dS
=4\pi\chi(X_t).
\end{equation}
\end{minipage}%
}\par

Here $\chi=N-L+O$ is the Euler characteristic, which counts connected
components $N$, loops $L$, and cavities $O$. $M_0$ is the phase volume
(saturation once divided by pore volume); $M_1$ the boundary area; $M_2$ the
mean curvature integrated over the fluid--fluid interface $\partial X'_t$,
whose local value sets the capillary pressure through the Young--Laplace
equation; and $M_3$ the topology, which snap-off and ganglion coalescence
change while barely moving voxel errors \citep{armstrong2016beyond}. We
report the MAE of $p$ and $\phi$ over pore voxels, the relative error
$|M_k(\widehat{X}_t)-M_k(X_t)|/|M_k(X_t)|$ of $M_0$--$M_2$, and for $M_3$
the MAE of the Euler characteristic $\chi$ per unit volume of the injected
phase, which keeps it comparable across domain sizes
(Appendix~\ref{app:metrics:errors}).

\section{Experiments and Results}
\label{sec:experiments}

The evaluation framework asks two questions: whether a 64-step rollout
keeps the interface geometry and topology that govern capillary dynamics
(Section~\ref{sec:exp:rollout}), and whether that fidelity survives the two
transfers applications demand, to scanned geometries and to larger domains,
on the three campaigns with paired geometry sources and domain sizes
(Section~\ref{sec:exp:shift}). Five architectures serve as references on
held-out geometries; their scores show where current models stand and
which limitations they share.

\suppressfloats[t]
\begin{table}[ht]
  \caption{\textbf{Field and geometric fidelity.}
  In-distribution rows report validation; Scale-Up and Geo-Shift rows
  report transfer tests. $p,\phi$: normalized MAE; $M_0$--$M_2$: relative errors as
  fractions; $M_3$: MAE of the Euler characteristic per $10^5$ voxels of the
  injected-phase volume $V$. Underfill descriptors score the wetting phase,
  others the non-wetting phase. Means of per-run means; lower is better;
  bold: lowest per column within a block.}
  \label{tab:all-results}
  \centering
  \scriptsize
  \setlength{\tabcolsep}{1.5pt}
  \renewcommand{\arraystretch}{1.02}
  \begin{adjustbox}{max width=\textwidth}
  \begin{tabular}{@{}llcccccccccccc@{}}
    \toprule
    & & \multicolumn{6}{c}{One-step prediction}
      & \multicolumn{6}{c}{Autoregressive rollout, step 64} \\
    \cmidrule(lr){3-8}\cmidrule(lr){9-14}
    Campaign & Model & $p$ & $\phi$ & $M_0$ & $M_1$ & $M_2$ & $M_3$
      & $p$ & $\phi$ & $M_0$ & $M_1$ & $M_2$ & $M_3$ \\
    & & \multicolumn{2}{c}{MAE} & \multicolumn{3}{c}{Relative error} & MAE/$V$
      & \multicolumn{2}{c}{MAE} & \multicolumn{3}{c}{Relative error} & MAE/$V$ \\
    \midrule
    \multirow{5}{*}{\shortstack[l]{Drainage\\(In-distribution)}} & UNet3D & 0.00571 & \textbf{0.00272} & \textbf{0.00584} & \textbf{0.00459} & 0.0342 & \textbf{0.0910} & 0.190 & 0.215 & 0.0579 & 0.146 & 0.554 & 4.74 \\
     & FNO3D & 0.00816 & 0.00524 & 0.0231 & 0.0231 & \textbf{0.0296} & 0.140 & 0.192 & 0.256 & 0.0860 & 0.153 & 0.447 & \textbf{1.84} \\
     & P3D & \textbf{0.00568} & 0.00337 & 0.00683 & 0.00506 & 0.0308 & 0.109 & \textbf{0.168} & 0.174 & \textbf{0.0261} & \textbf{0.0346} & \textbf{0.245} & 2.63 \\
     & Transolver++ & 0.00647 & 0.00522 & 0.0146 & 0.0114 & 0.0548 & 0.392 & 0.217 & 0.297 & 0.114 & 1.25 & 3.87 & 109 \\
     & AB-UPT & 0.00665 & 0.00391 & 0.0358 & 0.0203 & 0.158 & 0.455 & 0.198 & \textbf{0.126} & 0.0305 & 0.0598 & 0.368 & 2.23 \\
    \campaignrule
    \multirow{5}{*}{\shortstack[l]{Drainage\\(Scale-Up)}} & UNet3D & 0.00890 & \textbf{0.00317} & 0.0832 & 0.0427 & 0.207 & 0.415 & 0.205 & 0.220 & 0.296 & 0.429 & 0.433 & 2.00 \\
     & FNO3D & 0.00809 & 0.00588 & 0.112 & 0.111 & \textbf{0.102} & 0.516 & \textbf{0.171} & 0.212 & 0.139 & \textbf{0.161} & 1.39 & \textbf{1.14} \\
     & P3D & 0.00730 & 0.00401 & 0.113 & 0.0633 & 0.267 & \textbf{0.382} & 0.191 & 0.209 & 0.243 & 0.316 & \textbf{0.306} & 1.24 \\
     & Transolver++ & 0.00803 & 0.00658 & \textbf{0.0747} & 0.0563 & 0.231 & 1.36 & 0.203 & 0.304 & 0.355 & 2.07 & 5.22 & 130 \\
     & AB-UPT & \textbf{0.00691} & 0.00434 & 0.0872 & \textbf{0.0224} & 0.583 & 0.405 & 0.235 & \textbf{0.155} & \textbf{0.135} & 0.254 & 1.88 & 3.39 \\
    \campaignrule
    \multirow{5}{*}{\shortstack[l]{Drainage\\(Geo-Shift)}} & UNet3D & 0.00896 & 0.00430 & 0.0914 & \textbf{0.0469} & 0.280 & 1.15 & 0.182 & 0.176 & \textbf{0.0302} & 0.0637 & 0.294 & \textbf{1.59} \\
     & FNO3D & 0.0117 & 0.00931 & 0.215 & 0.216 & \textbf{0.0490} & 1.95 & 0.208 & 0.265 & 0.252 & 0.355 & 0.459 & 7.40 \\
     & P3D & 0.00823 & 0.00487 & \textbf{0.0789} & 0.0510 & 0.465 & \textbf{1.12} & 0.211 & 0.199 & 0.0490 & \textbf{0.0540} & \textbf{0.255} & 2.89 \\
     & Transolver++ & 0.00870 & 0.00785 & 0.0794 & 0.0675 & 1.17 & 3.18 & 0.271 & 0.326 & 0.124 & 1.88 & 3.92 & 149 \\
     & AB-UPT & \textbf{0.00744} & \textbf{0.00380} & 0.104 & 0.0651 & 0.802 & 1.21 & \textbf{0.164} & \textbf{0.123} & 0.0342 & 0.0605 & 0.502 & 2.49 \\
    \campaignrule
    \multirow{5}{*}{\shortstack[l]{GDL\\(In-distribution)}} & UNet3D & \textbf{0.00549} & \textbf{0.00515} & 0.0141 & 0.0120 & 0.0974 & 0.735 & 0.131 & 0.197 & 0.0835 & 0.109 & \textbf{0.447} & \textbf{6.02} \\
     & FNO3D & 0.00672 & 0.00942 & 0.0354 & 0.0355 & \textbf{0.0371} & 0.836 & 0.184 & 0.289 & 0.467 & 0.515 & 0.668 & 9.75 \\
     & P3D & 0.00620 & 0.00623 & \textbf{0.00760} & \textbf{0.00711} & 0.0969 & \textbf{0.556} & 0.170 & 0.229 & 0.172 & \textbf{0.0933} & 0.654 & 29.8 \\
     & Transolver++ & 0.00624 & 0.00618 & 0.0376 & 0.0367 & 0.0460 & 0.790 & 0.160 & 0.229 & 0.150 & 0.452 & 2.59 & 73.9 \\
     & AB-UPT & 0.00945 & 0.0119 & 0.0417 & 0.0349 & 1.22 & 2.96 & \textbf{0.116} & \textbf{0.179} & \textbf{0.0808} & 0.268 & 1.41 & 29.4 \\
    \campaignrule
    \multirow{5}{*}{\shortstack[l]{GDL\\(Scale-Up)}} & UNet3D & 0.00740 & \textbf{0.00764} & 0.171 & 0.0970 & 1.35 & 0.914 & 0.139 & 0.216 & 0.533 & 0.483 & 0.585 & 14.7 \\
     & FNO3D & 0.00609 & 0.00860 & 0.143 & 0.143 & \textbf{0.0224} & 1.35 & 0.162 & 0.221 & 0.466 & 0.561 & \textbf{0.387} & \textbf{12.2} \\
     & P3D & 0.00699 & 0.00852 & 0.183 & 0.0921 & 1.50 & \textbf{0.819} & 0.151 & 0.208 & 0.284 & \textbf{0.334} & 0.625 & 37.4 \\
     & Transolver++ & 0.00626 & 0.00825 & 0.143 & 0.143 & 0.0670 & 1.32 & 0.146 & 0.197 & 0.320 & 0.913 & 4.03 & 96.0 \\
     & AB-UPT & \textbf{0.00567} & 0.00834 & \textbf{0.0562} & \textbf{0.0315} & 3.44 & 0.933 & \textbf{0.112} & \textbf{0.187} & \textbf{0.203} & 0.618 & 2.25 & 27.8 \\
    \campaignrule
    \multirow{5}{*}{\shortstack[l]{GDL\\(Geo-Shift)}} & UNet3D & 0.0145 & \textbf{0.0104} & \textbf{0.0764} & \textbf{0.0523} & 0.906 & 17.2 & 0.184 & 0.313 & 0.405 & 0.442 & \textbf{0.466} & 12.5 \\
     & FNO3D & 0.0135 & 0.0122 & 0.290 & 0.254 & 0.341 & \textbf{13.1} & 0.141 & 0.259 & 0.301 & 0.365 & 0.525 & \textbf{12.1} \\
     & P3D & 0.0115 & 0.0128 & 0.0887 & 0.0832 & 0.436 & 16.6 & 0.175 & 0.221 & 0.119 & \textbf{0.120} & 1.19 & 33.0 \\
     & Transolver++ & 0.0137 & 0.0200 & 0.337 & 0.336 & \textbf{0.0437} & 25.9 & 0.245 & 0.362 & 0.307 & 0.421 & 12.2 & 46.1 \\
     & AB-UPT & \textbf{0.0113} & 0.0117 & 0.150 & 0.127 & 0.799 & 17.1 & \textbf{0.118} & \textbf{0.184} & \textbf{0.0876} & 0.212 & 1.59 & 34.8 \\
    \campaignrule
    \multirow{5}{*}{\shortstack[l]{Trapping\\(In-distribution)}} & UNet3D & 0.00855 & \textbf{0.00463} & 0.00126 & 0.00127 & \textbf{0.00851} & 0.0195 & 0.197 & 0.392 & 0.136 & \textbf{0.0920} & \textbf{0.204} & 15.6 \\
     & FNO3D & 0.00857 & 0.00654 & 0.00248 & 0.00229 & 0.0114 & \textbf{0.000566} & \textbf{0.154} & 0.287 & 0.132 & 0.106 & 0.260 & \textbf{1.27} \\
     & P3D & 0.00892 & 0.00568 & 0.00123 & 0.00129 & 0.00861 & 0.0347 & 0.195 & 0.359 & 0.136 & 0.125 & 0.212 & 2.14 \\
     & Transolver++ & 0.00880 & 0.00622 & 0.00174 & 0.00167 & 0.00888 & \textbf{0.000566} & \textbf{0.154} & 0.285 & 0.165 & 0.256 & 0.648 & 8.34 \\
     & AB-UPT & \textbf{0.00833} & 0.00472 & \textbf{0.000624} & \textbf{0.00102} & 0.00896 & 0.0139 & 0.181 & \textbf{0.284} & \textbf{0.0988} & 0.117 & 0.577 & 4.47 \\
    \campaignrule
    \multirow{5}{*}{\shortstack[l]{Trapping\\(Scale-Up)}} & UNet3D & 0.0117 & \textbf{0.00470} & 0.000776 & 0.000623 & 0.00754 & 0.0111 & 0.199 & 0.310 & 0.0686 & \textbf{0.0515} & 0.276 & 7.12 \\
     & FNO3D & 0.0134 & 0.0103 & 0.00171 & 0.00176 & 0.00880 & 0.00913 & 0.171 & 0.316 & 0.174 & 0.0838 & 0.328 & \textbf{1.05} \\
     & P3D & \textbf{0.0116} & 0.00587 & \textbf{0.000501} & \textbf{0.000459} & \textbf{0.00664} & 0.0565 & 0.222 & 0.308 & 0.0740 & 0.0915 & \textbf{0.188} & 1.65 \\
     & Transolver++ & 0.0122 & 0.00695 & 0.000657 & 0.000661 & 0.00814 & \textbf{0.00662} & \textbf{0.141} & \textbf{0.204} & \textbf{0.0589} & 0.147 & 0.409 & 4.50 \\
     & AB-UPT & \textbf{0.0116} & 0.00736 & 0.000715 & 0.000859 & 0.0139 & 0.0289 & 0.279 & 0.308 & 0.107 & 0.0625 & 0.445 & 3.17 \\
    \campaignrule
    \multirow{5}{*}{\shortstack[l]{Trapping\\(Geo-Shift)}} & UNet3D & 0.0150 & 0.00743 & 0.00173 & 0.00202 & 0.0110 & 0.0412 & 0.204 & 0.331 & \textbf{0.117} & \textbf{0.109} & \textbf{0.234} & 2.98 \\
     & FNO3D & 0.0155 & 0.00865 & 0.00196 & 0.00219 & \textbf{0.00945} & \textbf{0.00} & \textbf{0.160} & 0.348 & 0.148 & 0.142 & 0.283 & \textbf{1.89} \\
     & P3D & 0.0150 & 0.00911 & 0.00144 & 0.00180 & 0.0111 & 0.0611 & 0.231 & 0.429 & 0.164 & 0.148 & 0.276 & 6.25 \\
     & Transolver++ & 0.0143 & 0.00826 & 0.00148 & 0.00179 & 0.0132 & \textbf{0.00} & 0.203 & 0.338 & 0.216 & 0.485 & 1.00 & 15.1 \\
     & AB-UPT & \textbf{0.0119} & \textbf{0.00719} & \textbf{0.000999} & \textbf{0.00123} & 0.0135 & 0.0154 & 0.279 & \textbf{0.323} & 0.121 & 0.329 & 0.797 & 11.1 \\
    \campaignrule
    \multirow{5}{*}{\shortstack[l]{Underfill\\(In-distribution)}} & UNet3D & 0.00452 & \textbf{0.00204} & 0.000410 & 0.000745 & 0.0312 & 0.0194 & 0.0541 & 0.108 & 0.0971 & 0.105 & 1.16 & 1.62 \\
     & FNO3D & 0.0142 & 0.0114 & 0.000435 & 0.00145 & 0.0410 & 0.0364 & 0.0810 & 0.244 & 0.302 & 0.438 & 0.955 & 1.06 \\
     & P3D & 0.00420 & 0.00223 & \textbf{0.000272} & \textbf{0.000460} & \textbf{0.0230} & 0.0159 & 0.0438 & 0.114 & 0.0609 & 0.0722 & \textbf{0.426} & 0.803 \\
     & Transolver++ & \textbf{0.00418} & 0.00436 & 0.000914 & 0.00364 & 0.0760 & 0.143 & 0.114 & 0.174 & 0.116 & 0.455 & 2.26 & 2.60 \\
     & AB-UPT & 0.00676 & 0.00276 & 0.000312 & 0.000587 & 0.0573 & \textbf{0.0143} & \textbf{0.0171} & \textbf{0.0322} & \textbf{0.0171} & \textbf{0.0247} & 0.575 & \textbf{0.761} \\
    \bottomrule
  \end{tabular}
  \end{adjustbox}
\end{table}

\subsection{Experimental Setup}
\label{sec:exp:setup}

\textbf{Evaluation setup.}
All models predict the next saved $(\phi,\mathbf{u},p)$ frame, 5{,}000 LBM
steps ahead (1{,}000 for underfill), from the current frame, the solid mask,
$M$, and $\theta$. Only pore voxels are
scored, excluding reservoirs and boundary devices. We evaluate the voxel
models UNet3D \citep{cicek2016unet}, FNO3D \citep{li2021fourier}, and P3D
\citep{holzschuh2026p3d}, and the point models Transolver++
\citep{luo2025transolver} and AB-UPT \citep{alkin2025abupt}. Each is trained
frame to frame with its own recipe, then fine-tuned
with the pushforward trick
\citep{brandstetter2022message,hassan2023bubbleml}, which grades the model on
a step taken from its own prediction; Appendix~\ref{app:baselines} gives the
architectures and training details.

\textbf{Datasets and splits.}
Drainage, GDL, and trapping (Table~\ref{tab:campaigns}) each use 128
small-domain training runs and 32 validation runs, half generated and half
$\mu$CT; underfill uses 25 and 7. Scale-Up tests on 16 large-domain runs
per campaign, eight per geometry source, with each dimension doubled at fixed
resolution ($128^3\to256^3$; $128{\times}128{\times}64\to
256{\times}256{\times}128$ for GDL). Geo-Shift trains on the 64 generated
runs, validates on the 16 generated ones, and tests on 80 $\mu$CT runs per
campaign at the training size.

\subsection{Single-Step and Rollout Performance}
\label{sec:exp:rollout}

\begin{figure}[t]
  \centering
  \includegraphics[width=0.9\linewidth]{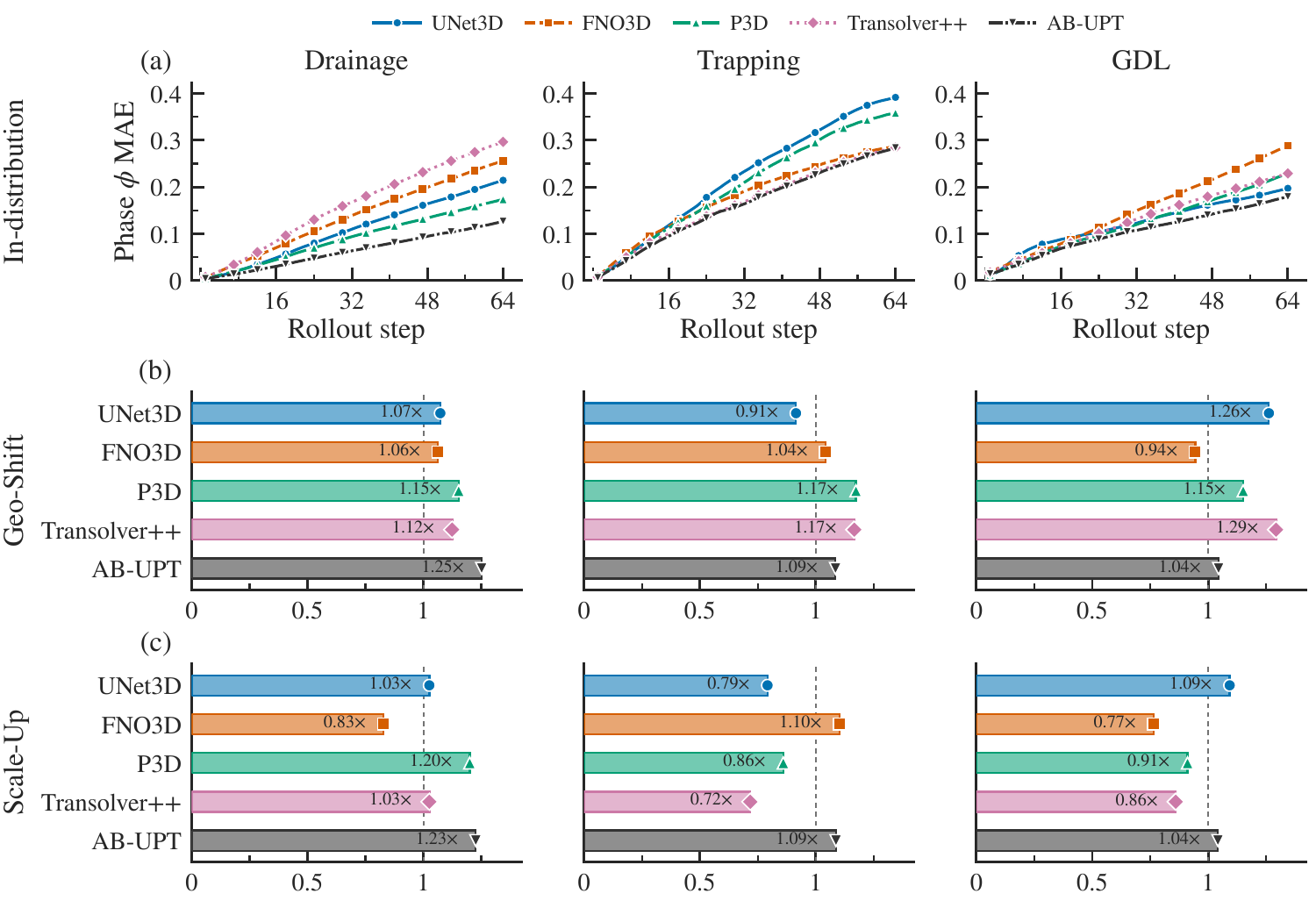}
  \caption{\textbf{In-distribution rollout curve and transfer capability.}
  \textbf{(a)}~In-distribution Phase MAE along the 64-step in-distribution rollout.
  \textbf{(b)} Geo-Shift and \textbf{(c)} Scale-Up: step-64 phase MAE
relative to the same model's in-distribution MAE (generated-only for
Geo-Shift). A ratio of 1.5 indicates 50\% higher error; the dashed
line at 1 indicates no change.}
  \label{fig:result}
\end{figure}

A model is useful only if it stays accurate over multiple steps.
In an autoregressive rollout,
$\widehat{\mathbf{x}}_{t+k\Delta t}=\widehat{\mathcal{G}}_{W}(\widehat{\mathbf{x}}_{t+(k-1)\Delta t},g,\boldsymbol{\lambda},\mathcal{B})$
for $k=1,\dots,64$, each prediction becomes the next input, so errors can
compound. We therefore score two regimes on the held-out runs: the one-step
prediction from a true frame, and step 64 of rollouts started from successive
windows along each trajectory. Each is measured with field errors and
Minkowski functionals, averaging within a run before averaging runs;
Appendix~\ref{app:results:other} gives further metrics.

\textbf{Results.} Over one step every model achieves low error ($\phi$ MAE
0.002--0.012), and the voxel models UNet3D and P3D lead 16 of the 24 columns
(Table~\ref{tab:all-results}). Over 64 steps the strengths spread out:
AB-UPT has the lowest $\phi$ MAE in all four campaigns
(Figure~\ref{fig:result}a), P3D the best interface geometry ($M_0$--$M_2$)
in drainage, and FNO3D and UNet3D the best topology in three campaigns.
Single-step accuracy thus does not by itself imply good rollout
performance. The Minkowski columns capture what the field error does not:
topology is the least well predicted quantity. The median $M_3$ error over
the 20 model--campaign pairs is 3.6 miscounts per $10^5$ voxels, as large as
the reference $\chi$ density itself (1--9), against a median $M_0$ error of
11\%, and
the model with the lowest $\phi$ MAE also has the lowest $M_3$ error in
only one campaign. A low field error does not guarantee the right
connectivity.

\subsection{Distribution Shift: Geo-Shift and Scale-Up}
\label{sec:exp:shift}

A model is useful when trained on affordable ground-truth data and applied to cases where such data are more costly to generate: in scanned materials and on larger
domains (Figure~\ref{fig:workflow}). \textbf{Geo-Shift} trains on generated
geometries, $g\sim P_{\mathrm{gen}}$, and tests on
$g\sim P_{\mu\mathrm{CT}}$, whose pore sizes overlap but whose morphology and
connectivity differ. \textbf{Scale-Up} applies weights trained on the small
domain to the large one at the same resolution: domain-size extrapolation,
not super-resolution, with more pores, longer connectivity paths, and more
frequent interactions between local events and the global flow. Both reuse
the trained weights without retraining under the rollout protocol of
Section~\ref{sec:exp:rollout}, and each result is read against the same
model's in-distribution error, which separates the cost of the shift from the
difficulty of the rollout. 

\textbf{Results.} $\phi$ rollout errors remain bounded across both shifts,
but $M_3$ (Euler characteristic) is consistently the least well predicted metric. AB-UPT achieves the lowest $\phi$ MAE in five
of six blocks (Table~\ref{tab:all-results}). Bulk geometry is more
sensitive: under Scale-Up, phase-volume and interfacial-area errors
increase more than field error. Topological fidelity warrants closer
attention during rollout, particularly for GDL, where a prediction that
engulfs a fiber or bridges two regions with a voxel-thin strand closes a
loop and lowers the Euler characteristic by one, with little effect on
the field error (Appendix~\ref{app:failure}). Transfer to scanned materials and larger
domains is encouraging, but low field and bulk geometric errors do not necessarily imply faithful local geometry or phase topology.
These results motivate future research that focuses on
preserving physically meaningful phase topology over long rollouts.

\FloatBarrier
\section{Conclusion}

\label{sec:conclusion}
\label{sec:conclusion}
PoreML aims to address the scarcity of time-resolved 3D data and unified
learning benchmarks for multiphase flow in porous media. It unifies a
validated open-source solver, datasets across four applications, and
reproducible training and evaluation tools. Its 64-step rollouts and
transfer tests assess field accuracy alongside interface geometry and
topology, measured through phase volume, interfacial area, integrated mean
curvature, and the Euler characteristic. These metrics connect model
evaluation to dynamics that matter for energy storage, fuel cells, and
chip packaging. The baselines provide initial performance references,
while explicit tests of mass and momentum conservation remain future work.
Limitations and future directions are discussed in
Appendix~\ref{app:future}. By enabling new data generation, consistent
model comparison, and extension to new conditions, PoreML provides a
shared foundation for advancing machine learning in porous media.

\subsection*{AI Use Statement}

We used generative AI tools to revise and polish manuscript text for
readability, assist with LaTeX formatting, and support literature retrieval
and discovery, including identifying related work. We also used these tools
to support research execution through software development and scripting.
We did not use generative AI tools to interpret results, formulate claims or develop proofs, or design experiments or research
methods. We have reviewed all AI-assisted work and take responsibility for
the final content of this work, including text, claims, software, and other
artifacts produced with the aid of generative AI.
\subsection*{Reproducibility Statement}

Every resource needed to reproduce this work is public and documented in
Appendix~\ref{app:access}: the solver, with its test suite, the validation
cases of Appendix~\ref{app:validation}, and the campaign drivers that
generated every run; the dataset and the input geometries, every run
carrying its solver configuration, geometry hash, and run conditions; and
the machine learning framework, with the configuration file of every
benchmark run, the frozen splits of the Geo-Shift and Scale-Up challenges,
the trained checkpoints behind every reported number, and a demo notebook
that runs the full workflow on a small subset of the data. Simulation
settings and parameter sampling are given in Appendix~\ref{app:setup},
and the baseline architectures and training recipes in
Appendix~\ref{app:baselines}.

\bibliography{poreml}
\bibliographystyle{iclr2027_conference}

\clearpage
\appendix
\raggedbottom

\section{Other Experiment Results}
\label{app:results}

This appendix collects results that complement the main text.

\subsection{Supplementary Metrics}
\label{app:results:other}

Table~\ref{tab:other-results} reports five further metrics for the same
campaign, study, and model rows as Table~\ref{tab:all-results}; blocks,
averaging, and contributing runs follow that table. Each answers a question
that the main table leaves open.

\textbf{Relative phase error ($\phi$, Rel.\ MAE).} The summed absolute phase
error divided by the summed magnitude of the reference field,
$\sum|\widehat{\phi}-\phi|/\sum|\phi|$ over the scored pore voxels. Because
$|\phi|$ is close to one away from the interface, it expresses the phase MAE
as a fraction of the field, which makes campaigns and domain sizes directly
comparable.

\textbf{Phase overlap (IoU).} Both fields are thresholded into the
non-wetting body, and the volume the two bodies share is divided by the
volume either occupies. IoU asks whether the fluid is in the right place, not
merely present in the right amount: a rollout with the correct saturation but
a ganglion in the wrong pore scores low, and one means agreement voxel for
voxel. It is the score printed on the panels of Appendix~\ref{app:vis}.

\textbf{Velocity error ($\mathbf{u}$).} The mean magnitude of the velocity
error vector, $\lVert\widehat{\mathbf{u}}-\mathbf{u}\rVert_2$ averaged over the
scored pore voxels, in normalized units.

\textbf{Saturation error ($S$).} The absolute difference between predicted
and reference saturation, the fraction of the pore volume that the scored
phase occupies. Saturation is the bulk variable in which continuum
descriptions of multiphase flow are written, and an error of 0.01 means that
the predicted phase volume is off by one percent of the pore volume. It is
the absolute counterpart of the relative $M_0$ error of
Table~\ref{tab:all-results}.

\textbf{Curvature distribution ($H$, $W_1/\sigma$).} Every meniscus has its
own mean curvature, set by the pore it sits in, so an interface carries a
distribution of $H$ rather than one value; $M_2$ keeps only its integral. We
histogram $H$ over the fluid--fluid interface and report the Wasserstein-1
distance between the predicted and reference histograms, divided by the
standard deviation of the reference. $W_1$ is the average amount by which
curvature values must be shifted to turn one distribution into the other, so
a score of one means the predicted curvatures are off by one reference
standard deviation on average, and through the Young--Laplace equation
(Appendix~\ref{app:metrics:curvature}) so are the local capillary pressures.

\begin{table}[!htbp]
  \caption{\textbf{Supplementary metrics.}
  Rows and blocks as in Table~\ref{tab:all-results}.
  $\phi$: MAE relative to the mean absolute value of the reference field;
  IoU: overlap of non-wetting-phase voxels;
  $\mathbf{u}$: mean magnitude of the velocity error vector in normalized
  units;
  $S$: absolute saturation error;
  $H$: Wasserstein-1 distance between predicted and reference
  mean-curvature distributions, in units of the reference spread.
  Underfill descriptors ($S$, $H$) score the wetting phase;
  others the non-wetting phase. Means of per-run means; lower is better
  except IoU. Bold as in Table~\ref{tab:all-results}.}
  \label{tab:other-results}
  \centering
  \scriptsize
  \setlength{\tabcolsep}{1.5pt}
  \renewcommand{\arraystretch}{1.02}
  \begin{adjustbox}{max width=\textwidth}
  \begin{tabular}{@{}llcccccccccc@{}}
    \toprule
    & & \multicolumn{5}{c}{One-step prediction}
      & \multicolumn{5}{c}{Autoregressive rollout, step 64} \\
    \cmidrule(lr){3-7}\cmidrule(lr){8-12}
    Campaign & Model & $\phi$ & $\phi$ & $\mathbf{u}$ & $S$ & $H$
      & $\phi$ & $\phi$ & $\mathbf{u}$ & $S$ & $H$ \\
    & & Rel. MAE & IoU & MAE & Abs. err. & $W_1/\sigma$
      & Rel. MAE & IoU & MAE & Abs. err. & $W_1/\sigma$ \\
    \midrule
    \multirow{5}{*}{\shortstack[l]{Drainage\\(In-distribution)}} & UNet3D & \textbf{0.00278} & 0.987 & 0.00988 & \textbf{0.000205} & 0.0541 & 0.219 & 0.738 & 0.116 & 0.0127 & 1.24 \\
     & FNO3D & 0.00535 & 0.971 & 0.0119 & 0.000635 & 0.0658 & 0.262 & 0.685 & 0.140 & 0.0246 & 1.40 \\
     & P3D & 0.00344 & \textbf{0.988} & 0.0101 & 0.000206 & \textbf{0.0487} & 0.178 & 0.820 & 0.129 & 0.00776 & \textbf{0.677} \\
     & Transolver++ & 0.00532 & 0.976 & 0.0105 & 0.000373 & 0.120 & 0.303 & 0.641 & 0.0828 & 0.0292 & 2.72 \\
     & AB-UPT & 0.00398 & 0.934 & \textbf{0.00927} & 0.000417 & 0.136 & \textbf{0.129} & \textbf{0.830} & \textbf{0.0804} & \textbf{0.00730} & 0.946 \\
    \campaignrule
    \multirow{5}{*}{\shortstack[l]{Drainage\\(Scale-Up)}} & UNet3D & \textbf{0.00321} & 0.932 & \textbf{0.00752} & 0.000640 & 0.0987 & 0.223 & 0.698 & 0.103 & 0.0514 & 0.929 \\
     & FNO3D & 0.00595 & 0.883 & 0.0118 & 0.000805 & 0.136 & 0.215 & 0.729 & 0.153 & 0.0329 & 1.08 \\
     & P3D & 0.00405 & \textbf{0.936} & 0.00797 & 0.000763 & \textbf{0.0893} & 0.212 & 0.740 & 0.122 & 0.0471 & \textbf{0.680} \\
     & Transolver++ & 0.00665 & 0.904 & 0.00808 & 0.000596 & 0.423 & 0.309 & 0.604 & \textbf{0.0727} & 0.0694 & 2.16 \\
     & AB-UPT & 0.00439 & 0.916 & 0.00869 & \textbf{0.000391} & 0.395 & \textbf{0.157} & \textbf{0.766} & 0.0780 & \textbf{0.0252} & 2.28 \\
    \campaignrule
    \multirow{5}{*}{\shortstack[l]{Drainage\\(Geo-Shift)}} & UNet3D & 0.00436 & \textbf{0.900} & 0.0106 & \textbf{0.000549} & 0.223 & 0.180 & 0.780 & 0.115 & \textbf{0.00801} & \textbf{0.858} \\
     & FNO3D & 0.00947 & 0.780 & 0.0142 & 0.00212 & \textbf{0.120} & 0.271 & 0.687 & 0.143 & 0.0728 & 1.66 \\
     & P3D & 0.00494 & 0.897 & 0.0110 & 0.000565 & 0.397 & 0.203 & 0.770 & 0.143 & 0.0114 & 0.868 \\
     & Transolver++ & 0.00796 & 0.864 & 0.0113 & 0.000684 & 0.861 & 0.333 & 0.592 & 0.101 & 0.0325 & 2.95 \\
     & AB-UPT & \textbf{0.00386} & 0.896 & \textbf{0.0101} & 0.000578 & 0.369 & \textbf{0.126} & \textbf{0.826} & \textbf{0.0879} & 0.00923 & 1.29 \\
    \campaignrule
    \multirow{5}{*}{\shortstack[l]{GDL\\(In-distribution)}} & UNet3D & \textbf{0.00526} & 0.973 & 0.0993 & 0.000665 & 0.0922 & 0.202 & \textbf{0.703} & 0.284 & 0.0145 & \textbf{1.18} \\
     & FNO3D & 0.00963 & 0.956 & 0.0994 & 0.000900 & \textbf{0.0705} & 0.296 & 0.575 & 0.233 & 0.0919 & 1.52 \\
     & P3D & 0.00637 & \textbf{0.978} & 0.103 & \textbf{0.000404} & 0.125 & 0.234 & 0.685 & 0.302 & 0.0397 & 1.82 \\
     & Transolver++ & 0.00631 & 0.956 & 0.105 & 0.00122 & 0.0953 & 0.235 & 0.614 & 0.393 & 0.0285 & 2.37 \\
     & AB-UPT & 0.0121 & 0.863 & \textbf{0.0983} & 0.000895 & 0.475 & \textbf{0.183} & 0.681 & \textbf{0.195} & \textbf{0.0136} & 2.27 \\
    \campaignrule
    \multirow{5}{*}{\shortstack[l]{GDL\\(Scale-Up)}} & UNet3D & \textbf{0.00773} & 0.899 & 0.0326 & 0.00107 & 0.295 & 0.219 & 0.624 & 0.218 & 0.0575 & \textbf{1.22} \\
     & FNO3D & 0.00871 & 0.851 & \textbf{0.0296} & 0.000976 & \textbf{0.0463} & 0.224 & 0.613 & 0.168 & 0.0728 & 1.85 \\
     & P3D & 0.00861 & \textbf{0.910} & 0.0361 & 0.00136 & 0.444 & 0.212 & \textbf{0.681} & 0.239 & 0.0304 & 1.91 \\
     & Transolver++ & 0.00833 & 0.851 & 0.0358 & 0.000984 & 0.118 & 0.200 & 0.636 & 0.342 & 0.0308 & 2.48 \\
     & AB-UPT & 0.00842 & 0.881 & 0.0355 & \textbf{0.000353} & 1.01 & \textbf{0.190} & 0.635 & \textbf{0.135} & \textbf{0.0248} & 2.44 \\
    \campaignrule
    \multirow{5}{*}{\shortstack[l]{GDL\\(Geo-Shift)}} & UNet3D & \textbf{0.0105} & 0.818 & 0.0684 & \textbf{0.00139} & 0.736 & 0.321 & 0.571 & 0.261 & 0.0782 & \textbf{0.951} \\
     & FNO3D & 0.0124 & 0.687 & 0.0683 & 0.00411 & 0.461 & 0.266 & 0.614 & 0.196 & 0.0706 & 1.29 \\
     & P3D & 0.0130 & \textbf{0.839} & 0.0803 & 0.00145 & 0.449 & 0.227 & \textbf{0.695} & 0.302 & 0.0269 & 1.69 \\
     & Transolver++ & 0.0203 & 0.660 & 0.0686 & 0.00527 & \textbf{0.0899} & 0.371 & 0.439 & 0.286 & 0.0656 & 1.89 \\
     & AB-UPT & 0.0119 & 0.782 & \textbf{0.0670} & 0.00229 & 0.468 & \textbf{0.188} & 0.691 & \textbf{0.178} & \textbf{0.0174} & 1.62 \\
    \campaignrule
    \multirow{5}{*}{\shortstack[l]{Trapping\\(In-distribution)}} & UNet3D & \textbf{0.00480} & \textbf{0.996} & 0.0125 & 0.000478 & 0.0246 & 0.406 & 0.634 & 0.114 & 0.0564 & 0.948 \\
     & FNO3D & 0.00677 & 0.993 & 0.0135 & 0.000934 & 0.0514 & 0.297 & 0.740 & 0.123 & 0.0537 & 1.01 \\
     & P3D & 0.00588 & 0.995 & 0.0129 & 0.000480 & \textbf{0.0235} & 0.371 & 0.681 & 0.154 & 0.0527 & \textbf{0.889} \\
     & Transolver++ & 0.00644 & 0.994 & 0.0129 & 0.000688 & 0.0238 & 0.295 & \textbf{0.762} & 0.0968 & 0.0619 & 1.46 \\
     & AB-UPT & 0.00489 & 0.995 & \textbf{0.0124} & \textbf{0.000265} & 0.0268 & \textbf{0.294} & 0.727 & \textbf{0.0960} & \textbf{0.0366} & 1.11 \\
    \campaignrule
    \multirow{5}{*}{\shortstack[l]{Trapping\\(Scale-Up)}} & UNet3D & \textbf{0.00487} & \textbf{0.996} & \textbf{0.00935} & 0.000376 & \textbf{0.0142} & 0.321 & 0.722 & 0.108 & 0.0309 & \textbf{0.808} \\
     & FNO3D & 0.0107 & 0.995 & 0.0109 & 0.000874 & 0.0207 & 0.327 & 0.717 & 0.115 & 0.0795 & 0.996 \\
     & P3D & 0.00608 & \textbf{0.996} & 0.00992 & \textbf{0.000246} & 0.0150 & 0.319 & 0.752 & 0.149 & 0.0328 & 0.902 \\
     & Transolver++ & 0.00720 & \textbf{0.996} & 0.00980 & 0.000322 & 0.0172 & \textbf{0.211} & \textbf{0.837} & \textbf{0.0896} & \textbf{0.0263} & 1.25 \\
     & AB-UPT & 0.00762 & \textbf{0.996} & 0.00956 & 0.000341 & 0.0328 & 0.319 & 0.720 & 0.103 & 0.0473 & 1.35 \\
    \campaignrule
    \multirow{5}{*}{\shortstack[l]{Trapping\\(Geo-Shift)}} & UNet3D & 0.00775 & \textbf{0.994} & \textbf{0.0142} & 0.000829 & 0.0316 & 0.344 & 0.678 & 0.137 & \textbf{0.0427} & \textbf{0.731} \\
     & FNO3D & 0.00901 & 0.993 & 0.0143 & 0.000910 & \textbf{0.0270} & 0.360 & 0.677 & 0.154 & 0.0555 & 1.14 \\
     & P3D & 0.00950 & 0.993 & 0.0147 & 0.000672 & 0.0295 & 0.445 & 0.611 & 0.172 & 0.0561 & 0.940 \\
     & Transolver++ & 0.00861 & 0.993 & 0.0144 & 0.000669 & 0.0319 & 0.350 & \textbf{0.714} & \textbf{0.108} & 0.0781 & 1.77 \\
     & AB-UPT & \textbf{0.00749} & \textbf{0.994} & 0.0143 & \textbf{0.000446} & 0.0326 & \textbf{0.335} & 0.702 & 0.139 & 0.0432 & 1.29 \\
    \campaignrule
    \multirow{5}{*}{\shortstack[l]{Underfill\\(In-distribution)}} & UNet3D & \textbf{0.00205} & 0.998 & 0.0330 & 0.000130 & 0.0924 & 0.108 & 0.950 & 0.124 & 0.0257 & 0.881 \\
     & FNO3D & 0.0114 & 0.998 & 0.0693 & 0.000201 & 0.0916 & 0.245 & 0.813 & 0.163 & 0.0964 & 2.24 \\
     & P3D & 0.00224 & \textbf{0.999} & 0.0329 & \textbf{0.000101} & \textbf{0.0675} & 0.115 & 0.954 & 0.125 & 0.0200 & 0.954 \\
     & Transolver++ & 0.00438 & 0.995 & 0.0432 & 0.000470 & 0.182 & 0.175 & 0.891 & 0.0900 & 0.0456 & 2.04 \\
     & AB-UPT & 0.00277 & 0.996 & \textbf{0.0230} & 0.000158 & 0.174 & \textbf{0.0323} & \textbf{0.983} & \textbf{0.0661} & \textbf{0.00561} & \textbf{0.779} \\
    \bottomrule
  \end{tabular}
  \end{adjustbox}
\end{table}

\section{Geometric Metrics}
\label{app:metrics}

\subsection{Curvature and Capillary Pressure}
\label{app:metrics:curvature}

In full, the Young--Laplace equation of Section~\ref{sec:dataset:metrics}
reads
\begin{equation}
\label{eq:young-laplace}
p_c=p_{\mathrm{nw}}-p_{\mathrm{w}}=2\sigma H
=\sigma(\kappa_1+\kappa_2),
\end{equation}
where $p_c$ is the capillary pressure, $p_{\mathrm{nw}}$ and $p_{\mathrm{w}}$
are the local non-wetting- and wetting-phase pressures, $\sigma$ is the
fluid-fluid interfacial tension, and $H=(\kappa_1+\kappa_2)/2$ is the mean
curvature. The principal curvatures $\kappa_1=1/r_1$ and $\kappa_2=1/r_2$
are defined by the signed principal radii $r_1$ and $r_2$, with positive
curvature for a convex non-wetting body. At fixed interfacial tension,
a curvature error therefore translates directly into an error in the
inferred capillary pressure, even when predicted phase volume is correct.
To measure this geometry, the curvature pipeline extracts a marching-cubes
surface, applies non-shrinking Taubin smoothing, fits local principal
curvatures, and excludes a neighborhood of the three-phase contact line
\citep{aljaberi2026spatially}. The integral of Eq.~\ref{eq:minkowski-m2}
runs over this fluid--fluid interface, outside the excluded neighborhood, so
$M_2$ is the total mean curvature of the menisci; a complete spherical
interface of radius $r$ gives $4\pi r$.

\subsection{Error Measures}
\label{app:metrics:errors}

A metric composes an error measure $e$ with a descriptor $d$ as
$e(d(\widehat{\mathbf{x}}),d(\mathbf{x}))$, where $\widehat{\mathbf{x}}$ and
$\mathbf{x}$ are the predicted and reference fields. Field errors are the
mean absolute error of $p$ and $\phi$ over the pore voxels of the sample, in
the normalized units of the stored frames: an interface that arrives one
voxel early is physically benign yet costly under a squared error, whereas
MAE keeps the units of the field and grows linearly with misplaced phase
volume. For $M_0$, $M_1$, and $M_2$ we report the relative error
$|M_k(\widehat{X}_t)-M_k(X_t)|/|M_k(X_t)|$, which makes them comparable across
campaigns and domain sizes. For $M_3$ we report the MAE of the Euler
characteristic $\chi=M_3/4\pi$ per unit volume of the injected phase: for
each run, the sum over its scored frames of $|\chi(\widehat{X}_t)-\chi(X_t)|$
divided by the summed reference phase volume of those frames, then the
mean over runs, in miscounts per $10^5$ voxels. $\chi$ is a signed integer
that passes through zero as ganglia form and loops close, so a relative
error is undefined there, whereas the miscount is defined on every frame
and reads as the number of components, loops, and cavities a prediction
gets wrong.

\section{The color-gradient lattice Boltzmann model}
\label{app:model}

\subsection{Color-gradient lattice Boltzmann method}
\label{app:model:basics}

At the continuum scale, pore-scale immiscible flow is governed by the
Navier--Stokes equations coupled to interfacial forces and wetting boundary
conditions. The color-gradient lattice Boltzmann method (CG-LBM) represents
this physics through component distribution functions whose hydrodynamic
limit recovers the Navier--Stokes equations, thereby advancing fluid motion
and interface evolution on a common lattice. Introduced by
\citet{gunstensen1991lattice}, the method has been successively developed in
its recoloring, forcing, and wetting formulations
\citep{latvakokko2005diffusion,reis2007lattice,halliday2007lattice,
leclaire2017generalized,akai2018wetting}.

LBM is particularly well suited to pore-scale simulation: complex solid
boundaries are represented by local lattice rules, allowing voxelized
micro-CT images to serve directly as computational domains without an
explicit meshing stage, while the local update structure maps efficiently to
parallel hardware \citep{liu2016multiphase}.

\begin{figure}[!htbp]
\begin{center}
\includegraphics[width=0.4\linewidth]{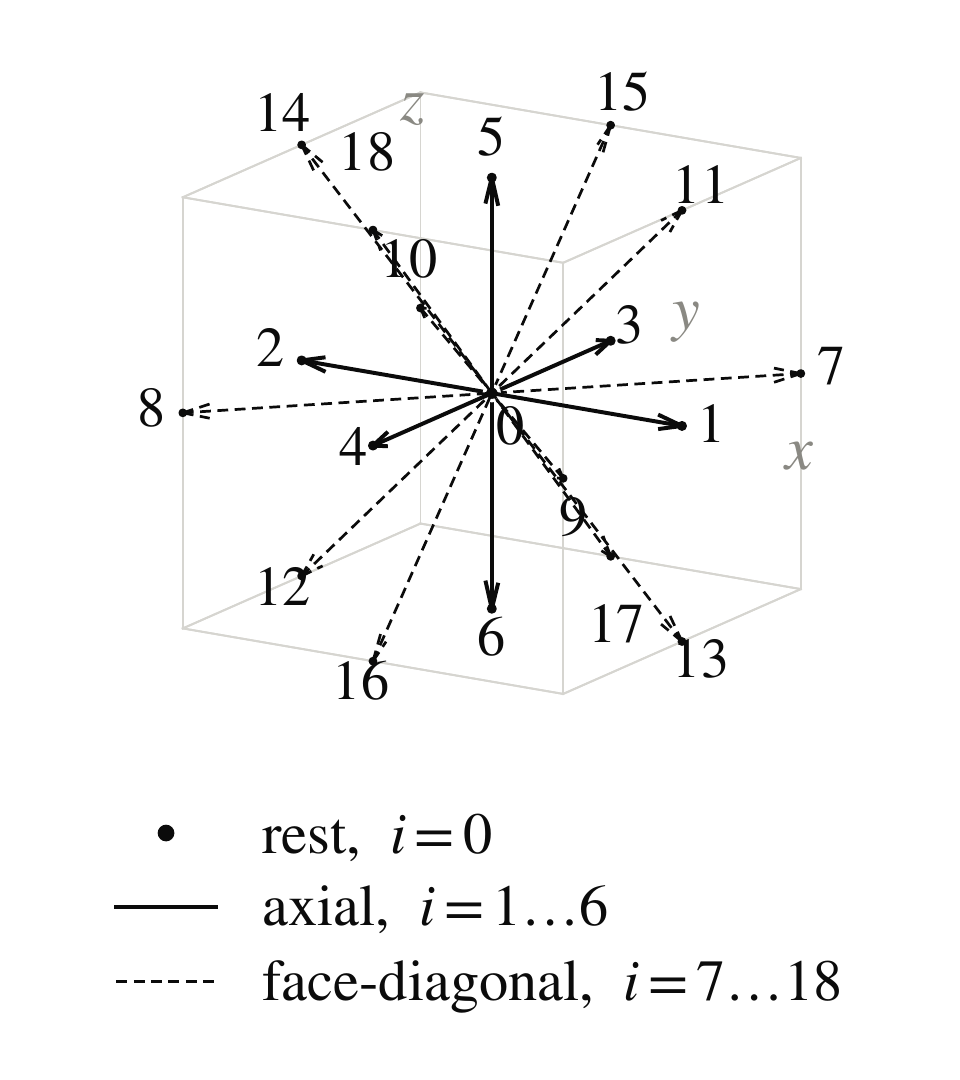}
\end{center}
\caption{The D3Q19 lattice stencil of Eq.~\ref{eq:lattice}: a rest
velocity, six axial neighbors (solid), and twelve face-diagonal neighbors
(dashed). Arrows are numbered in the solver's direction ordering, with
opposite directions in adjacent pairs.}
\label{fig:d3q19}
\end{figure}

We consider two immiscible fluids, labeled red (the non-wetting phase) and
blue (the wetting phase), with particle distribution functions
$f_i^{k}(\vx, t)$, $k \in \{r, b\}$, defined on the D3Q19 lattice
(Figure~\ref{fig:d3q19}) with discrete velocities
\begin{equation}
\label{eq:lattice}
\ve_i =
\begin{cases}
(0, 0, 0), & i = 0,\\
(\pm 1, 0, 0),\ (0, \pm 1, 0),\ (0, 0, \pm 1), & i = 1, \ldots, 6,\\
(\pm 1, \pm 1, 0),\ (\pm 1, 0, \pm 1),\ (0, \pm 1, \pm 1), & i = 7, \ldots, 18,
\end{cases}
\end{equation}
where lattice spacing and time step are set to unity. The color densities
and the total fluid density and momentum are moments of the distributions,
\begin{equation}
\label{eq:moments}
\rho_k = \sum_i f_i^{k}, \qquad
\rho = \rho_r + \rho_b, \qquad
\rho \vu = \sum_i f_i \ve_i + \tfrac{1}{2} {\bm{F}},
\end{equation}
where $f_i = f_i^{r} + f_i^{b}$ is the total distribution, ${\bm{F}}$ is
the total body force, and the $\tfrac{1}{2}{\bm{F}}$ term is the standard
half-force correction of \citet{guo2002discrete}. One time step applies,
in order: collision with forcing, recoloring, and streaming.

\textbf{Collision.} The total distribution relaxes toward local
equilibrium,
\begin{equation}
\label{eq:collision}
f_i^{*}(\vx, t) = f_i(\vx, t) + \Omega_i(\vx, t) + \Phi_i(\vx, t),
\end{equation}
where $f_i^{*}$ is the post-collision distribution, $\Omega_i$ the
collision operator, and $\Phi_i$ the forcing source term. In the
single-relaxation-time form \citep{qian1992lattice},
$\Omega_i = -\frac{1}{\tau} ( f_i - f_i^{\mathrm{eq}} )$, with the
second-order equilibrium
\begin{equation}
\label{eq:feq}
f_i^{\mathrm{eq}} = \rho\, w_i \left[
  1 + \frac{\ve_i \cdot \vu}{c_s^2}
    + \frac{(\ve_i \cdot \vu)^2}{2 c_s^4}
    - \frac{\vu \cdot \vu}{2 c_s^2} \right],
\qquad
w_i =
\begin{cases}
1/3, & i = 0,\\
1/18, & i = 1, \ldots, 6,\\
1/36, & i = 7, \ldots, 18,
\end{cases}
\end{equation}
where $c_s^2 = 1/3$ is the lattice sound speed. Chapman--Enskog analysis
recovers the Navier--Stokes equations with kinematic viscosity
$\nu = c_s^2 (\tau - 1/2)$. In production we replace the
single-relaxation-time operator by its multiple-relaxation-time
generalization \citep{dhumieres2002multiple}, detailed in
Section~\ref{app:solver:numerics}. A viscosity contrast between the phases is
realized by assigning each fluid its own relaxation time and blending the
inverse viscosities harmonically across the interface, which keeps the shear
stress continuous.

Spatially varying body forces (the interfacial force of
Section~\ref{app:model:surface} and any external drive) enter through the
source term of \citet{guo2002discrete},
\begin{equation}
\label{eq:guo}
\Phi_i = w_i \left( 1 - \frac{1}{2\tau} \right)
\left[ \frac{\ve_i - \vu}{c_s^2}
     + \frac{(\ve_i \cdot \vu)}{c_s^4}\, \ve_i \right] \cdot {\bm{F}},
\end{equation}
evaluated with the force-corrected velocity of Eq.~\ref{eq:moments}.

\textbf{Phase field and recoloring.} The phase composition is tracked by
the color function
\begin{equation}
\label{eq:phi}
\rho^{N} = \frac{\rho_r - \rho_b}{\rho_r + \rho_b} \in [-1, 1],
\end{equation}
whose gradient locates the interface; the interface unit normal is
$\vn = -\nabla \rho^{N} / |\nabla \rho^{N}|$. All spatial derivatives of
the phase field are evaluated with the fourth-order isotropic scheme
\begin{equation}
\label{eq:iso}
\frac{\partial \psi(\vx)}{\partial x_\alpha}
  = \frac{1}{c_s^2} \sum_i w_i\, \psi(\vx + \ve_i)\, e_{i\alpha},
\end{equation}
for any field $\psi$. Collision does not by itself keep the fluids
separated; immiscibility is enforced by the recoloring step of
\citet{latvakokko2005diffusion}, which splits the post-collision
distribution between the colors such that red and blue counter-diffuse
along the color gradient while total mass and momentum are conserved:
\begin{equation}
\label{eq:recolor}
f_i^{r} = \frac{\rho_r}{\rho} f_i^{*}
  + \beta\, \frac{\rho_r \rho_b}{\rho}\, w_i \cos \varphi_i\, |\ve_i|,
\qquad
f_i^{b} = \frac{\rho_b}{\rho} f_i^{*}
  - \beta\, \frac{\rho_r \rho_b}{\rho}\, w_i \cos \varphi_i\, |\ve_i|,
\end{equation}
where $\cos \varphi_i = \ve_i \cdot \nabla\rho^{N} /
(|\ve_i|\, |\nabla\rho^{N}|)$ is the angle between the lattice vector and
the color gradient, and the segregation parameter $\beta \in (0, 1]$
controls the interface width (we use $\beta = 0.99$). Eq.~\ref{eq:recolor}
includes the $|\ve_i|$ factor as in \citet{akai2018wetting} and
\citet{sedahmed2024}; the variant without it
\citep{leclaire2017generalized} conserves the same invariants, and the
difference is confined to the diagonal links. After recoloring, each color
streams to its neighbors,
$f_i^{k}(\vx + \ve_i,\, t + 1) = f_i^{k}(\vx, t)$.

\FloatBarrier

\subsection{Interfacial tension}
\label{app:model:surface}

Two established formulations generate interfacial tension in the CG-LBM.
The \emph{perturbation} operator of \citet{reis2007lattice}, in its
three-dimensional form \citep{leclaire2017generalized}, adds an
anisotropic stress contribution along the color gradient
${\bm{C}} = \nabla \rho^{N}$ during collision,
\begin{equation}
\label{eq:perturb}
\Delta f_i = \frac{9}{4}\, \omega\, \sigma\, |{\bm{C}}|
\left[ w_i \frac{(\ve_i \cdot {\bm{C}})^2}{|{\bm{C}}|^2} - B_i \right],
\end{equation}
where $\sigma$ sets the surface tension, $\omega = 1/\tau$, and the $B_i$
are lattice constants chosen so that mass and momentum are conserved.
Alternatively, the \emph{continuum surface force} (CSF) of
\citet{brackbill1992continuum}, introduced into the CG-LBM by
\citet{halliday2007lattice}, applies the interfacial force as a body force
through Eq.~\ref{eq:guo},
\begin{equation}
\label{eq:csf}
{\bm{F}}_s = \tfrac{1}{2}\, \sigma\, \kappa\, \nabla \rho^{N},
\qquad
\kappa = -\nabla_{\!s} \cdot \vn,
\end{equation}
where $\kappa$ is the interface curvature and
$\nabla_{\!s} = (\mI - \vn \vn^{\top}) \nabla$ the surface gradient, all
derivatives evaluated with Eq.~\ref{eq:iso}. Both formulations recover the
Young--Laplace law with closely similar interface physics. We pair each wetting boundary condition with the
surface-tension formulation of its source publication
(Section~\ref{app:model:wetting}), so that every configuration we run
matches a published, validated combination.

\subsection{Wetting boundary condition}
\label{app:model:wetting}

At the pore scale, wettability controls capillary entry, imbibition, and
trapping. We define the contact angle $\theta$ through the non-wetting red
phase, and require the boundary condition to impose it accurately on
voxelized surfaces.

The fictitious-density condition of \citet{latvakokko2005static} assigns a
color value to solid nodes and influences wetting indirectly through the
near-wall color gradient. Its realized angle can require calibration and its
accuracy deteriorates away from neutral wetting
\citep{leclaire2017generalized,akai2018wetting}. Geometric conditions instead
reorient the interface normal to satisfy the prescribed wall angle directly.
Building on the diffuse-interface construction of \citet{ding2007wetting} and
its two-dimensional CG-LBM adaptation by \citet{xu2017lattice},
\citet{leclaire2017generalized} extended the approach to arbitrary
three-dimensional surfaces, while \citet{akai2018wetting} derived the
closed-form rotation used here. At boundary fluid nodes, the predicted
interface normal $\vn^{*}$ is replaced by $\vn^{\theta}$ such that
$\vn^{\theta}\cdot\vn_s=\cos\theta$; Leclaire obtains this direction
iteratively, whereas Akai selects between two closed-form candidates. Wall
normals $\vn_s$ are precomputed from the solid mask.

We implement both geometric conditions; a systematic comparison by
\citet{sedahmed2024} found the Akai scheme more accurate for static
contact angles and the Leclaire scheme more accurate for dynamic
capillary-tube flow, so retaining both is of practical value. Following
the pairing principle of Section~\ref{app:model:surface}, the Leclaire
condition runs with the Reis--Phillips perturbation operator
(Eq.~\ref{eq:perturb}) and the Akai condition with the CSF force
(Eq.~\ref{eq:csf}) together with the solid-node extrapolation of the color
function and interface normal of \citet{sedahmed2024}; the Akai--CSF
combination is the solver default.

\section{Solver implementation}
\label{app:solver}

\subsection{Overview}
\label{app:solver:why}

Established porous-media implementations include the GPU-enabled LBPM
package \citep{mcclure2021lbpm} and the Fortran/MPI MF-LBM solver
\citep{chen2019inertial}. Our solver provides a complementary, modern
implementation written end-to-end in JAX \citep{jax2018github}, retaining
high-performance accelerator execution while integrating directly with
Python-based machine-learning workflows.

The complete time step --- collision, perturbation or CSF forcing,
recoloring, wetting reorientation, streaming, and boundary conditions --- is
a pure function on arrays, JIT-compiled and composable with
\texttt{scan}; it supports CPU, single-GPU, and multi-GPU execution through
halo exchange. The implementation combines geometric wetting with a
configuration-driven validation suite evaluated below against published
analytical solutions and pore-scale experiments.

\subsection{Numerics}
\label{app:solver:numerics}

\textbf{Collision.} We use a multiple-relaxation-time (MRT) collision
operator \citep{dhumieres2002multiple}: populations are mapped to moment
space, each moment relaxes at its own rate, and the result maps back. The
shear-stress moments relax at the rate $\omega$ set by the fluid viscosity.
For the remaining bulk and non-hydrodynamic (ghost) moments the solver offers
two spectra. The first uses fixed rates (Table~\ref{tab:params}); the second
ties every non-shear moment to the local shear rate as $\chi\omega$ with
$0<\chi\le1$, which improves stability and reduces to single-relaxation-time
collision at $\chi=1$; we use $\chi=0.8$ as in
\citet{leclaire2017generalized}. Drainage, trapping, and GDL intrusion use
the second, underfill the first. Either choice suppresses the staggered-mode
instabilities that single-relaxation-time (BGK) collision develops at low
viscosity and near walls. Body forces enter through the scheme of
\citet{guo2002discrete}, applied per moment with the standard $(1 - S/2)$
correction.

\textbf{Viscosity contrast.} The two fluids may have different viscosities
(relaxation rates $\omega_r$, $\omega_b$); in interface cells the local rate
is the harmonic blend of the two inverse viscosities weighted by the color
field, which keeps the stress continuous across the interface.

\textbf{Walls.} Solid nodes are handled by half-way (link) bounce-back:
each color's populations reflect at the wall link midpoint, solids store no
fluid state, and no color leaks through the wall. Wall normals for the
wetting condition are precomputed once per geometry from the solid mask.

\subsection{Boundary conditions}
\label{app:solver:bc}

The solver provides the following boundary conditions; the first three drive
the dataset protocols (Appendix~\ref{app:setup}):
\begin{itemize}
\item \textbf{Piston (velocity) inlet} --- the non-wetting phase is injected at
  an exact volumetric rate $u_{\mathrm{in}} A$ per step behind a moving
  piston front, with the inlet pressure left floating. This drives the
  displacement experiments, where the capillary number is set directly by
  $u_{\mathrm{in}}$.
\item \textbf{Pressure inlet/outlet} --- per-color Zou--He pressure
  boundaries \citep{zou1997pressure} fix the densities of reservoir planes.
  Holding $\Delta P = 0$ yields purely capillary-driven (spontaneous)
  imbibition; raising the inlet-outlet density difference in steps yields
  pressure-controlled drainage and capillary-pressure curves.
\item \textbf{Outflow} --- a density-rescaled copy of the penultimate plane
  lets the wetting phase leave the domain freely.
\item \textbf{Velocity inlet} --- the inlet plane is reset to the full
  equilibrium at the prescribed velocity with the density taken by zero
  gradient from its neighbor, which keeps every moment consistent and is
  stable under MRT collision.
\item \textbf{Reservoir inlet} --- a slab of pure non-wetting fluid held at
  equilibrium, the Dirichlet counterpart of the piston.
\item \textbf{Periodic faces and solid walls} --- streaming is periodic by
  default; any face can be sealed with half-way bounce-back.
\end{itemize}

\subsection{Parameters}
\label{app:solver:params}

Table~\ref{tab:params} collects the lattice constants and default numerical
parameters; Table~\ref{tab:mapping} gives the mapping from lattice
parameters to the physical control variables used throughout the paper. The
full D3Q19 moment matrix is the standard set of
\citet{dhumieres2002multiple}.

\begin{table}[!htbp]
\caption{Lattice constants and default numerical parameters.}
\label{tab:params}
\centering
\small
\begin{tabularx}{\linewidth}{@{}L{0.38\linewidth}Y@{}}
\toprule
\textbf{Quantity} & \textbf{Value} \\
\midrule
Lattice & D3Q19 (rest + 6 axial + 12 diagonal) \\
Weights $w_i$ & $1/3,\; 1/18,\; 1/36$ (rest / axial / diagonal) \\
Sound speed & $c_s^2 = 1/3$ \\
Recoloring strength & $\beta = 0.99$ (Eq.~\ref{eq:recolor}) \\
MRT non-shear rates & $0.8\,\omega$ (drainage, trapping, GDL);
  fixed $1.19,\; 1.2,\; 1.98,\; 1.4$ (underfill) \\
Wall BC & half-way bounce-back \\
Wetting BC & Akai (geometric); Leclaire available \\
Surface tension & CSF (Eq.~\ref{eq:csf}); perturbation available \\
\bottomrule
\end{tabularx}
\end{table}

\begin{table}[!htbp]
\caption{Mapping from lattice parameters to physical control variables.}
\label{tab:mapping}
\centering
\small
\begin{tabularx}{\linewidth}{@{}L{0.32\linewidth}Y@{}}
\toprule
\textbf{Control variable} & \textbf{Definition} \\
\midrule
Capillary number $\mathrm{Ca}$ & $\nu_0\, u_{\mathrm{in}} / \sigma$ \\
Viscosity ratio $M$ & $\nu_{\mathrm{nw}} / \nu_{\mathrm{w}}$ \\
Contact angle $\theta$ & angle of the non-wetting red phase at the wall \\
Surface tension $\sigma$ & Laplace coefficient, $\Delta p = 2\sigma/R$ \\
\bottomrule
\end{tabularx}
\end{table}
\FloatBarrier

\section{Solver validation}
\label{app:validation}

We validate the solver against three-dimensional analytical benchmarks and
published pore-scale experiments. A summary is provided in
Table~\ref{tab:validation}.

\begin{table}[!htbp]
\caption{Validation suite at a glance. All cases are 3D (D3Q19).}
\label{tab:validation}
\centering
\small
\begin{tabularx}{\textwidth}{@{}
    >{\raggedright\arraybackslash}p{0.40\textwidth}
    >{\raggedright\arraybackslash}X@{}}
\toprule
\textbf{Case} & \textbf{Reference} \\
\midrule
Laplace law (\S\ref{app:val:laplace}) & $\Delta p = 2\sigma/R$ \\
Layered Poiseuille (\S\ref{app:val:layered}) & two-layer analytic profile \\
Contact angle, flat (\S\ref{app:val:flat}) & imposed $\theta$ and
  \citet{akai2018wetting} \\
Contact angle, curved (\S\ref{app:val:curved}) & sphere-on-sphere geometry \\
Washburn imbibition (\S\ref{app:val:washburn}) & two-fluid Washburn law \\
Capillary pressure (\S\ref{app:val:pc}) & curvature-correct entry
  pressures \\
Micromodel drainage and imbibition (\S\ref{app:val:micromodel}) &
  single-junction experiments \citep{zacharoudiou2017porefilling} \\
Displacement regimes (\S\ref{app:val:lenormand}) & Lenormand phase diagram \\
\bottomrule
\end{tabularx}
\end{table}
\FloatBarrier

\subsection{Laplace law}
\label{app:val:laplace}

\begin{figure}[!htbp]
\begin{center}
\includegraphics[width=\linewidth]{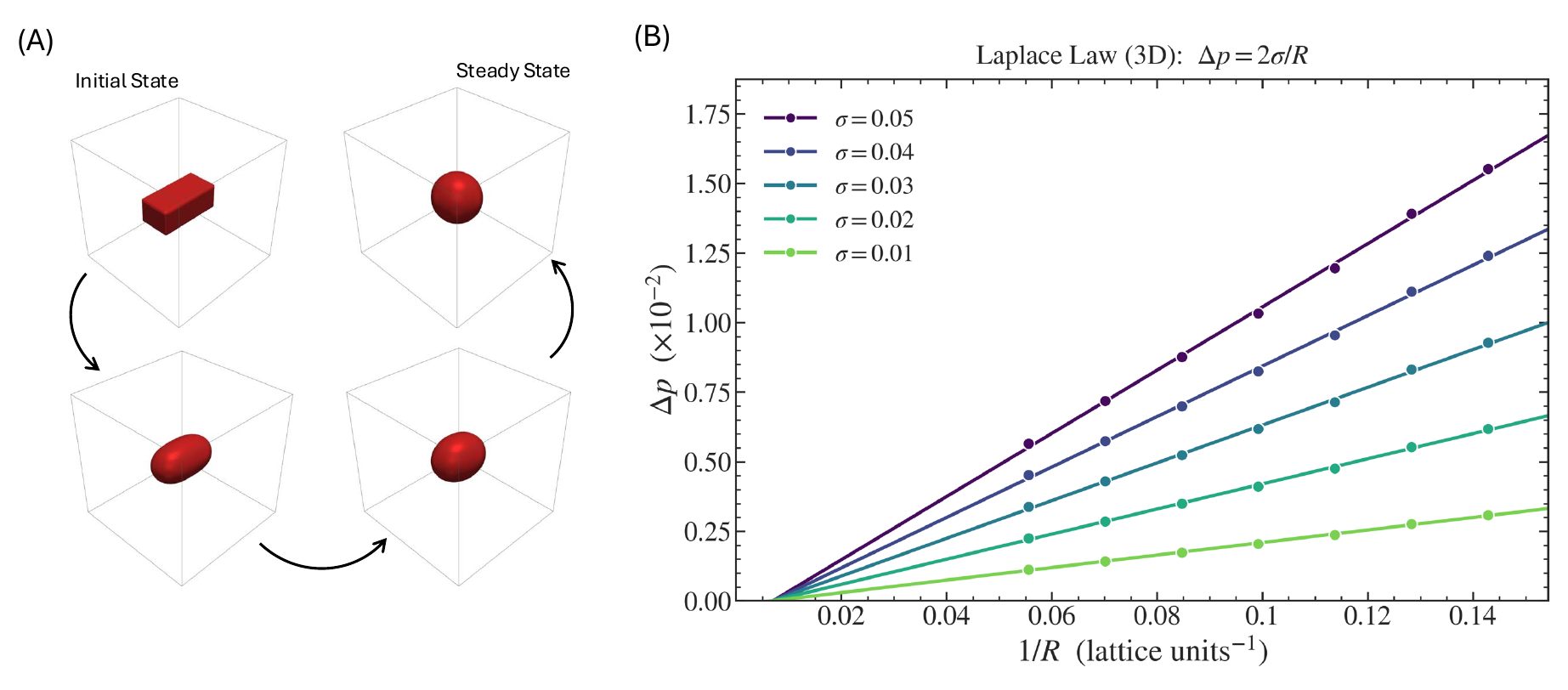}
\end{center}
\caption{Laplace-law validation. (A)~A deliberately non-spherical initial
slab relaxes to a sphere under the interfacial force alone (sphericity
$2.504 \to 1.000$). (B)~Measured pressure jump $\Delta p$ across
equilibrated 3D droplets versus inverse radius $1/R$, swept over input
surface tension $\sigma = 0.01$--$0.05$: one fitted line per $\sigma$,
each linear with $R^2 \approx 0.999$.}
\label{fig:val-laplace}
\end{figure}

\textbf{Purpose.} Surface tension drives every capillary phenomenon in the
dataset. This test determines whether the simulated pressure jump follows
the Young--Laplace relation over the surface-tension range used in the porous
cases.
\textbf{Setup.} Spherical droplets of the red fluid, over a range of radii
$R$, equilibrate in a periodic box of blue fluid; the input surface
tension is swept over $\sigma = 0.01$--$0.05$, the range the porous cases
use. A companion run starts from a deliberately non-spherical slab, which
the interfacial force alone must relax to a sphere. \textbf{Reference.}
The Young--Laplace law: the pressure jump across the interface is
$\Delta p = 2\sigma_{\mathrm{eff}}/R$. \textbf{Result.} The slab rounds to
a sphere (sphericity $2.504 \to 1.000$), and at every $\sigma$ the
measured $\Delta p$ is linear in $1/R$ across the radius range
(Figure~\ref{fig:val-laplace}; $R^2 \approx 0.999$ per fit).

\FloatBarrier

\subsection{Layered Poiseuille flow}
\label{app:val:layered}

\begin{figure}[!htbp]
\begin{center}
\includegraphics[width=0.5\linewidth]{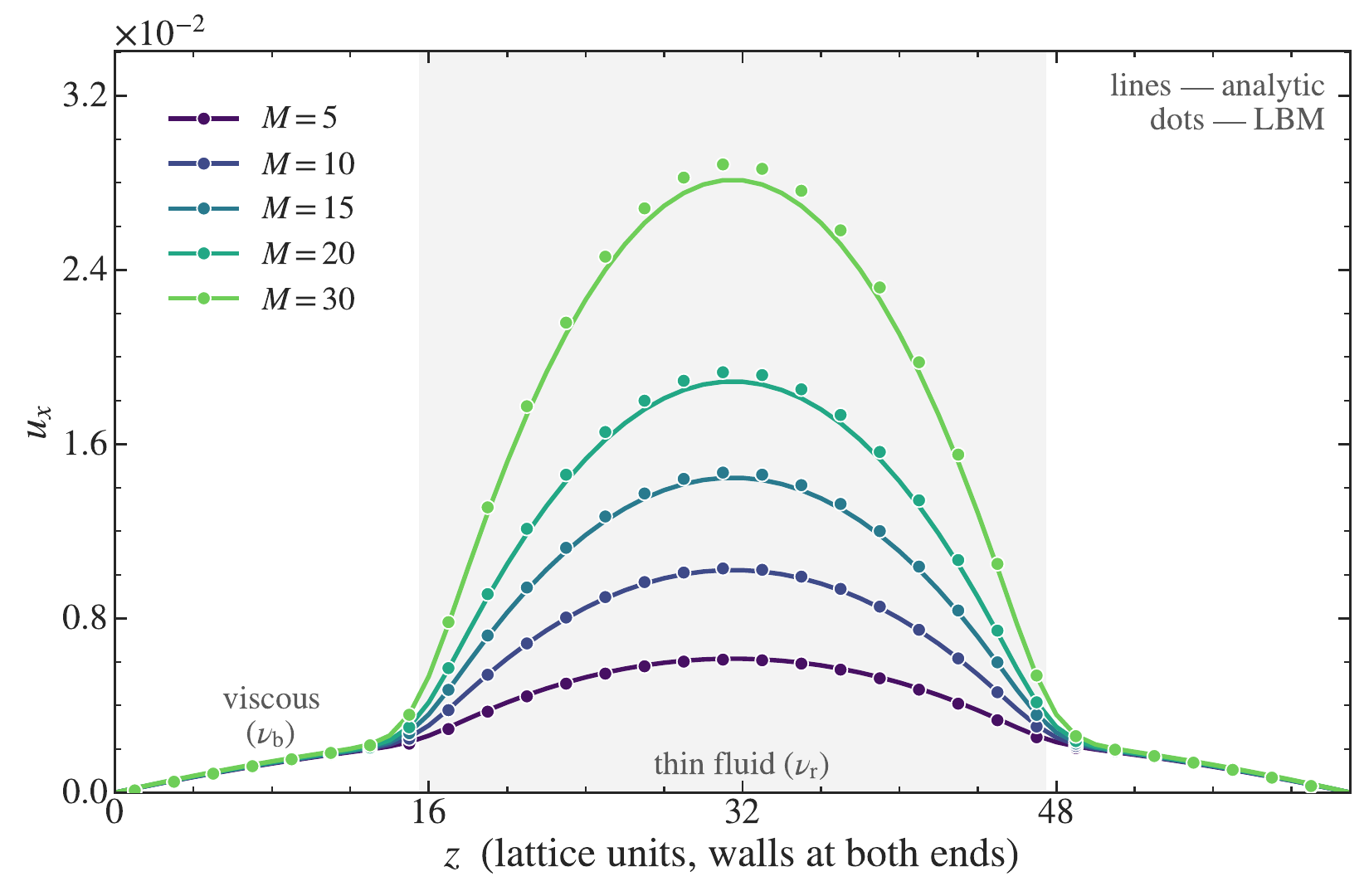}
\end{center}
\caption{Layered Poiseuille validation in 3D: steady velocity profiles
$u_x(z)$ for viscosity ratios $M = 5$--$30$ (LBM dots) against the
two-fluid analytic solution (lines). The shaded band marks the central
low-viscosity layer; maximum relative error is $2.58\%$ at $M = 30$.}
\label{fig:val-layered3d}
\end{figure}

\textbf{Purpose.} The viscosity ratio $M$ is one of the dataset's two
control knobs, and its physical meaning rests on correct viscous momentum
exchange across the fluid-fluid interface --- the mechanism that decides
whether a displacement front is stabilized or fingers. This case verifies
that exchange in the simplest configuration with an exact solution,
isolated from any capillary effect.
\textbf{Setup.} A $64 \times 8 \times 128$ channel with bounce-back walls
on the spanwise faces, filled with a symmetric three-band layering: the
more viscous fluid ($\nu_b = 0.3$) against both walls and a thinner fluid
in the central band, with the viscosity ratio $M = \nu_b/\nu_r$ swept over
$5$--$30$. The flow is driven by the per-color Zou--He pressure boundaries
of Section~\ref{app:solver:bc} rather than a body force, so the case also
exercises the pressure drive used in the dataset protocols.
\textbf{Reference.} The layered Poiseuille solution of
$\mathrm{d}/\mathrm{d}z\,(\nu\, \mathrm{d}u_x/\mathrm{d}z) = -G$,
evaluated on the realized diffuse interface profile and the realized
pressure gradient --- a piecewise parabola, nearly flat in the viscous
wall bands with a tall central peak that grows with $M$.
\textbf{Result.} At steady state the simulated profiles sit on the
analytic curves for all five ratios (Figure~\ref{fig:val-layered3d}), with
maximum relative error $0.95$--$2.58\%$, largest at $M = 30$ where the
center fluid reaches $\nu_r = 0.01$.

\FloatBarrier

\subsection{Contact angle on a flat wall}
\label{app:val:flat}

\begin{figure}[!htbp]
\begin{center}
\includegraphics[width=\linewidth]{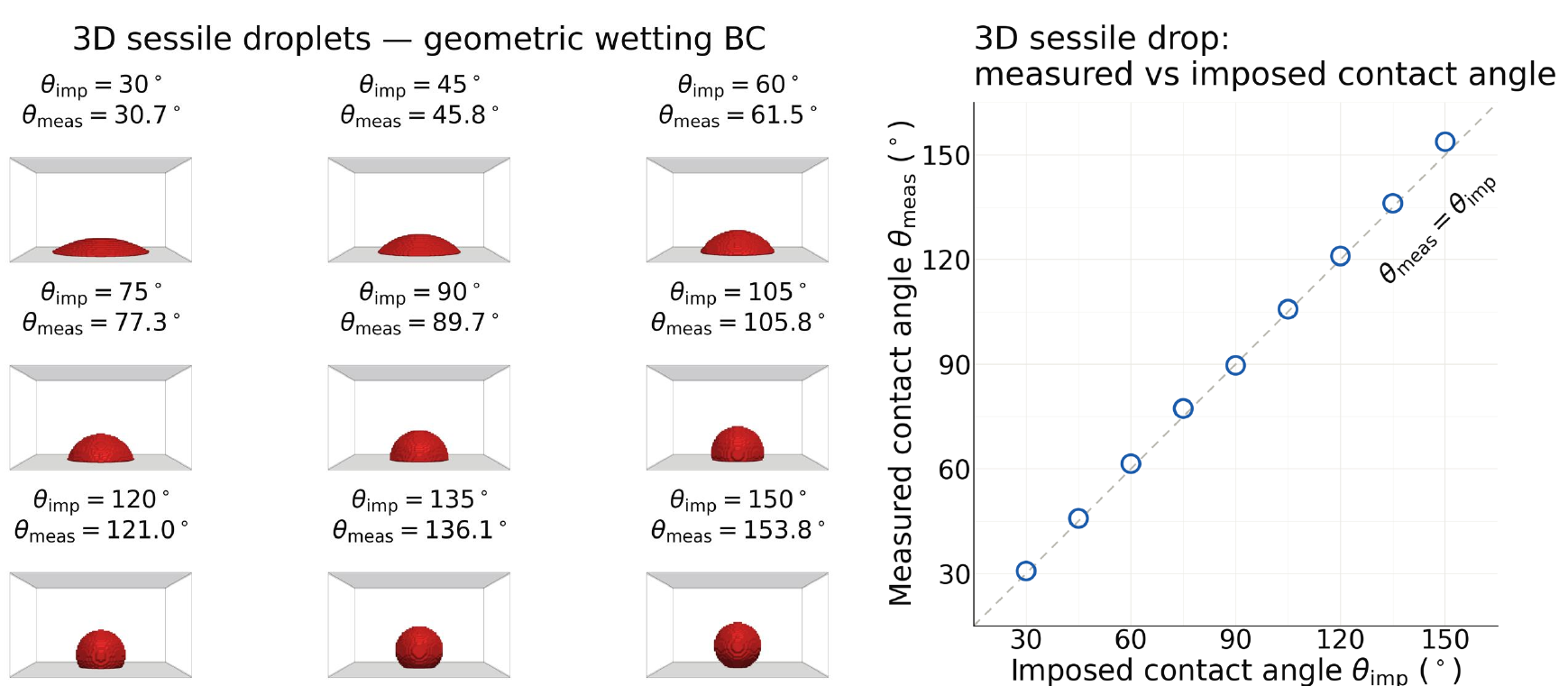}
\end{center}
\caption{Contact-angle validation on a flat wall. Left: equilibrated
sessile droplets for imposed angles $30^{\circ}$--$150^{\circ}$, with the
measured angle annotated per panel. Right: measured versus imposed angle
across the sweep. The realized angle is within $2.3^{\circ}$ of the
imposed value everywhere except the $150^{\circ}$ extreme
($3.8^{\circ}$).}
\label{fig:val-wetting-flat}
\end{figure}

\textbf{Purpose.} Wettability is a controlled parameter of the dataset:
every trajectory is labeled by its imposed contact angle, and those labels
are only trustworthy if the solver realizes the angle it is asked for.
The static accuracy established here underlies everything wettability
controls downstream --- entry pressures, imbibition rates, and residual
trapping.
\textbf{Setup.} Sessile red droplets on a flat wall relax to equilibrium
under imposed contact angles spanning $30$--$150^{\circ}$; the realized
angle is measured by fitting a spherical cap to the converged interface.
\textbf{Reference.} The imposed angle itself --- the geometric condition
makes $\theta$ a direct input, so the realized angle is a pure accuracy
check. \textbf{Result.} The measured angles follow a near-linear parity
relation with the imposed angles across the $30$--$150^{\circ}$ sweep
(Figure~\ref{fig:val-wetting-flat}), consistent with the analytical
validation reported by \citet{akai2018wetting}.

\FloatBarrier

\subsection{Contact angle on a curved wall}
\label{app:val:curved}

\begin{figure}[!htbp]
\begin{center}
\includegraphics[width=\linewidth]{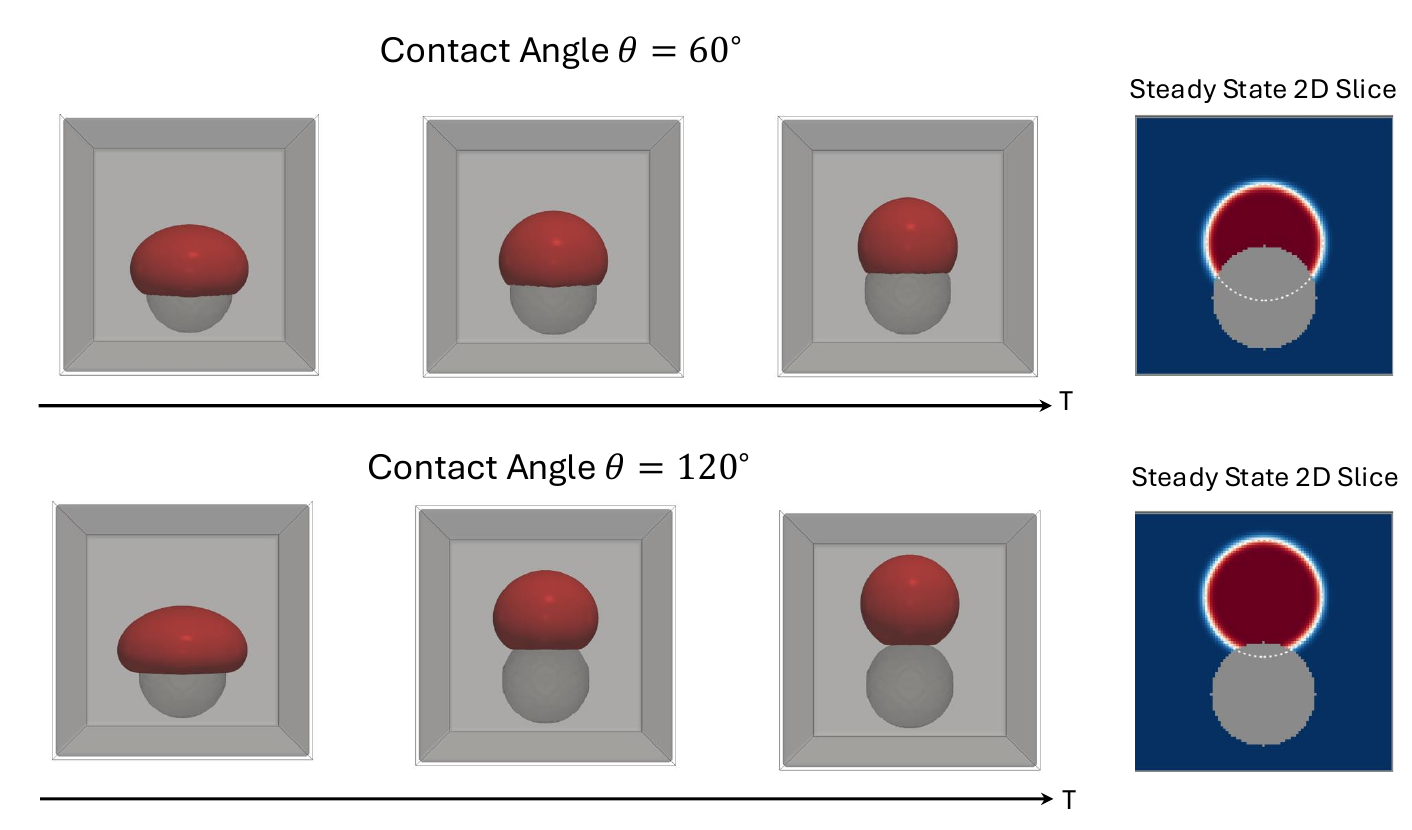}
\end{center}
\caption{Contact-angle validation on a curved wall for imposed
$\theta = 60^{\circ}$ (A, top row) and $\theta = 120^{\circ}$ (B, bottom
row). The first three panels of each row are snapshots of the droplet
(red) relaxing on the spherical grain (gray) between the volume-matched
half-shell collar and the equilibrium cap; the rightmost panel is the
final midplane slice, with the dotted circle marking the analytical
sphere-on-sphere solution.}
\label{fig:val-wetting-curved}
\end{figure}

\textbf{Purpose.} The dataset's geometries have no flat walls inside a sphere pack or rock crop: most
contact line lives on a curved, voxelized
surface, where the wall normals are computed from the solid mask rather
than known analytically. This case ensures the wetting fidelity of
Section~\ref{app:val:flat} carries over to the surfaces the dataset
actually uses.
\textbf{Setup.} Following \citet{akai2018wetting}, a red droplet relaxes
on a spherical solid grain, initialized as a volume-matched half-shell
collar coating the upper hemisphere. \textbf{Reference.} The analytic sphere-on-sphere
configuration: for droplet curvature radius $R_1$, grain radius $R_2$, and
contact angle $\theta$, the center-to-center distance satisfies
$R_3 = \sqrt{R_1^2 + R_2^2 - 2 R_1 R_2 \cos\theta}$.
\textbf{Result.} At $\theta = 60^{\circ}$ the fitted drop gives
$\theta_{\mathrm{meas}} = 60.2^{\circ}$ with $R_1 = 22.0$ against a target
of $22.5$; at $\theta = 120^{\circ}$,
$\theta_{\mathrm{meas}} = 124.4^{\circ}$ with $R_1 = 21.9$
(Figure~\ref{fig:val-wetting-curved}).

\FloatBarrier

\subsection{Washburn imbibition}
\label{app:val:washburn}

\begin{figure}[!htbp]
\begin{center}
\includegraphics[width=\linewidth]{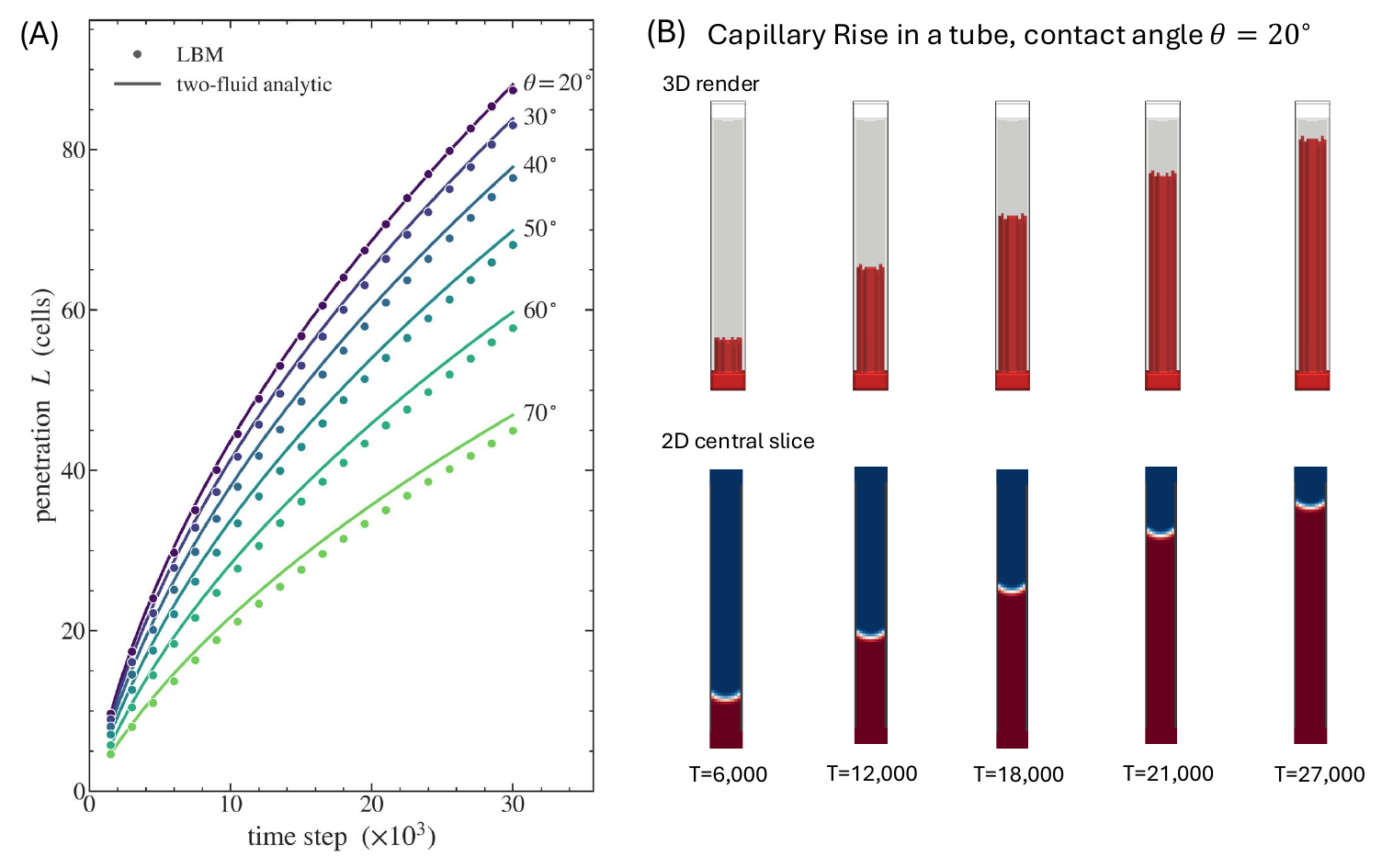}
\end{center}
\caption{Washburn imbibition validation. (A)~Penetration length $L(t)$
from the simulation (dots) against the two-fluid Washburn solution
(lines) for contact angles $\theta = 20$--$70^{\circ}$. (B)~The
$\theta = 20^{\circ}$ tube at five successive times, rendered upright
(wetting phase red, rising bottom to top under capillary suction alone,
$\Delta P = 0$): 3D view above, midplane slice below.}
\label{fig:val-washburn}
\end{figure}

\textbf{Purpose.} The benchmark's trajectories are time-resolved, so the
\emph{dynamics} must be right, not only the equilibria. Spontaneous
imbibition is the test of the moving contact line: the capillary
force at the wall is the only driver, and any error in dynamic wetting,
in the pressure boundaries, or in two-phase viscous friction shows up
directly in the penetration rate. Passing it gives confidence that the
time axis of every imbibition trajectory in the dataset is physical.
\textbf{Setup.} A wetting fluid spontaneously imbibes a cylindrical
capillary tube (radius $r = 5$) from a frictionless reservoir, displacing
the non-wetting fluid under zero applied pressure difference
($\Delta P = 0$ per-color pressure boundaries at both ends), for a sweep of
contact angles.
\textbf{Reference.} The two-fluid Washburn law
\citep{washburn1921dynamics}: the penetration length obeys
$L^2 \propto t$ when the displaced viscosity is negligible, with the
finite-viscosity two-fluid generalization used for the fit.
\textbf{Result.} The penetration curves follow the two-fluid analytic
solution across the full angle sweep (Figure~\ref{fig:val-washburn}): the
rectified curves are linear in $t$ ($R^2 = 1.000$ to three decimals for
every angle), and the measured
imbibition-rate coefficient is within $1$--$6\%$ of the analytic value
($A/A_{\mathrm{an}} = 1.01$ to $0.94$, decreasing monotonically from
$\theta = 20^{\circ}$ to $70^{\circ}$).

\FloatBarrier

\subsection{Drainage and capillary pressure}
\label{app:val:pc}

\begin{figure}[!htbp]
\begin{center}
\includegraphics[width=0.9\linewidth]{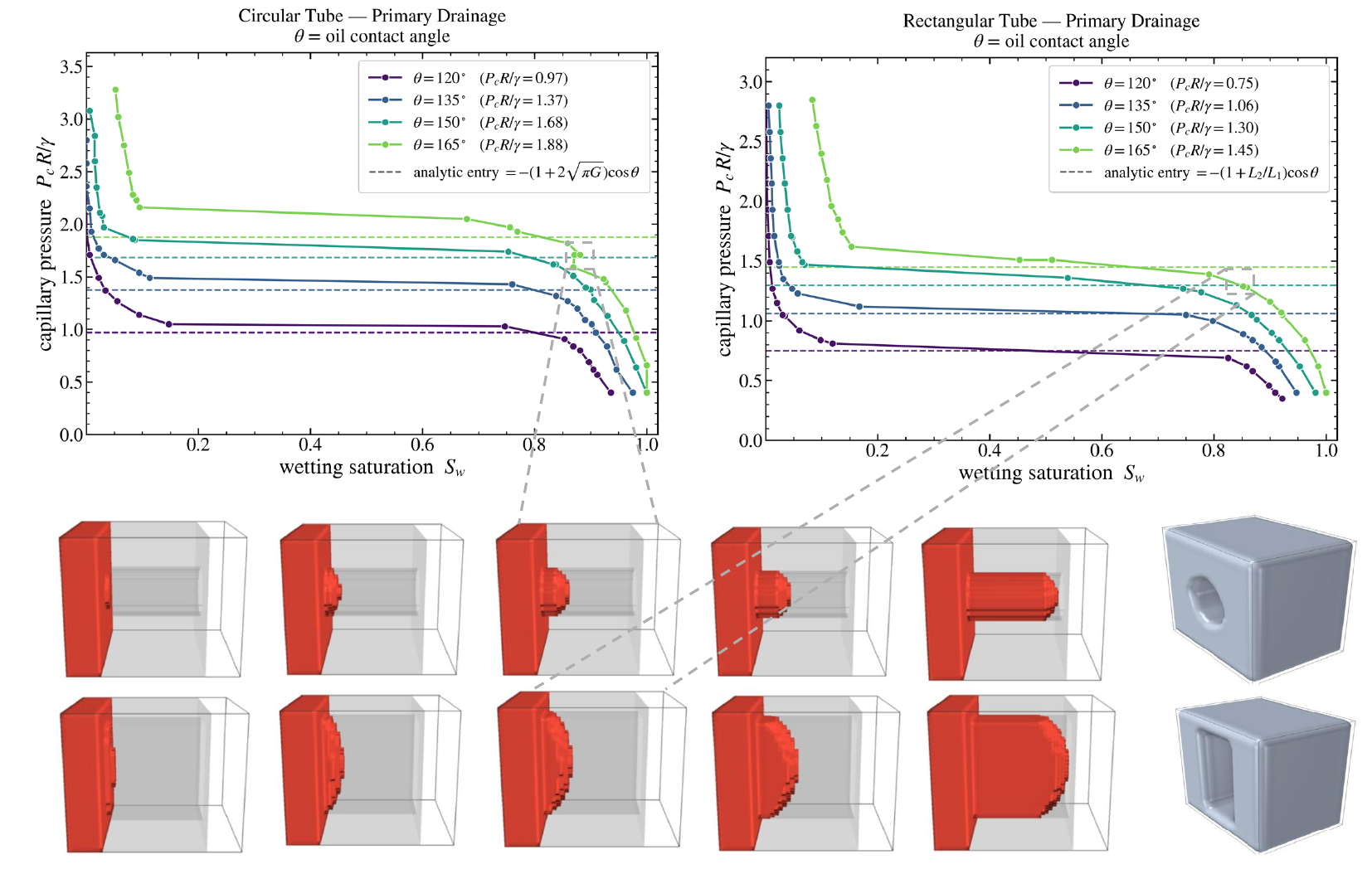}
\end{center}
\caption{Primary-drainage capillary-pressure curves ($P_c R / \gamma$
versus wetting saturation $S_w$) for the circular tube (top left) and the
rectangular tube (top right), swept over contact angle. Dashed lines mark
each angle's curvature-correct analytic entry pressure (circle:
Mason--Morrow shape-factor form on the voxelized cross-section;
rectangle: two-curvature Laplace). Both tubes drain along a flat plateau
that tracks the analytic entry per angle. Bottom: the non-wetting phase
(red) crossing the throat and filling the duct
piston-like (upper sequence: circular tube; lower sequence: rectangular
tube; geometry at right).}
\label{fig:val-pc}
\end{figure}

\textbf{Purpose.} Drainage through a pore network proceeds as a sequence
of threshold events: the non-wetting phase enters a throat only once the
applied pressure exceeds that throat's entry pressure, and each crossing
is released as a sudden, intermittent pore-filling event --- a Haines
jump. The invasion pathway, the macroscopic $P_c$--$S_w$ curve, and where
the wetting phase ends up trapped are all consequences of these
thresholds. Verifying entry pressures against curvature-correct analytic
values in elementary throat geometries therefore secures the mechanism by
which every drainage trajectory in the dataset unfolds.
\textbf{Setup.} Pressure-controlled primary drainage in two throat
geometries --- a voxelized circular tube (radius $r = 6$ in a
$32 \times 32$ cross-section) and an elongated rectangular tube
($24 \times 12$) --- each swept over non-wetting contact angles
$\theta = 120$--$165^{\circ}$ (wetting angles
$\theta_w = 15$--$60^{\circ}$). The non-wetting inlet density is fixed and
the outlet density lowered in a refined ladder of capillary pressures;
the equilibrium saturation of the duct pore is recorded at each step.
\textbf{Reference.} Curvature-correct analytic entry pressures: the
\citet{mason1991capillary} shape-factor form
$P_c R/\gamma = (1 + 2\sqrt{\pi G})\cos\theta_w$ for the circular tube,
with the inscribed radius $R$ and shape factor $G$ measured from the
voxelized cross-section the solver actually sees (the form reduces to the
Laplace value $2\cos\theta_w$ for a perfect circle); and the
two-curvature Laplace value $(1 + L_2/L_1)\cos\theta_w$ for the
rectangle, whose meniscus spans the short gap and curves more weakly
along the long side.
\textbf{Result.} In both tubes the drainage curve is a sharp step: the
non-wetting phase waits at the throat until the applied pressure crosses
the entry value, then sweeps the uniform cross-section piston-like, so
saturation drops along a flat plateau at the entry pressure. The realized
plateaus track each angle's analytic entry closely and are correctly
ordered and spaced across the four contact angles
(Figure~\ref{fig:val-pc}).

\FloatBarrier

\subsection{Drainage and imbibition compared to experiment in a micromodel}
\label{app:val:micromodel}

\begin{figure}[!htbp]
  \centering
  \includegraphics[width=0.55\linewidth]{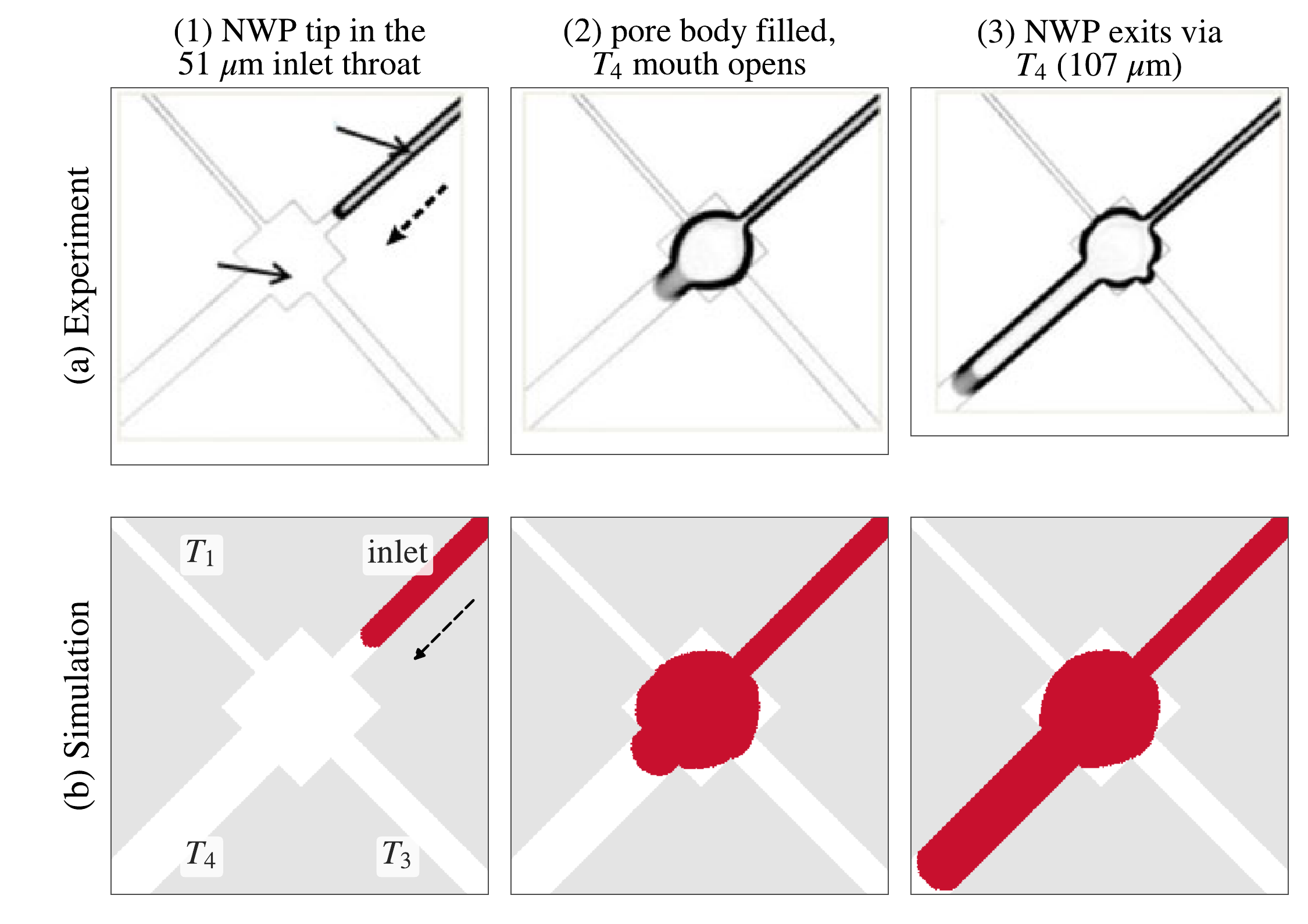}
  \caption{Dynamic primary drainage in a four-throat micromodel.
  (a)~Experimental images adapted from Figure~4(a) of
  \citet{zacharoudiou2017porefilling}.
  (b)~Midplane slices of our three-dimensional color-gradient LBM
  simulation at corresponding meniscus positions
  ($\mathrm{Ca}\approx10^{-4}$, imposed $\theta_w=26^{\circ}$).
  From left to right, the non-wetting phase advances through the inlet,
  fills the pore body, and invades the widest outlet throat, $T_4$
  ($107\,\mu\mathrm{m}$ in the experiment, $108\,\mu\mathrm{m}$ in
  the simulation); $T_1$ and $T_3$ remain predominantly wetting-filled.
  In the simulation, red denotes the non-wetting phase, white the wetting
  phase, and gray the solid. The dashed arrow indicates the inlet flow
  direction.}
  \label{fig:micromodel-drainage}
\end{figure}

\begin{figure}[!htbp]
  \centering
  \includegraphics[width=\linewidth]{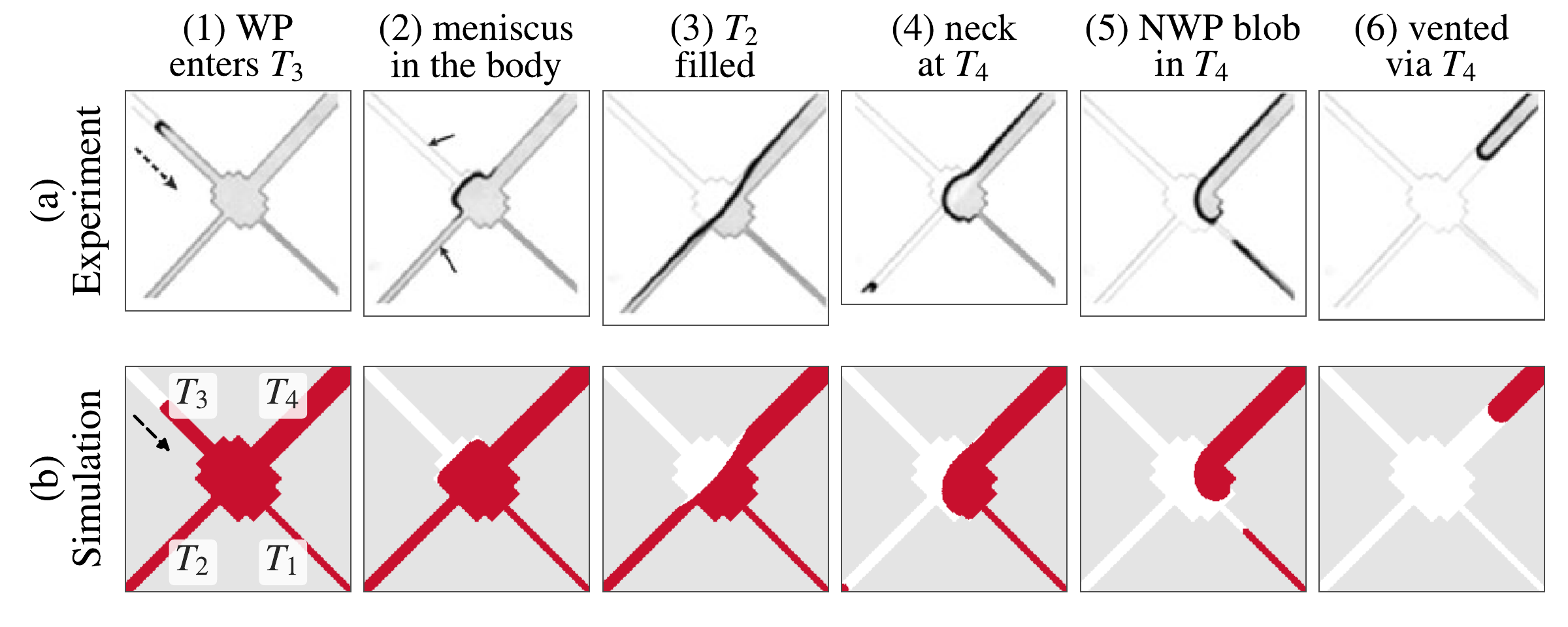}
  \caption{Spontaneous imbibition in a four-throat micromodel.
  (a)~Experimental images adapted from Figure~8(a) of
  \citet{zacharoudiou2017porefilling}.
  (b)~Midplane slices of our three-dimensional color-gradient LBM
  simulation at corresponding meniscus positions
  (imposed $\theta_w=16^{\circ}$, $\Delta P=0$).
  From left to right, the wetting phase enters through $T_3$, advances
  into the pore body, and invades the adjacent throat $T_2$ in stage (3),
  filling it by stage (4), before the narrowest throat $T_1$.
  The stage-(3) heading describes the experiment; the simulation at that
  stage shows partial invasion of $T_2$.
  Their experimental widths are $47$ and $27\,\mu\mathrm{m}$,
  represented by $48$ and $27\,\mu\mathrm{m}$ in the simulation.
  The remaining non-wetting phase then
  withdraws through the widest throat, $T_4$.
  In the simulation, red denotes the non-wetting phase, white the wetting
  phase, and gray the solid. The dashed arrow indicates the inlet flow
  direction.}
  \label{fig:micromodel-imbibition}
\end{figure}

\textbf{Purpose.} At a pore junction, the evolving meniscus determines
which downstream throat is invaded and how the displaced phase escapes.
These local decisions govern the displacement pathways represented in the
dataset. Micromodel experiments resolve individual filling events in a
controlled pore geometry, providing a direct test of the interface dynamics
beyond the single-capillary validations above.
\textbf{Setup.} We simulate drainage and spontaneous imbibition in
idealized three-dimensional representations of two four-throat micromodel
junctions, discretized at $3\,\mu\mathrm{m}$ per voxel.
For drainage, a prescribed inlet flux drives the non-wetting phase through
a square pore at $\mathrm{Ca}\approx10^{-4}$ and an imposed wetting-phase
contact angle
$\theta_w=26^{\circ}$. For imbibition, the wetting phase enters a patterned
pore at an imposed $\theta_w=16^{\circ}$ with equal pressure at all four ports.
Throat widths quoted below identify the experimental throats: the drainage
outlet $T_4$ is represented by $108\,\mu\mathrm{m}$, and the imbibition
throats $T_1$, $T_2$, $T_3$, and $T_4$ by $27$, $48$, $57$, and
$105\,\mu\mathrm{m}$, respectively. The simulated channel depths are
$45\,\mu\mathrm{m}$ for drainage and $54\,\mu\mathrm{m}$ for imbibition.
\textbf{Reference.} The single-junction experiments of
\citet{zacharoudiou2017porefilling} resolve both displacement sequences:
air displacing \emph{n}-decane during dynamic drainage, and \emph{n}-decane
displacing air during spontaneous imbibition. Drainage selects the widest
downstream throat, consistent with capillary entry-pressure ordering.
Spontaneous imbibition instead fills an adjacent throat before the
narrowest one, demonstrating the role of pore-body geometry in a filling
sequence that departs from quasi-static predictions.
\textbf{Result.} The solver reproduces the observed throat-filling order
in both cases.
During drainage, the non-wetting phase fills the pore body and exits
through the widest throat, $T_4$ ($107\,\mu\mathrm{m}$), while the two
side throats remain predominantly wetting-filled
(Figure~\ref{fig:micromodel-drainage}). During imbibition, the wetting
phase fills the adjacent throat $T_2$ ($47\,\mu\mathrm{m}$) before
$T_1$ ($27\,\mu\mathrm{m}$), and the remaining non-wetting phase
retreats through $T_4$ (Figure~\ref{fig:micromodel-imbibition}).
Together, these cases support the solver's ability to capture pore-scale
pathway selection under both forced and spontaneous displacement.

\FloatBarrier

\subsection{Displacement regimes in 3D pore geometry}
\label{app:val:lenormand}

\begin{figure}[!htbp]
\begin{center}
\includegraphics[width=0.7\linewidth]{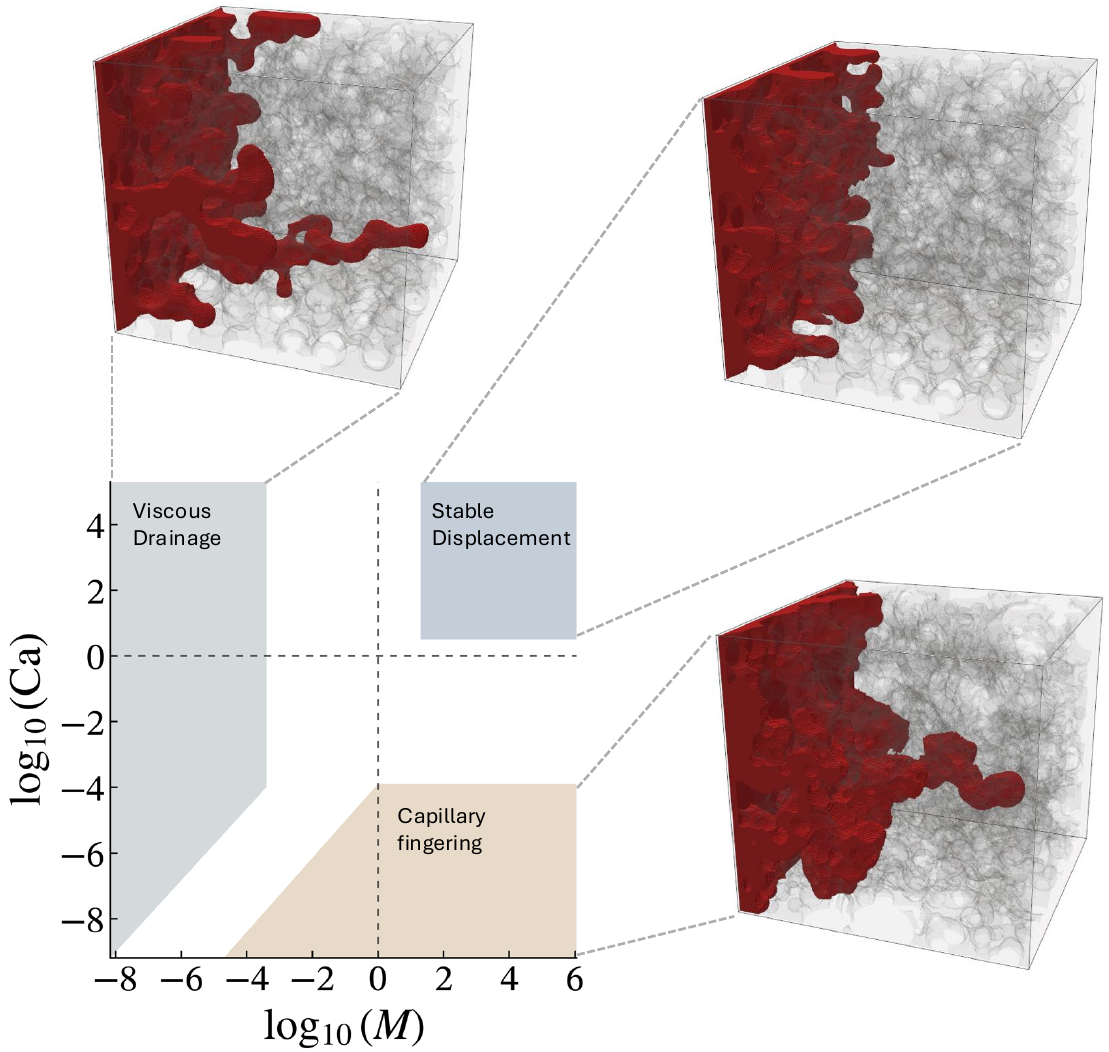}
\end{center}
\caption{Displacement regimes on the $256^3$ sphere pack, varying only
$(\mathrm{Ca}, M)$; the non-wetting phase is red, solid translucent, flow
left to right. The central map sketches the
\citet{lenormand1988numerical} phase diagram in
$(\log_{10} M,\, \log_{10} \mathrm{Ca})$, and each regime region is
linked to a snapshot of the corresponding run: stable displacement at
$(10^{-3},\, 30)$ --- a compact front sweeping the pore space; viscous
fingering at $(10^{-3},\, 0.01)$ --- thin branched fingers that screen
the surrounding pore space; capillary fingering at $(10^{-5},\, 30)$ ---
a rough, capillarity-controlled front.}
\label{fig:val-lenormand}
\end{figure}

\textbf{Purpose.} This test evaluates whether the validated numerical
components jointly reproduce the established dependence of displacement
morphology on $(\mathrm{Ca},M)$ in a real pore geometry.
\textbf{Setup.} Immiscible displacement through a $256^3$ voxelized sphere
pack, moving only the two control knobs of
Table~\ref{tab:mapping}: $(\mathrm{Ca}, M) = (10^{-3}, 30)$,
$(10^{-3}, 0.01)$, and $(10^{-5}, 30)$.
\textbf{Reference.} The classical phase diagram of
\citet{lenormand1988numerical}: these three corners must produce,
respectively, \emph{stable displacement} (compact, near-flat front),
\emph{viscous fingering} (thin branched fingers), and \emph{capillary
fingering} (isotropic invasion-percolation clusters).
\textbf{Result.} Varying only $(\mathrm{Ca},M)$ produces stable displacement,
viscous fingering, and capillary fingering, with the distinct front
morphologies expected from the Lenormand phase diagram
(Figure~\ref{fig:val-lenormand}).

\FloatBarrier

\section{Simulation Setup}
\label{app:setup}

This appendix specifies how the trajectories summarized in
Table~\ref{tab:campaigns} were produced: the solver configuration shared by
all runs, the geometry pipeline, and the domain, boundary conditions, and
termination condition of each campaign. Table~\ref{tab:setup-runs} lists the resulting
run lengths.

\subsection{Common Configuration}
\label{app:setup:common}

All runs use the production stack of Appendix~\ref{app:solver}: D3Q19, MRT
collision with Guo forcing, CSF interfacial tension with $\sigma=0.05$,
recoloring with $\beta=0.99$, the Akai wetting condition, and half-way
bounce-back walls. In the three porous-media campaigns every non-shear moment
relaxes at $0.8\,\omega$, the choice of \citet{leclaire2017generalized},
which improves stability; underfill uses the fixed
rates of Table~\ref{tab:params} (Appendix~\ref{app:solver:numerics}).
Both fluids have unit density and no body force acts, so buoyancy plays no
role and the flow is set by capillarity, viscosity, and the boundary drive.
All quantities are in lattice units.

\textbf{Viscosity ratio.} In the three porous-media campaigns the viscosity
ratio $M=\nu_{\mathrm{nw}}/\nu_{\mathrm{w}}$ is imposed by splitting a fixed
reference viscosity $\nu_0=0.04$ geometrically,
$\nu_{\mathrm{nw}}=\nu_0\sqrt{M}$ and $\nu_{\mathrm{w}}=\nu_0/\sqrt{M}$, so
changing $M$ moves both fluids while their geometric mean stays put; at
$M=0.05$ this gives $\nu_{\mathrm{nw}}=0.0089$ and $\nu_{\mathrm{w}}=0.179$.
Underfill instead fixes the displaced air at $\nu=0.025$ and sets the
encapsulant to $0.025\,M$.

\textbf{Capillary number.} The injection velocity follows from
$\mathrm{Ca}=\nu_0 u_{\mathrm{in}}/\sigma=10^{-5}$ (Table~\ref{tab:mapping}),
i.e.\ $u_{\mathrm{in}}=1.25\times10^{-5}$ lattice units per step in every
displacement, independent of $M$. Because $\mathrm{Ca}$ is referenced to
$\nu_0$, the capillary number of the invading fluid alone is
$10^{-5}\sqrt{M}$ in drainage and GDL intrusion and $10^{-5}/\sqrt{M}$ in the
trapping flood.

\textbf{Domain conventions.} Arrays are stored as $(z,y,x)$ with $x$ the flow
axis, inlet at $x=0$. The four lateral faces are sealed no-slip walls that
carry the sample's contact angle. Along $x$ the sample is extended by an inlet
reservoir and an outlet buffer (and, for drainage, a porous plate); the
benchmark scores the sample span only (Section~\ref{sec:exp:setup}). The
constant-rate drive is the moving-piston inlet of
Appendix~\ref{app:solver:bc}: the $x=0$ plane is sealed and the first fluid
slab receives exactly the volume $u_{\mathrm{in}}A$ per step, with the inlet
pressure left floating.

\textbf{Stored data.} A frame holds the phase field $\phi$, density, pressure,
velocity magnitude, and the velocity vector on the full domain, in single
precision, with $\phi$ undefined inside the solid; the solid mask, the
geometry record, and the run metadata (parameters, regions, termination reason)
are stored beside the trajectory. Frames are written every 5{,}000 steps,
every 1{,}000 for underfill, including the initial state.

\subsection{Geometries}
\label{app:setup:geometry}

\textbf{Scanned.} Rock samples are random-origin cubic crops, with random axis
permutations and flips, of segmented $1000^3$ $\mu$CT volumes of Bentheimer,
CastleGate, and Buff Berea sandstone at 2.25\,$\mu$m per voxel
\citep{neumann2021high,neumann2020sandstones}. GDL samples are random-origin crops of
carbon-paper scans at 5\%, 20\%, and 40\% PTFE loading and 2.05\,$\mu$m per
voxel \citep{shojaei2022minimal,shojaei2023xray}, oriented so that the through-plane direction
is the flow axis.

\textbf{Generated.} Sphere media are overlapping monodisperse spheres (radius
14--28 voxels, target porosity 0.20--0.35); blob media threshold a Gaussian
random field (target porosity 0.22--0.32); polydisperse media overlap spheres
with log-normally distributed radii (median 8--14 voxels at $128^3$, porosity
0.25--0.32); fiber mats are in-plane cylinders of radius 3--5 voxels with an
out-of-plane tilt of at most $20^\circ$ (target porosity 0.65--0.80). After
post-processing, the samples used span porosities of 0.17--0.42 for the
scanned sandstones, 0.19--0.38 for the generated rock families, 0.43--0.68 for
the scanned GDLs, and 0.53--0.77 for the fiber mats.

\textbf{Post-processing and screening.} Every rock and fiber sample is opened
morphologically with a 3-voxel ball, which removes throats too narrow for
the solver to resolve \citep{li2020discretization}, and pore pockets disconnected
from the faces are filled. A sample enters a campaign only if its pore space
percolates along the flow axis with the lateral faces sealed. Drainage and
trapping additionally require a critical throat radius above five voxels and
at least 1.3 times the smallest radius the drainage pressure limit can invade
at the run's contact angle, so that the termination condition below cannot
fire before the sample is entered.

\textbf{Flip-chip gaps.} An underfill geometry is analytic: two solid plates
(substrate and chip) separated by the gap height, joined by a staggered array
of truncated-sphere solder balls, with a ball-free margin of one diameter at
the entry and the vent. Across the 32 geometries the ball diameter is 32--54
voxels, the gap height 20--46, the pitch 1.55--2.09 diameters, and the array
holds 8--46 balls, for a gap porosity of 0.82--0.93. The domain is
$(h+2)\times482\times(432\text{--}476)$ voxels for gap height $h$. One voxel
corresponds to 2.5\,$\mu$m if a 32-voxel ball is read as an 80\,$\mu$m bump.

\subsection{Primary Drainage}
\label{app:setup:drainage}

\textbf{Domain.} Along the flow axis: a 7-slab inlet reservoir, the rock
($128$ or $256$ voxels), a 20-voxel porous plate, and a 6-slab outlet
buffer, i.e.\ $128{\times}128{\times}161$ and $256{\times}256{\times}289$
voxels in total. The plate is a solid slab pierced by square $5\times5$-voxel
pores on an 8-voxel pitch and is strongly wetting
($\theta_{\mathrm{plate}}=150^\circ$). It plays the role of the porous plate
of a laboratory drainage experiment: the wetting phase leaves through it
freely, while the non-wetting phase cannot enter until the pressure drop
reaches its entry pressure.

\textbf{Protocol.} The reservoir starts filled with non-wetting fluid and the
rest of the domain with wetting fluid. The piston injects non-wetting fluid at
$\mathrm{Ca}=10^{-5}$; the outlet is a Zou--He pressure boundary at unit
density that admits wetting fluid only. A run ends when the pressure drop
across the domain reaches half the plate entry pressure or when 0.75 rock
pore volumes have been injected, whichever comes first. Breakthrough is
recorded but does not end a run, so trajectories
continue into the late, capillary-controlled filling of smaller pores.

\subsection{Capillary Trapping}
\label{app:setup:trapping}

\textbf{Domain.} A 10-slab inlet reservoir, the rock, and a 6-slab outlet
buffer, with no plate: $128{\times}128{\times}144$ and
$256{\times}256{\times}272$ voxels.

\textbf{Initial state.} No drainage is simulated first. The non-wetting phase
is placed with the pore-morphology-based initializer of
\citet{tavakkoli2025pore}, which mimics the end state of a drainage: the
non-wetting phase fills the wide pore bodies and the wetting phase stays in
corners and narrow throats. Every run starts at a non-wetting saturation of
about 60\% (median 0.60); the exact value follows from the pore geometry of
the sample, so it serves as a randomized starting point. The buffers are
filled with wetting fluid.

\textbf{Stage 1: equilibration.} To make the initial state physically
accurate as well, the run first relaxes toward capillary equilibrium with the
piston effectively off ($\mathrm{Ca}=10^{-12}$) and an open outflow outlet. The
pressure field adjusts to the Young--Laplace jump and the sample imbibes
spontaneously until the wetting saturation changes by less than
$5\times10^{-6}$ per 1{,}000 steps for three consecutive blocks, or for at
most 200{,}000 steps.

\textbf{Stage 2: flood.} The piston then injects wetting fluid at
$\mathrm{Ca}=10^{-5}$ through the same boundaries; non-wetting fluid reaching
the outlet buffer is removed and counted as produced. Saturation is a poor
termination signal, since ganglia can move at constant saturation. The run
therefore ends when the non-wetting body itself stops moving: fewer than
0.3\% of its voxels change over a 20{,}000-step window, in three consecutive
windows, with a cap of 1.2 rock pore volumes injected. Equilibration lowers
the median non-wetting saturation from 0.60 to 0.51, and the flood leaves a
median residual of 0.39.

\subsection{GDL Water Intrusion}
\label{app:setup:gdl}

The layout is that of drainage without the plate: a 7-slab reservoir, the
layer (64 or 128 voxels through-plane), and a 6-slab buffer, i.e.\
$128{\times}128{\times}77$ and $256{\times}256{\times}141$ voxels. Liquid
water, the non-wetting phase on the hydrophobic fibers, is injected
through-plane by the piston at $\mathrm{Ca}=10^{-5}$ against the same
wetting-only Zou--He pressure outlet. Because a thin layer breaks through
early, the run continues to 1.2 times the breakthrough step or to 0.5 pore
volumes injected, whichever comes first, which keeps the post-breakthrough
redistribution in the trajectory.

\subsection{Flip-Chip Underfill}
\label{app:setup:underfill}

The gap is open at both ends of the flow axis and sealed laterally; the
encapsulant is the wetting phase and enters from a 22-slab strip at the
inlet. Inlet and outlet are Zou--He pressure boundaries held at the same
density, so no pressure difference is applied and the front advances by
capillary suction alone, as in capillary underfill dispensing. A single
contact angle applies to the plates and the balls. A run ends ten blocks
after the fluid fills 99\% of the ball array. The viscosity ratios
$M=5$--$30$ are far below that of a real encapsulant--air pair
(${\sim}10^{3}$--$10^{4}$); they preserve a more viscous invading fluid while
remaining within the solver's stable range, and should be read as a
qualitative regime rather than a material match.

\subsection{Parameter Sampling}
\label{app:setup:sampling}

Each small-domain campaign crosses four viscosity ratios with four contact
angles (Table~\ref{tab:campaigns}). The 80 generated runs and the 80 scanned
runs each cover this $4\times4$ grid five times, with the geometry families
balanced within each group (28/26/26 runs). The large-domain campaigns cross the four
viscosity ratios with $\theta\in\{120^\circ,150^\circ\}$ once per geometry
source (8 + 8 runs) and run on four GPUs with the domain sharded along $z$.
Underfill assigns the 12 $(M,\theta)$ pairs of Table~\ref{tab:campaigns} to 32
distinct geometries, two or three per pair and eight per viscosity ratio.

\begin{table}[!htbp]
\caption{Run lengths per campaign and domain size. Domains include the
reservoir, plate, and buffer slabs; frames include the initial state.
Frames per run are medians.}
\label{tab:setup-runs}
\centering
\small
\begin{tabular}{@{}llrrrr@{}}
\toprule
Campaign & Domain (voxels) & Runs & Frames per run & Frame interval & Frames \\
\midrule
Drainage  & $128{\times}128{\times}161$ & 160 & 376 & 5{,}000 & 58{,}830 \\
          & $256{\times}256{\times}289$ &  16 & 694 & 5{,}000 & 10{,}643 \\
Trapping  & $128{\times}128{\times}144$ & 160 & 105 & 5{,}000 & 19{,}424 \\
          & $256{\times}256{\times}272$ &  16 & 239 & 5{,}000 &  3{,}616 \\
GDL       & $128{\times}128{\times}77$  & 160 & 292 & 5{,}000 & 45{,}437 \\
          & $256{\times}256{\times}141$ &  16 & 486 & 5{,}000 &  8{,}217 \\
Underfill & $(22\text{--}48){\times}482{\times}(432\text{--}476)$
          &  32 & 342 & 1{,}000 & 12{,}379 \\
\midrule
Total     &                             & 560 &                 &         & 158{,}546 \\
\bottomrule
\end{tabular}
\end{table}
\FloatBarrier

\section{Input Geometry Examples}
\label{app:geo}

\begin{figure}[!htbp]
  \centering
  \includegraphics[width=\linewidth]{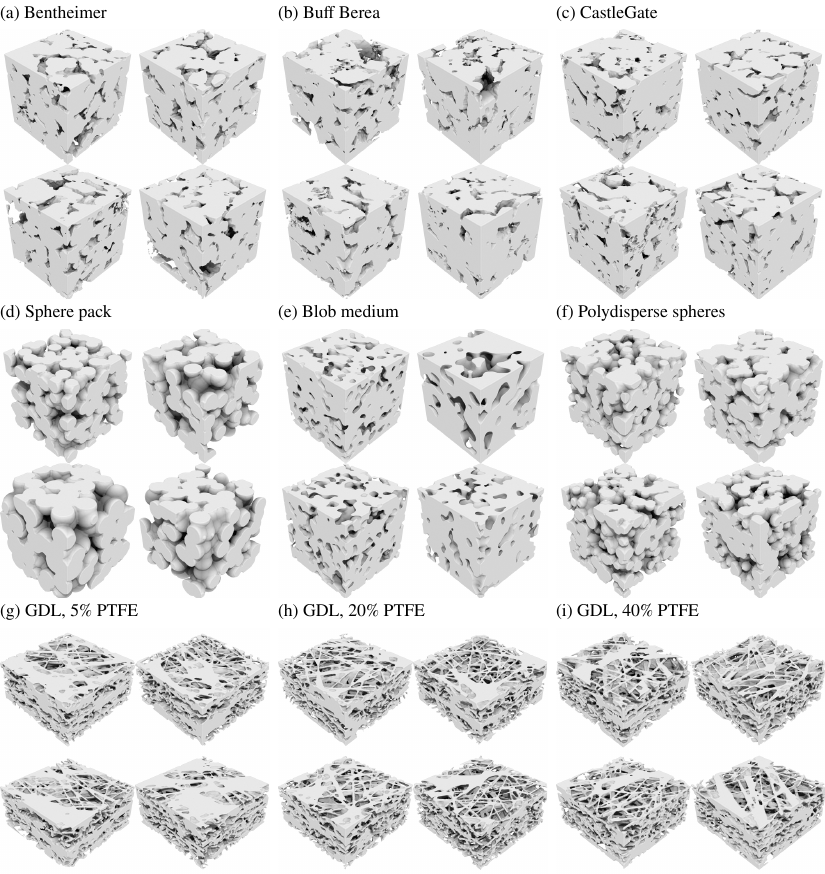}
  \caption{\textbf{Input geometry examples at the large domain size.} Each
  panel renders the solid phase of four samples of one family.
  \textbf{(a)--(c)}~Sandstones from $\mu$CT scans, $256^3$ voxels: Bentheimer,
  Buff Berea, and CastleGate. \textbf{(d)--(f)}~Generated rock-like media,
  $256^3$ voxels: sphere packs, blob media, and polydisperse overlapping
  spheres. \textbf{(g)--(i)}~Carbon-paper GDLs from $\mu$CT scans at 5\%,
  20\%, and 40\% PTFE loading, $256{\times}256{\times}128$ voxels, with the
  through-plane flow direction vertical. Samples of one family share a
  morphology but differ in porosity and connectivity.}
  \label{fig:geo-gallery}
\end{figure}

Figure~\ref{fig:geo} shows one large sample of each geometry family beside
two small ones. Figure~\ref{fig:geo-gallery} adds further large-domain
samples of the scanned sandstones, the generated rock-like families, and the
scanned GDLs; Appendix~\ref{app:setup:geometry} describes how each family is
produced and screened. The full geometry collection, every family at both
domain sizes together with the flip-chip gaps, is available at
\url{https://huggingface.co/datasets/PoreML/PoreML_geo}.

\FloatBarrier

\section{Baseline Models}
\label{app:baselines}

All baselines share the prediction contract of
Section~\ref{sec:exp:setup}. A run provides a solid mask $s$, the
viscosity ratio $M$, the contact angle $\theta$, and saved frames
$\mathbf{q}_t=(\phi_t,\mathbf{u}_t,p_t)$ with five scalar channels per
voxel, normalized by fixed per-field offsets and scales and set to a
constant fill inside the solid. A network $g_\psi$ with parameters
$\psi$ maps the input
$\mathbf{x}_t=(s,\log_{10}M,\theta/180^{\circ},\mathbf{q}_t)$ to the next
frame in residual form,
\begin{equation}
  \hat{\mathbf{q}}_{t+1}=\mathbf{q}_t+g_\psi(\mathbf{x}_t),
  \label{eq:baseline-residual}
\end{equation}
with a zero-initialized output layer. The two run conditions enter as
constant input channels for every model. Voxel models consume the dense grid
$\mathbf{x}_t\in\R^{C\times D\times H\times W}$. Point models consume the
set of pore voxels $\{(\boldsymbol{\xi}_i,\mathbf{x}_i)\}_{i=1}^{N}$ with
positions $\boldsymbol{\xi}_i\in(0,1)^3$ and, in place of the mask, a
binary wall flag and the local solid fraction in cubic neighborhoods of
radius $r$ voxels, so that the wall is seen rather than implied by the
point set; their predictions are scattered back to the grid before
scoring, so all models are evaluated on the same voxels with the same
metrics. Capacities are matched near $10^7$ parameters (UNet3D 11.2M,
FNO3D 9.7M, P3D 10.7M, AB-UPT 10.5M); Transolver++ keeps its published
1.7M setting because its cost grows faster than its parameter count.
The four published architectures are ports of the authors' public code,
verified against the original implementations by copying weights and
comparing outputs. Every model trains
for the same number of run passes on the porous-media campaigns (ten
epochs over the 128-run split, twenty over the 64-run generated split)
and for twenty epochs on underfill, with the masked mean-squared error
over pore voxels unless stated otherwise.
Table~\ref{tab:baseline-config} summarizes the configurations.
Every model is then fine-tuned with the pushforward trick
\citep{brandstetter2022message}, following the training loop of
\citet{hassan2023bubbleml}. A pushed batch first advances its input one step on the
model's own prediction without gradients and is then scored against the
true frame two steps ahead; the probability of pushing a batch rises
linearly from 0.5 to 1.0 over the run, and the other batches receive
Gaussian noise of standard deviation 0.01 on the pore voxels of their
flow channels, with the solid mask and run conditions left exact.

\begin{table}[!htbp]
\caption{Baseline configurations, shared by all campaigns. Parameter
counts are those recorded for the drainage models; the underfill
variants of UNet3D and FNO3D are given in the text. $B$: batch size.}
\label{tab:baseline-config}
\centering
\begin{adjustbox}{max width=\textwidth}
\begin{tabular}{@{}llrlll@{}}
\toprule
Model & Input & Params & Architecture & Optimizer and schedule & Loss \\
\midrule
UNet3D & voxels & 11.2M & $c_0=20$, $L=4$, 4 norm groups & AdamW $3{\times}10^{-4}$, constant, $B=1$ & MSE \\
FNO3D & voxels & 9.7M & $k_{\max}=24$, width 18, $T=4$, pad $1/16$ & AdamW $3{\times}10^{-4}$, wd $10^{-4}$, halved every fifth of the run, $B=1$ & rel.\ $H^1$ \\
P3D & voxels & 10.7M & preset S: width 64, 4 heads, $w=4$, depth $(2,2,2,2,2)$ & AdamW $2{\times}10^{-4}$, one halving, clip 1, $B=1$ & MSE \\
Transolver++ & points & 1.7M & width 256, 4 blocks, 8 heads, $G=32$, dropout 0.1 & AdamW $10^{-3}$ one-cycle, clip 1, $B=1$ & MSE \\
AB-UPT & points & 10.5M & width 192, 3 heads, $S=A=16{,}384$, $8+8$ anchor blocks & AdamW $3{\times}10^{-4}$ one-cycle, wd 0.05, clip 1, $B=1$, bf16 & MSE \\
\bottomrule
\end{tabular}
\end{adjustbox}
\end{table}

\subsection{UNet3D}
\label{app:baselines:unet}

The 3D U-Net \citep{cicek2016unet} is the dense-grid convolutional
baseline: an encoder--decoder whose encoder halves the resolution while
doubling the channel width at every level, and whose decoder restores
the resolution while concatenating the encoder features of the same
level through skip connections, so that fine local structure and coarse
context meet at every scale. With widths $c_\ell=c_0\,2^{\ell}$ at levels
$\ell=0,\dots,L$, each level applies two $3^3$ convolutions, each
followed by group normalization and a SiLU activation; levels are joined
by stride-2 convolutions on the way down and stride-2 transposed
convolutions on the way up, and a $1^3$ convolution maps the
full-resolution features to the five output channels. Every operation is
local, a bias that mirrors the streaming--collision update of the solver,
while the bottleneck sees the whole domain. In our configuration,
$c_0=20$ and $L=4$, so a $128^3$ domain is compressed to $8^3$ at the
bottleneck with widths from 20 to 320, and group normalization uses four
groups. The underfill gap is only a few tens of voxels thick, so its
model uses $c_0=16$ and $L=1$, with axes that do not divide
replicate-padded and cropped after. The model trains with AdamW at a
constant learning rate of $3\times10^{-4}$ and batch size 1.

\subsection{FNO3D}
\label{app:baselines:fno}

The Fourier neural operator \citep{li2021fourier,kovachki2023neural}
learns a map between function spaces by parameterizing the kernel of an
integral operator in Fourier space. After a pointwise lifting
$v_0=P(\mathbf{x}_t)$, each of $T$ layers updates the latent field as
\begin{equation}
  v_{\ell+1}(\mathbf{r})=\sigma\!\left(W_\ell\,v_\ell(\mathbf{r})
  +\mathcal{F}^{-1}\!\big(R_\ell\cdot\mathcal{F}v_\ell\big)(\mathbf{r})\right),
  \label{eq:fno}
\end{equation}
where $\mathcal{F}$ is the three-dimensional Fourier transform over the
spatial axes, $R_\ell$ is a complex weight tensor acting on the
$k_{\max}$ lowest frequencies of each axis and discarding the rest, and
$W_\ell$ is a pointwise linear map; a projection $Q$ returns to the
output channels. The weights index frequencies rather than voxels, so the
same weights apply unchanged to a grid of a different size. We follow the
reference library rather than the 2021 paper: each layer adds a
channel-wise MLP with a gated skip and GELU activations, $P$ and $Q$ are
two-layer, and three coordinate channels are appended to the input.
Because the rock is periodic on no axis, each axis is zero-padded by one
sixteenth on both sides before the transforms and cropped afterwards.
The three transform axes are spatial; time stepping is left to the
autoregressive contract rather than to the space--time variant of the
original paper. In our configuration, $k_{\max}=24$ modes per axis,
$T=4$ layers and a latent width of 18 channels, which meets the
parameter budget of the other voxel models after wider variants overfit
the correlated training windows. The underfill model keeps 11 modes
along the gap axis and 16 in-plane with a width of 16. The model trains
with AdamW at $3\times10^{-4}$, weight decay $10^{-4}$, the learning
rate halved at every fifth of the run, and the relative $H^1$ loss of the
reference library, which penalizes errors in the finite-difference
gradients of the fields alongside their values.

\subsection{P3D}
\label{app:baselines:p3d}

P3D \citep{holzschuh2026p3d} is a hybrid convolutional and transformer
surrogate designed for high-resolution three-dimensional fields. A
convolutional stem compresses the volume by a factor of four in each
direction, a U-shaped transformer backbone operates on the compressed
volume, and a convolutional decoder with additive skips restores full
resolution. Convolutions carry local structure and attention carries
long-range interaction, but attention is only ever evaluated inside
non-overlapping windows of $w^3$ cells,
\begin{equation}
  \mathrm{Attn}(Q,K,V)=\mathrm{softmax}\!\left(
  \lambda\,\hat{Q}\hat{K}^{\top}+B\right)V,
  \label{eq:p3d-attention}
\end{equation}
with $\ell_2$-normalized queries and keys $\hat{Q},\hat{K}$, a learned
logit scale $\lambda$ and a relative-position bias $B$ produced by a
small network on log-spaced offsets, which keeps the cost linear in the
number of cells. Each block is modulated, in the manner of diffusion
transformers, by shift, scale and gate vectors derived from a
conditioning embedding. The original work also provides a probabilistic
flow-matching variant and a patch-fusion scheme that assembles $512^3$
solutions from sub-volumes; we use neither, because the benchmark asks
for deterministic one-step surrogates and a $128^3$ domain fits in one
region. The conditioning embedding is held constant, since $M$ and
$\theta$ already enter as input channels. In our configuration, we use
the published S preset: latent width 64 with four heads, stem widths
$(32,32,64)$, two blocks per stage over two encoder stages, one latent
stage and two decoder stages, and $w=4$ without window shifting, as in
the released presets. A $128^3$ domain is thus attended at $32^3$ and
reduced to $8^3$ in the latent stage; the underfill domain is
replicate-padded to a multiple of 16 and cropped back. The model trains
with AdamW at $2\times10^{-4}$ without weight decay, gradient clipping
at 1, and one halving of the learning rate at two thirds of the run.

\subsection{Transolver++}
\label{app:baselines:transolver}

Transolver++ \citep{luo2025transolver} extends the physics-attention of
Transolver \citep{wu2024transolver}: instead of attending among mesh
points, each layer assigns the $N$ points to a small number $G$ of
latent physical states, attends among those states, and broadcasts the
result back, so that the cost is linear in $N$ and the model applies to
arbitrary point sets. For a head with point tokens $\mathbf{h}_i\in\R^{d}$,
the assignment weights are a softmax over learned logits with a learned
per-point temperature $\tau_i$, perturbed by Gumbel noise $\gamma$
during training,
\begin{equation}
  w_{ig}=\frac{\exp\big((\mathbf{h}_i^{\top}\mathbf{a}_g+\gamma_{ig})/\tau_i\big)}
  {\sum_{g'}\exp\big((\mathbf{h}_i^{\top}\mathbf{a}_{g'}+\gamma_{ig'})/\tau_i\big)},
  \qquad
  \mathbf{s}_g=\frac{\sum_i w_{ig}\,\mathbf{h}_i}{\sum_i w_{ig}},
  \label{eq:transolver-slices}
\end{equation}
which the authors call eidetic states: the temperature lets a point
commit to one state where the physics is sharp and spread over several
where it is smooth. Standard attention runs among the $G$ state tokens,
$\mathbf{s}'=\mathrm{Attn}(\mathbf{s}W_Q,\mathbf{s}W_K,\mathbf{s}W_V)$,
and every point recovers $\mathbf{h}'_i=\sum_g w_{ig}\,\mathbf{s}'_g$.
Blocks alternate this attention with a feed-forward network under
pre-layer normalization. In our configuration, every pore voxel of a
sample is a point, with the coordinates concatenated to its features,
and the local solid fraction is supplied at radii $r\in\{1,2,4,8\}$. The
reference-grid position encoding and the global condition embedding of
the original are not used, because the points already lie on a regular
grid and $M$ and $\theta$ enter as per-point features. Evaluation uses
the noise-free assignment so that predictions are deterministic. We keep
the published aircraft setting of width 256, four blocks, eight heads,
$G=32$ states per head, feed-forward ratio 2 and dropout 0.1, and train
with AdamW without weight decay at a one-cycle peak learning rate of
$10^{-3}$, batch size 1 and gradient clipping at 1.

\subsection{AB-UPT}
\label{app:baselines:abupt}

The anchored-branched universal physics transformer
\citep{alkin2025abupt} scales transformer neural operators to point
clouds of millions of points by separating three roles. A geometry
branch pools the boundary points into $S$ supernodes by radius-graph
message passing and encodes them with transformer blocks; a surface
branch and a volume branch hold the points at which fields are
predicted; and within each branch a random subset of $A$ anchor points
carries full self-attention while every remaining point, a query, only
cross-attends to the anchors, so that the quadratic cost is confined to
$A^2$ and the rest is linear in $N$. Position enters through a
sinusoidal embedding of the coordinates and rotary position embeddings
inside every attention. A shared trunk applies perceiver cross-attention
from the tokens to the geometry and then alternates attention within a
branch with attention across the two branches; branch-specific
anchor-attention blocks and linear decoders produce the outputs. In our
configuration, the surface branch holds the wall-adjacent pore voxels
and the volume branch the remaining pore voxels, both predicting the same
five fields; the wall points also form the geometry, pooled into
$S=16{,}384$ supernodes within a radius of three voxels and encoded by
one block, and the local solid fraction is supplied at radii
$r\in\{1,2,4\}$. Because the original is a steady-state model driven by
positions alone, we add a linear embedding of the per-point features to
the position embedding of every token, which is how the current frame
and the conditions enter. We use the published width of 192 with three
heads, the published trunk pattern, eight anchor blocks per branch and
$A=16{,}384$ anchors per branch. Training follows the authors'
anchor-centric recipe: each window is a random subset of 32{,}768 pore
voxels, so that nearly every training point is an anchor, with AdamW at
a one-cycle peak of $3\times10^{-4}$, weight decay 0.05, batch size 1,
gradient clipping at 1 and bf16 autocast. Evaluation and rollout use
every pore voxel, with anchors drawn from a fixed seed so that results
are reproducible.

\section{Failure Mode Analysis}
\label{app:failure}

Section~\ref{sec:exp:shift} shows that a rollout can retain a low field
error while losing the topology of the fluid, most severely in the GDL
campaign. We attribute this to two mechanisms, specks and tunnels, which
we define, quantify over every GDL validation window, and illustrate on
one representative window.

\textbf{Mechanisms.} A speck is a small island of the invading phase
within the other fluid, or a small bubble of the other fluid within the
invading phase, produced where the predicted phase field oscillates about
the threshold; it is discretization noise rather than transported fluid. A
tunnel is a handle of the invading phase: a loop it closes around a solid
fiber or a pocket of the other fluid, as a torus encloses its hole. GDL
geometries favor tunnels. Their fibers are thin, and the reference retains
only a thin wetting film between a fiber and the invading water, so a
prediction that fills the film encircles the fiber, and one whose phase
fragments into thin strands encloses pockets of the other fluid.

\begin{figure}[!htbp]
  \centering
  \includegraphics[width=0.8\linewidth]{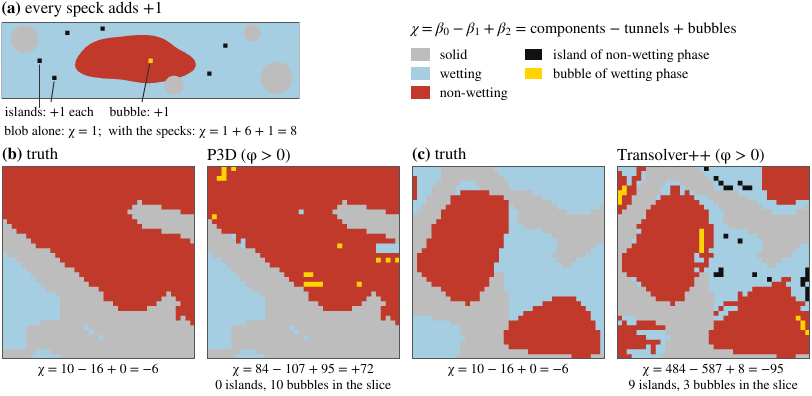}
  \caption{\textbf{Specks corrupt the Euler number.}
  \textbf{(a)}~$\chi=\beta_0-\beta_1+\beta_2$ counts pieces, tunnels, and
  enclosed bubbles of the non-wetting phase, so every island of non-wetting
  phase and every bubble of wetting phase adds $+1$, however small: the
  blob alone has $\chi=1$, with the specks $\chi=8$. \textbf{(b, c)}~One
  $40{\times}40$-voxel crop of a slice across the flow, fiber GDL, rollout
  step 64: the reference and the thresholded prediction ($\phi>0$) of P3D
  and of Transolver++, with speck islands (black) and speck bubbles
  (yellow) marked. Under each panel, $\chi$ of the whole 3D frame with its
  three terms. The predictions' 84 and 484 pieces and 95 and 8 bubbles are
  almost all specks: P3D's $\chi=+72$ has the wrong sign against the
  reference's $-6$, and Transolver++'s $-95$ hides 587 tunnels behind 484
  islands.}
  \label{fig:app-euler-speckles}
\end{figure}

\begin{figure}[!htbp]
  \centering
  \includegraphics[width=0.8\linewidth]{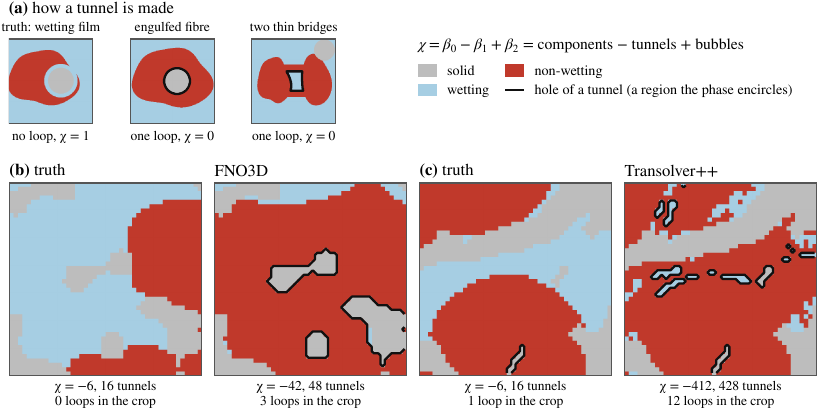}
  \caption{\textbf{Tunnels corrupt the Euler number and cannot be filtered.}
  \textbf{(a)}~A loop of the non-wetting phase is a tunnel through it and
  lowers $\chi$ by one; being part of the phase, no size filter removes it.
  Where the reference keeps a wetting film between a fiber and the phase
  there is no loop ($\chi=1$); a prediction that engulfs the fiber, or
  joins two regions by two voxel-thin bridges, closes one ($\chi=0$).
  \textbf{(b)}~Reference and
  FNO3D in a plane along the flow, where the fibers are cut across
  ($44{\times}44$ voxels): FNO3D encircles three fiber cross-sections that
  in the reference touch wetting fluid. \textbf{(c)}~Reference and
  Transolver++ in a plane across the flow: twelve pockets of wetting fluid
  enclosed by thin strands of phase. A region the phase encircles in the
  plane, a fiber cross-section or a pocket of wetting fluid, is the hole of
  a tunnel and is outlined in black. Under each panel, $\chi$ and the
  tunnel count of the whole frame, and the loop count of the crop.}
  \label{fig:app-euler-tunnels}
\end{figure}

\textbf{Topological accounting.} Both mechanisms enter the Euler
characteristic through its Betti-number decomposition
$\chi=\beta_0-\beta_1+\beta_2$, where $\beta_0$ is the number of connected
pieces of the phase, $\beta_1$ the number of tunnels through it, and
$\beta_2$ the number of bubbles of the other fluid it encloses; a solid
ball has $\chi=1$, a torus $0$, and a ball with an internal cavity $2$.
Each speck, island or bubble, therefore adds $+1$ to $\chi$, and each
tunnel subtracts $1$ (Figures~\ref{fig:app-euler-speckles}a
and~\ref{fig:app-euler-tunnels}a).

\textbf{Contribution to the error.} Over all GDL validation windows at
the end of the rollout, with islands and bubbles under 30 voxels counted as specks, specks
account for roughly a third to two thirds of the Euler discrepancy,
depending on the model, and for P3D, Transolver++, and AB-UPT they reverse
the sign of the error in most windows, so that the raw $\chi$ reports too
few tunnels where there are too many. The rest is tunnels: every model
predicts more tunnels than the reference in 82--98\% of the windows. For
P3D, Transolver++, and AB-UPT the excess consists mostly of voxel-thin
bridges, which a one-voxel morphological opening removes; for FNO3D and
UNet3D it consists mostly of engulfed fibers, which survive it. The
picture is the same under Geo-Shift, except that Transolver++'s error
there is mostly specks. Neither mechanism displaces many voxels, so the
field error is nearly insensitive to both: in Table~\ref{tab:all-results},
AB-UPT attains the lowest step-64 $\phi$ MAE in GDL with an $M_3$ error
almost five times UNet3D's. Since that $M_3$ counts $\chi$ without
filtering, both mechanisms enter it.

\textbf{Case study.} Figures~\ref{fig:app-euler-speckles}
and~\ref{fig:app-euler-tunnels} illustrate both mechanisms on one GDL
validation window at the last rollout step, a generated fiber geometry scored with the
frames behind Table~\ref{tab:all-results}, whose reference has ten pieces,
sixteen tunnels, and $\chi=-6$. In the $40{\times}40$-voxel crop with the
most speck voxels, P3D leaves bubbles of wetting fluid within the water and
Transolver++ islands of water within the wetting fluid; over the whole
frame, P3D's 84 pieces and 95 bubbles are almost all specks and its
$\chi=+72$ has the wrong sign, while Transolver++'s 484 islands conceal 587
tunnels behind $\chi=-95$. In a plane along the flow, which cuts the fibers
transversely, FNO3D encircles three fiber cross-sections that the reference
separates from the water by wetting fluid, each an engulfed fiber; in a
plane across the flow, Transolver++'s strands enclose twelve pockets of
wetting fluid. Over the whole frame, FNO3D and Transolver++ have 48 and 428
tunnels against the reference's 16.

\FloatBarrier

\section{Limitation and Future Work}
\label{app:future}

\textbf{Physics consistency beyond geometry.} PoreML checks physical
consistency through the geometry and topology of the fluid configuration:
the Minkowski functionals, whose role as state variables of two-fluid flow
in porous media is established
\citep{armstrong2016beyond,mcclure2018geometric,armstrong2019minkowski},
and the mean curvature that the Young--Laplace equation ties to capillary
pressure \citep{blunt2017multiphase}. The conservation laws themselves,
mass and momentum balance, are not scored here. Enforcing or measuring
them in learned surrogates is a broad field of its own, from residual
losses \citep{raissi2019physics,karniadakis2021physics} to architectures
that satisfy a constraint by construction
\citep{beucler2021enforcing,richterpowell2022neural}, and the limits of
residual losses for nonlinear two-phase transport in porous media are
documented \citep{fuks2020limitations}. Whether a surrogate that keeps the
topology of Appendix~\ref{app:failure} also conserves mass along a rollout,
and which constraint helps it do so, is a natural next study on this
dataset.

\textbf{Methodology improvements.} PoreML is a framework and benchmark,
so its effort goes into the data, the protocol, and the metrics rather than
into new methodology: the baselines are published architectures with their
published recipes, each trained once from a single seed
(Appendix~\ref{app:baselines}). Tuning, seed variation, and training
strategies for long rollouts should all raise their scores. The failure
mode of Appendix~\ref{app:failure}, in which voxel-level errors close loops
and fill films that a field loss barely penalizes, points to where new
architectures and objectives can act: representations and losses that see
the connectivity of the phase, not only its voxels. The framework makes
such work directly comparable, since a new model plugs into the same
splits, protocol, and metrics.

\section{Accessibility and Maintenance}
\label{app:access}

\subsection{Resources and Links}
\label{app:access:links}

PoreML is released as five public resources that together cover the path
from a pore geometry to a scored model.

\textbf{Solver.} The lattice Boltzmann solver is available at
\url{https://github.com/PoreML/bob}. The repository contains the test suite,
the validation cases of Appendix~\ref{app:validation}, and the campaign
drivers that generated every run of the dataset, so each trajectory can be
regenerated from its geometry and recorded parameters.

\textbf{Dataset.} All 560 trajectories are hosted on Hugging Face at
\url{https://huggingface.co/datasets/PoreML/PoreML_data}. The repository is
organized by scenario, geometry family, and domain size, so a single run, a
family, or a whole scenario can be downloaded without fetching the full
3.3\,TB. Future versions of the dataset will be published at the same
address.

\textbf{Input geometries.} The solid masks used as simulation input, together
with further samples of every family at both domain sizes, are hosted at
\url{https://huggingface.co/datasets/PoreML/PoreML_geo}
(Appendix~\ref{app:geo}). Each sample is accompanied by its generation
parameters and pore-network statistics, and the collection can be used on its
own to set up new simulations.

\textbf{Machine learning framework.} Code for training, inference, and
evaluation of all benchmark models is available at
\url{https://github.com/PoreML/poreml}. The repository holds the
configuration file of every benchmark run and the frozen data splits,
including those of the Geo-Shift and Scale-Up challenges, and a download
utility that fetches exactly the runs a given split names. The trained
weights are provided as well, at
\url{https://huggingface.co/PoreML/PoreML_checkpoint}: for each of the five
models and every training split, the checkpoint trained frame to frame and
its pushforward fine-tuned counterpart, which is the one behind the reported
results, each with its configuration and training log.

\textbf{Documentation and tutorial.} The dataset cards on Hugging Face
describe the directory layout, the stored fields, the metadata attached to
every run, and common pitfalls when reading the files. A demo notebook,
\url{https://github.com/PoreML/poreml/blob/main/demo/demo.ipynb}, walks
through the minimal workflow on a small subset of the data: it downloads two
drainage trajectories, trains a small UNet for next-step prediction,
fine-tunes it with the pushforward strategy, and evaluates and visualizes a
64-step autoregressive rollout on a geometry unseen during training.

\subsection{Maintenance and Long-Term Preservation}
\label{app:access:maintenance}

The authors are committed to maintaining PoreML as
the solver, dataset, and framework are used in future research. Maintenance
includes tracking and resolving issues raised by the community after
release; feedback is monitored through the GitHub issue trackers of the two
code repositories and the community pages of the three Hugging Face
repositories. All Hugging Face repositories are version-controlled, so
revisions to the data remain traceable and earlier states remain retrievable.

The release follows the FAIR principles.
\textbf{Findable:} the dataset and the geometry collection each reside at a
single public Hugging Face address and carry descriptive metadata and
searchable tags; all present and future versions will share these addresses.
\textbf{Accessible:} all data and metadata can be downloaded without
registration, through the Hugging Face command-line and Python clients or
through the download utility of the framework, which selects runs by
scenario, geometry family, domain size, or split.
\textbf{Interoperable:} trajectories are stored as standard HDF5 files with
Zstandard compression, readable with common libraries such as \texttt{h5py},
and each is paired with an XDMF index for direct visualization in ParaView.
Geometries are stored as NumPy arrays. Every run carries its solver
configuration, geometry provenance, and run conditions as JSON, and its
scalar diagnostics as CSV.
\textbf{Reusable:} the solver, the framework, and the dataset are released
under the MIT License, and the geometry collection under the Creative
Commons Attribution 4.0 International License. Every run records the solver
settings, the software environment, and a hash of its input geometry, so a
trajectory can be traced to the exact inputs that produced it.

\section{Model Comparison Visualization}
\label{app:vis}

The tables above summarize accuracy over whole campaigns. This appendix
shows what those averages look like inside a single trajectory. For each
campaign and each evaluation setting we take one test run and render the
same 64-step autoregressive rollout for the reference simulation and for
all five baselines. The two figures of a pair share one run: the first
shows the three-dimensional phase field, the second a mid-plane slice
through the same volume, both at rollout steps
$h\in\{7,18,30,41,53,64\}$. In the three campaigns shown, frames are stored
every 5{,}000 LBM steps,
and the counter above each column is the absolute step of the underlying
simulation. Every predicted panel carries its phase overlap (IoU) against
the reference at that step, so that visual drift and metric can be read
together.

\subsection{In-Distribution Test Results}
\label{app:vis:indist}

Geometry family, domain size and parameter ranges all match training
here, so the only extrapolation is in time. In each of the three runs
below the displacement advances a long way inside the window --- a
ganglion inflates, a third of the trapped phase is swept out of the pore
space, the water content of a fiber mat doubles --- and how far it
advances is what separates the baselines: the step-64 overlaps fall from
$0.77$--$0.93$ in the drainage run to $0.51$--$0.67$ in the trapping run
and $0.45$--$0.71$ in the GDL run.

\begin{figure}[!htbp]
  \centering
  \includegraphics[width=0.92\linewidth]{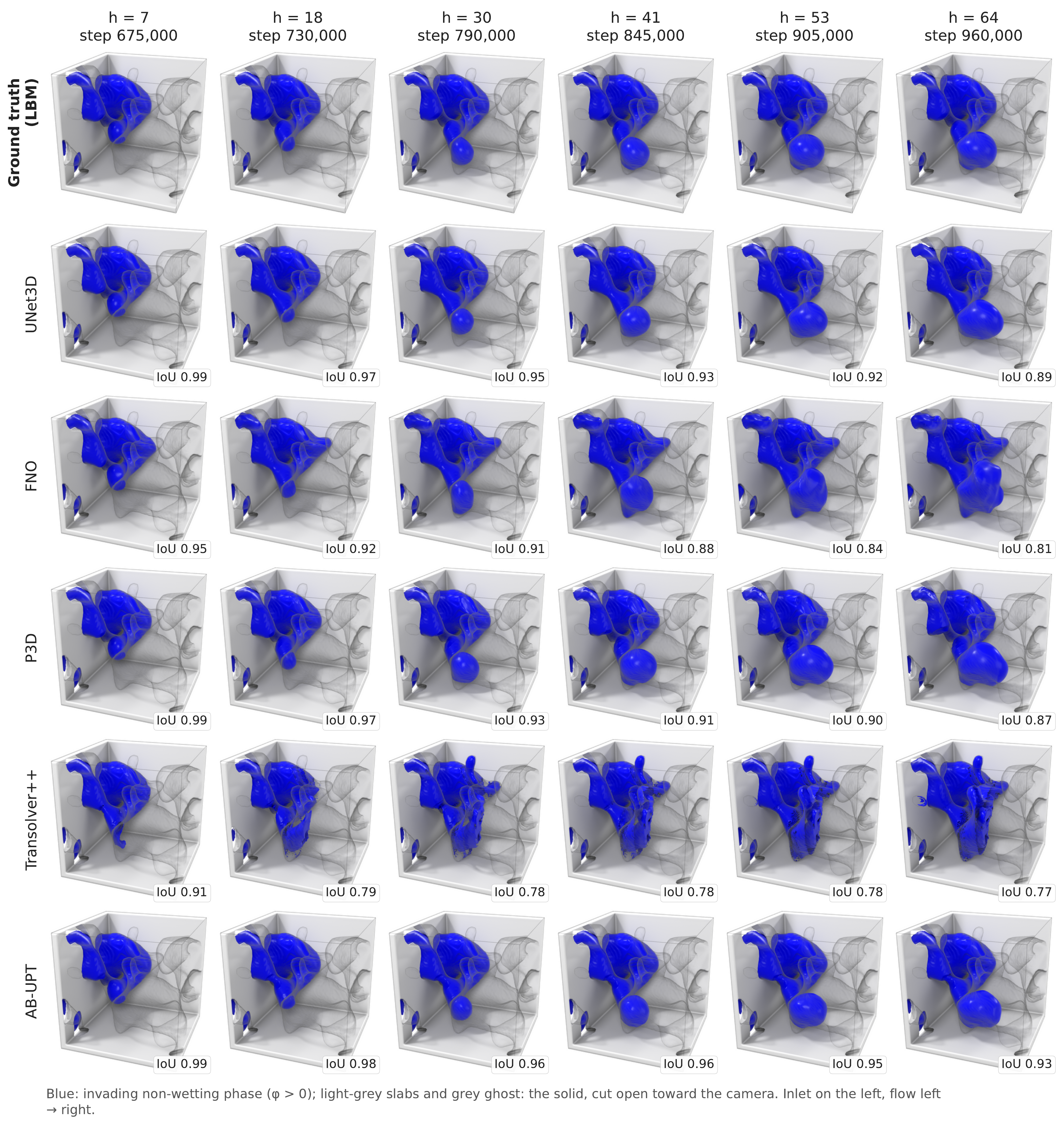}
  \caption{\textbf{Drainage, in-distribution: three-dimensional rollout.}
  A generated blob medium at $128^3$ voxels with porosity $0.24$, $M=0.2$
  and $\theta=150^{\circ}$; the rollout starts at stored frame 128. Rows are the reference simulation and the five
  baselines, columns the six rollout steps, and each predicted panel is
  labeled with its IoU. Over the window the reference inflates a single
  ganglion in the center of the domain, from a bud at step 7 to a rounded
  body at step 64, and the non-wetting saturation rises by half, from
  $0.26$ to $0.39$. AB-UPT reproduces that growth at IoU $0.93$ and ends
  on the reference saturation; UNet3D ($0.89$) and P3D ($0.87$) follow it
  slightly full; FNO3D ($0.81$) over-invades to $0.49$, and Transolver++
  ($0.77$) fragments the interface as the rollout proceeds.}
  \label{fig:vis-indist-drainage-3d}
\end{figure}

\begin{figure}[!htbp]
  \centering
  \includegraphics[width=0.92\linewidth]{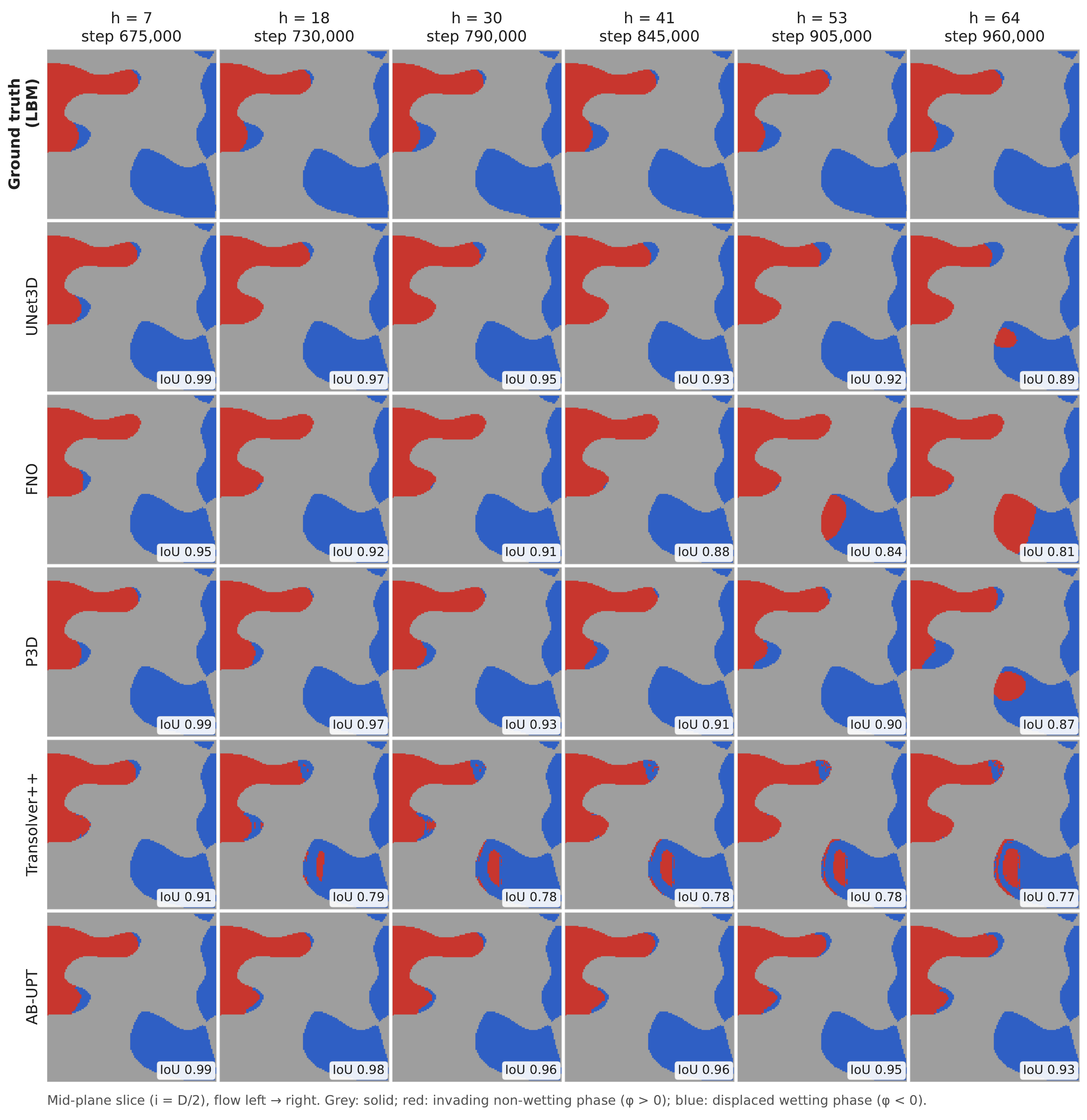}
  \caption{\textbf{Drainage, in-distribution: mid-plane slice.}
  The same run and steps as Figure~\ref{fig:vis-indist-drainage-3d},
  sliced through the middle of the domain. The inflating ganglion lies
  off this plane, so the reference slice holds its shape for the whole
  window --- which makes the figure a test of what a rollout invents
  where nothing happens. Every model eventually opens a spurious
  non-wetting patch inside the wetting-filled pore below the main body:
  Transolver++ carries one from step 18 to the end, FNO3D grows a large
  one from step 53, and UNet3D and P3D each open one at step 64.}
  \label{fig:vis-indist-drainage-2d}
\end{figure}
\FloatBarrier

\begin{figure}[!htbp]
  \centering
  \includegraphics[width=0.92\linewidth]{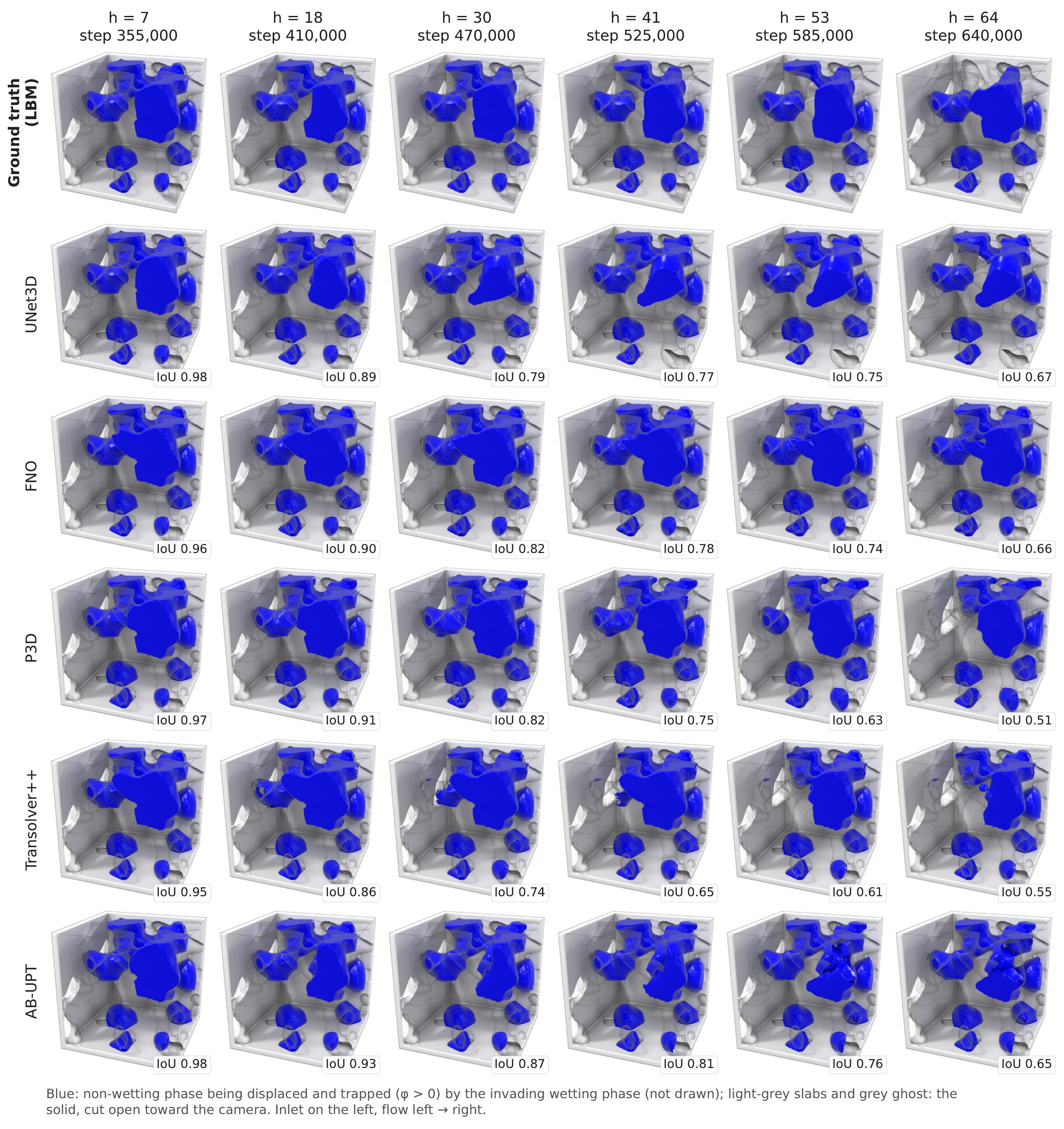}
  \caption{\textbf{Capillary trapping, in-distribution: three-dimensional
  rollout.} A generated blob medium at $128^3$ voxels with porosity $0.23$,
  $M=0.2$ and $\theta=120^{\circ}$; the rollout starts at stored frame 64.
  The wetting flood removes nearly a third of
  the trapped phase, from $0.52$ to $0.35$ saturation, and breaks the
  connected body in the middle of the domain into separate ganglia. Every
  baseline under-mobilizes: the endpoint saturations run from $0.38$
  (UNet3D) to $0.44$ (FNO3D) against the reference $0.35$, and the
  step-64 overlaps spread from $0.67$ (UNet3D), $0.66$ (FNO3D) and
  $0.65$ (AB-UPT) down to $0.55$ (Transolver++) and $0.51$ (P3D), which
  keeps the central body nearly intact.}
  \label{fig:vis-indist-trapping-3d}
\end{figure}

\begin{figure}[!htbp]
  \centering
  \includegraphics[width=0.92\linewidth]{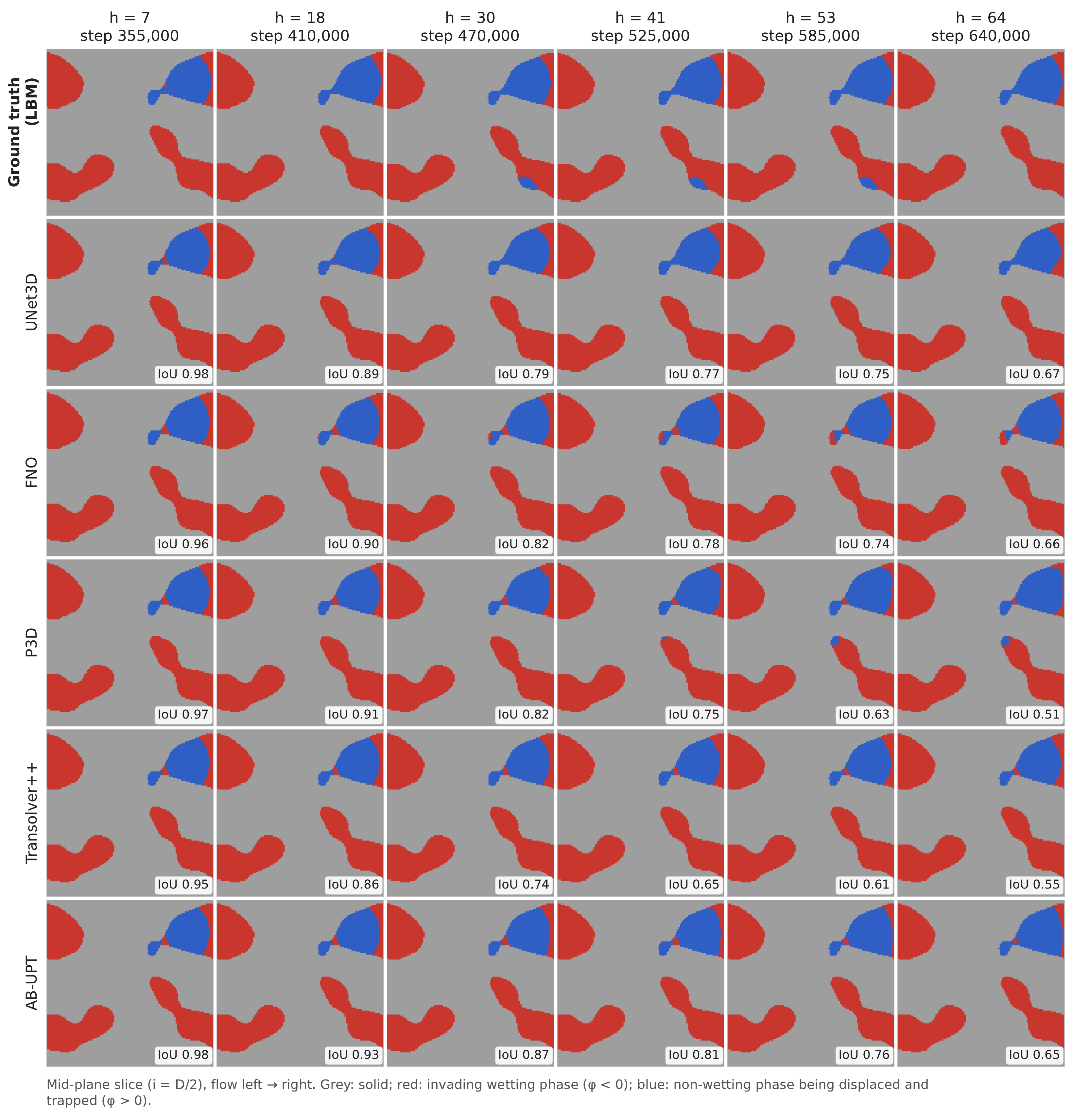}
  \caption{\textbf{Capillary trapping, in-distribution: mid-plane slice.}
  The same run and steps as Figure~\ref{fig:vis-indist-trapping-3d};
  here red is the invading wetting phase and blue the non-wetting phase
  being displaced. The reference thins the trapped blob at the top of the
  slice from the side facing the invading phase and narrows the neck that
  connects it to the throat on its left. All five models keep the blob
  fuller than the reference; P3D keeps it widest and its overlap decays
  furthest, from $0.97$ at step 7 to $0.51$ at step 64.}
  \label{fig:vis-indist-trapping-2d}
\end{figure}
\FloatBarrier

\begin{figure}[!htbp]
  \centering
  \includegraphics[width=\linewidth]{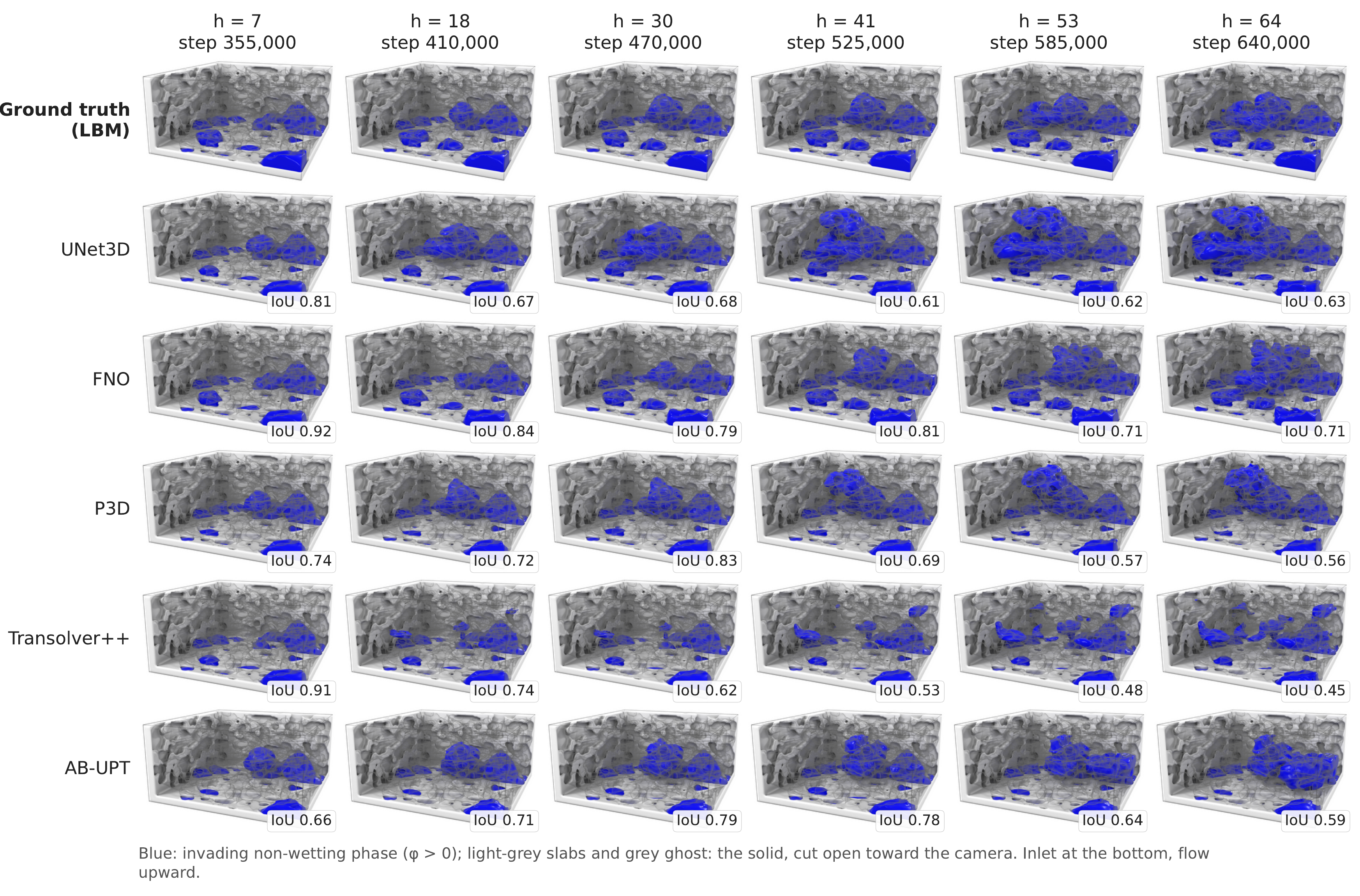}
  \caption{\textbf{GDL intrusion, in-distribution: three-dimensional
  rollout.} A generated fiber mat at $128\times128\times64$ voxels with
  porosity $0.55$, $M=5$ and $\theta=140^{\circ}$; the rollout starts at
  stored frame 64. Water climbs
  through the mat from the inlet at the bottom and the saturation
  doubles, from $0.11$ to $0.21$ --- the largest relative change of any
  run in this appendix. The baselines bracket that growth from both
  sides: UNet3D ends half again too wet at $0.31$, FNO3D at $0.25$,
  AB-UPT closest at $0.22$, while P3D stops at $0.19$ and Transolver++ at
  $0.12$, barely above the saturation it started from. The step-64
  overlaps follow the same order: FNO3D $0.71$, UNet3D $0.63$, AB-UPT
  $0.59$, P3D $0.56$, Transolver++ $0.45$.}
  \label{fig:vis-indist-gdl-3d}
\end{figure}

\begin{figure}[!htbp]
  \centering
  \includegraphics[width=\linewidth]{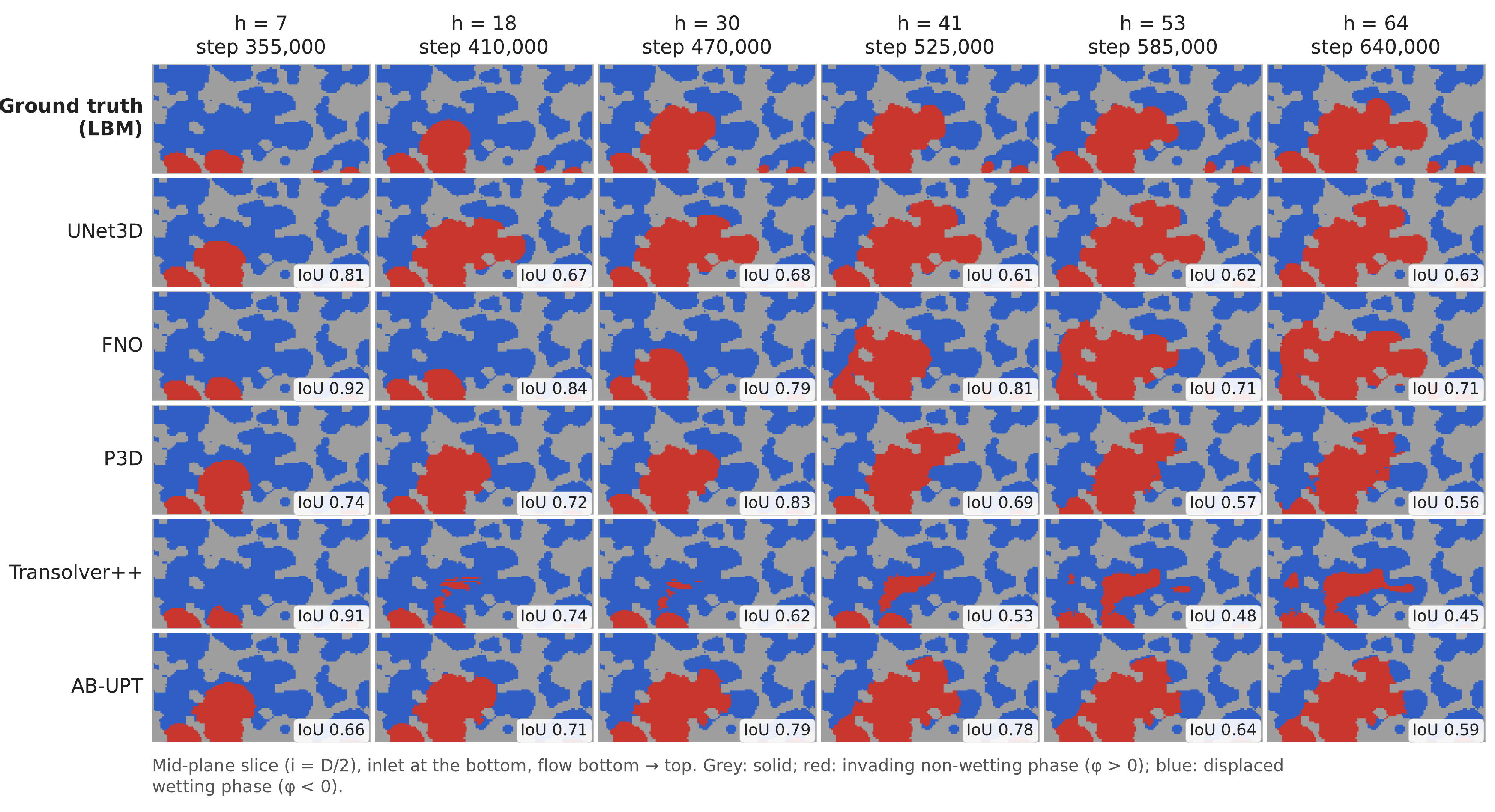}
  \caption{\textbf{GDL intrusion, in-distribution: mid-plane slice.}
  The same run and steps as Figure~\ref{fig:vis-indist-gdl-3d}. In the
  slice the invading phase goes from two small patches at the bottom to a
  single body spanning the middle of the mat. The errors are errors of
  timing: UNet3D, P3D and AB-UPT merge the two patches into one cluster
  already at step 7, ahead of the reference, while Transolver++ starts
  correctly and then stops growing, leaving a thin ragged filament that
  never fills the pore space --- the slice-level counterpart of its
  $0.12$ endpoint saturation.}
  \label{fig:vis-indist-gdl-2d}
\end{figure}
\FloatBarrier

\subsection{Geo-Shift}
\label{app:vis:geoshift}

These runs apply weights trained on generated geometries to $\mu$CT rock
and carbon-paper scans at the training domain size. The displacement
remains qualitatively right, but the spread between models widens, and in
the trapping and GDL runs below the five baselines bracket the reference
saturation from both sides rather than erring together in one direction.

\begin{figure}[!htbp]
  \centering
  \includegraphics[width=0.92\linewidth]{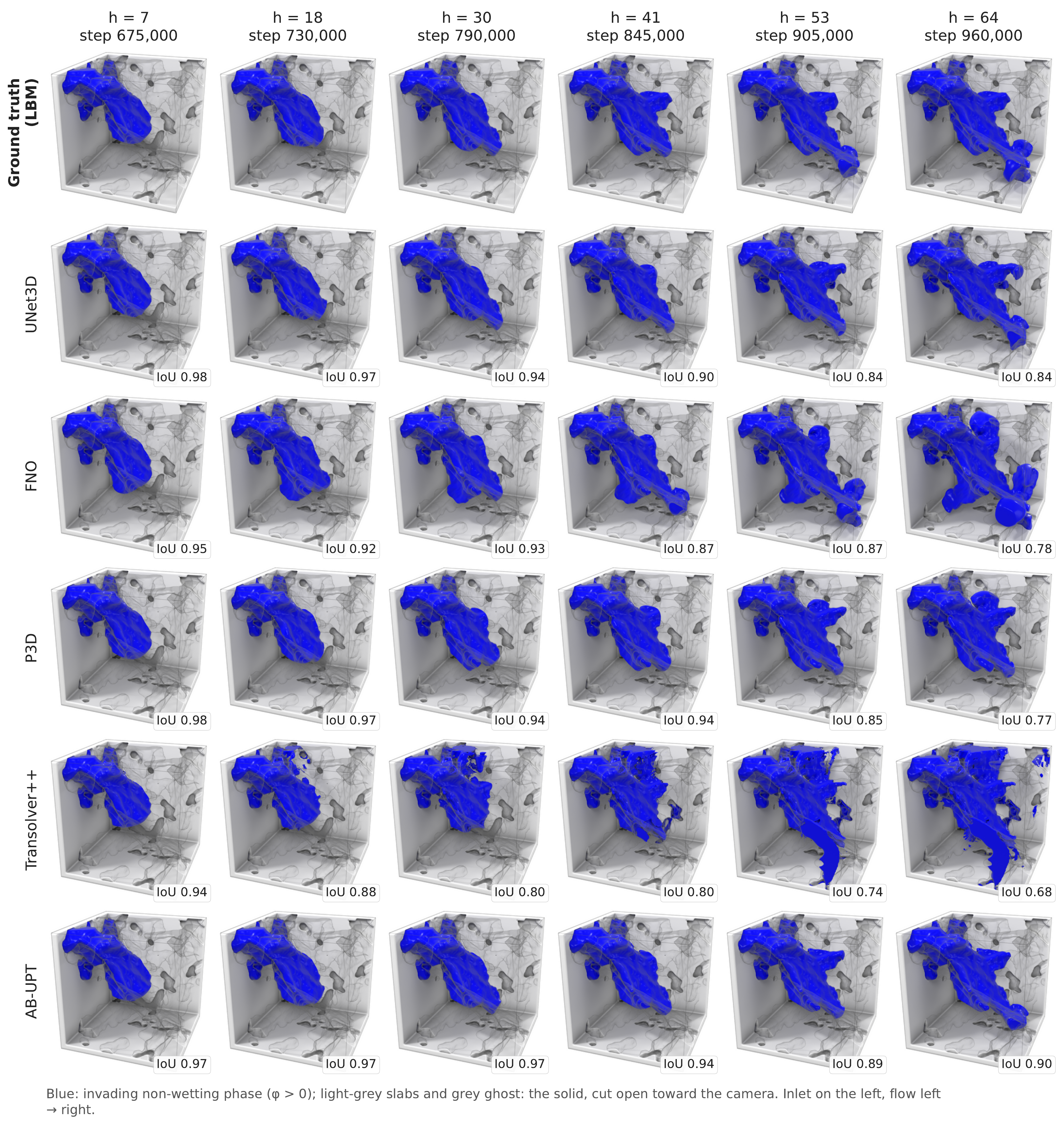}
  \caption{\textbf{Drainage, Geo-Shift: three-dimensional rollout.}
  A CastleGate sandstone sample at $128^3$ voxels with porosity $0.20$,
  $M=0.2$ and $\theta=120^{\circ}$; the rollout starts at stored frame 128.
  The reference gains almost half its
  non-wetting volume, from $0.30$ to $0.44$ saturation, extending the
  cluster along the pore and opening a new branch late in the window. How
  much of that branch a model produces separates the rollouts: AB-UPT
  reaches IoU $0.90$ and is the only baseline to end below the reference
  saturation, at $0.41$; UNet3D $0.84$ and P3D $0.77$ end just above it;
  Transolver++ $0.68$ roughens the interface, and FNO3D $0.78$
  over-invades to $0.54$.}
  \label{fig:vis-geoshift-drainage-3d}
\end{figure}

\begin{figure}[!htbp]
  \centering
  \includegraphics[width=0.92\linewidth]{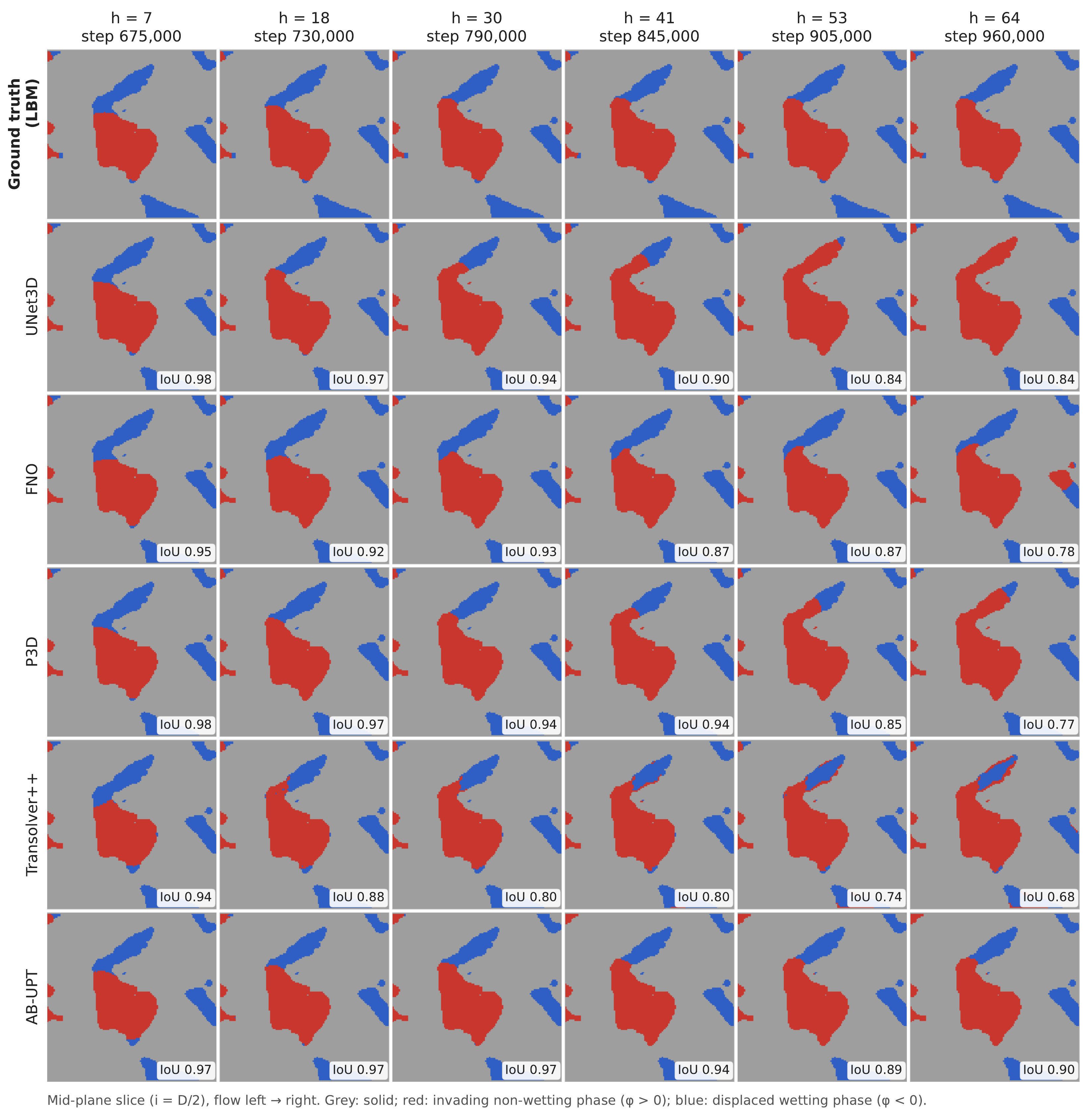}
  \caption{\textbf{Drainage, Geo-Shift: mid-plane slice.}
  The same run and steps as Figure~\ref{fig:vis-geoshift-drainage-3d}.
  The slice isolates the new branch: the arm at the top of the slice is
  wetting-filled at step 7 and half non-wetting by step 64. UNet3D, P3D
  and AB-UPT advance that front at about the right rate, Transolver++
  carries it higher up the arm than the reference, and FNO3D additionally
  opens a pore on the right that the reference leaves filled with the
  wetting phase --- which is where its saturation excess accumulates.}
  \label{fig:vis-geoshift-drainage-2d}
\end{figure}
\FloatBarrier

\begin{figure}[!htbp]
  \centering
  \includegraphics[width=0.92\linewidth]{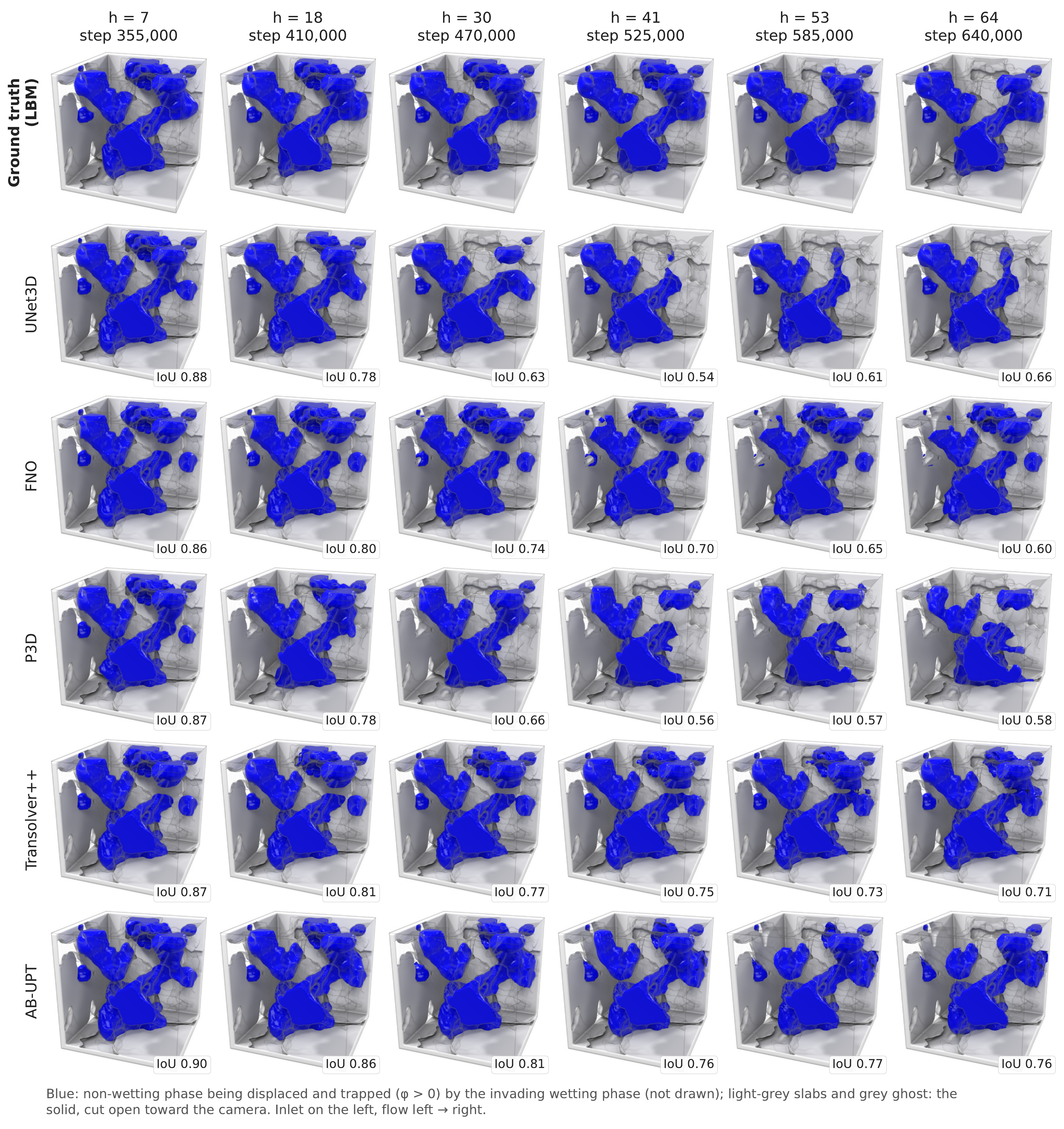}
  \caption{\textbf{Capillary trapping, Geo-Shift: three-dimensional
  rollout.} A Bentheimer sandstone sample at $128^3$ voxels with porosity
  $0.21$, $M=0.05$ and $\theta=130^{\circ}$; the rollout starts at stored
  frame 64. The wetting flood strips about a fifth of
  the trapped non-wetting phase, from $0.55$ to $0.44$ saturation, and
  the models bracket that endpoint: P3D and UNet3D remove too much and
  end at $0.38$, FNO3D and Transolver++ too little at $0.48$ and $0.52$,
  and AB-UPT lands closest at $0.45$ with the highest overlap, $0.76$.}
  \label{fig:vis-geoshift-trapping-3d}
\end{figure}

\begin{figure}[!htbp]
  \centering
  \includegraphics[width=0.92\linewidth]{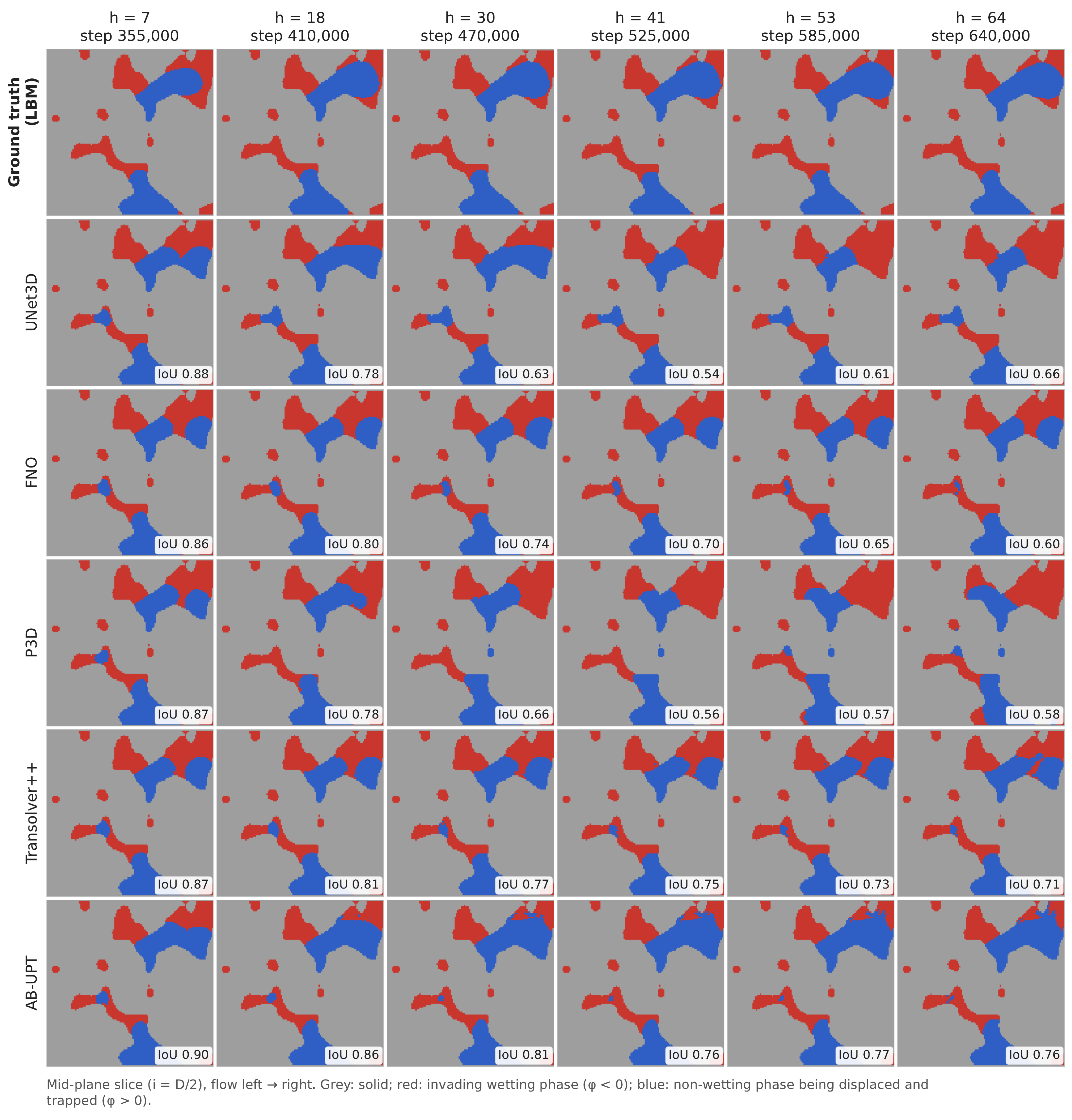}
  \caption{\textbf{Capillary trapping, Geo-Shift: mid-plane slice.}
  The same run and steps as Figure~\ref{fig:vis-geoshift-trapping-3d};
  red is the invading wetting phase and blue the non-wetting phase being
  displaced. The wetting phase eats into the upper body from the right,
  and where a model stops that erosion sets its endpoint. The overlap
  need not decay monotonically: UNet3D dips to $0.54$ at step 41, when it
  has cleared too much of the body, and recovers to $0.66$ by step 64 as
  the reference catches up with it.}
  \label{fig:vis-geoshift-trapping-2d}
\end{figure}
\FloatBarrier

\begin{figure}[!htbp]
  \centering
  \includegraphics[width=\linewidth]{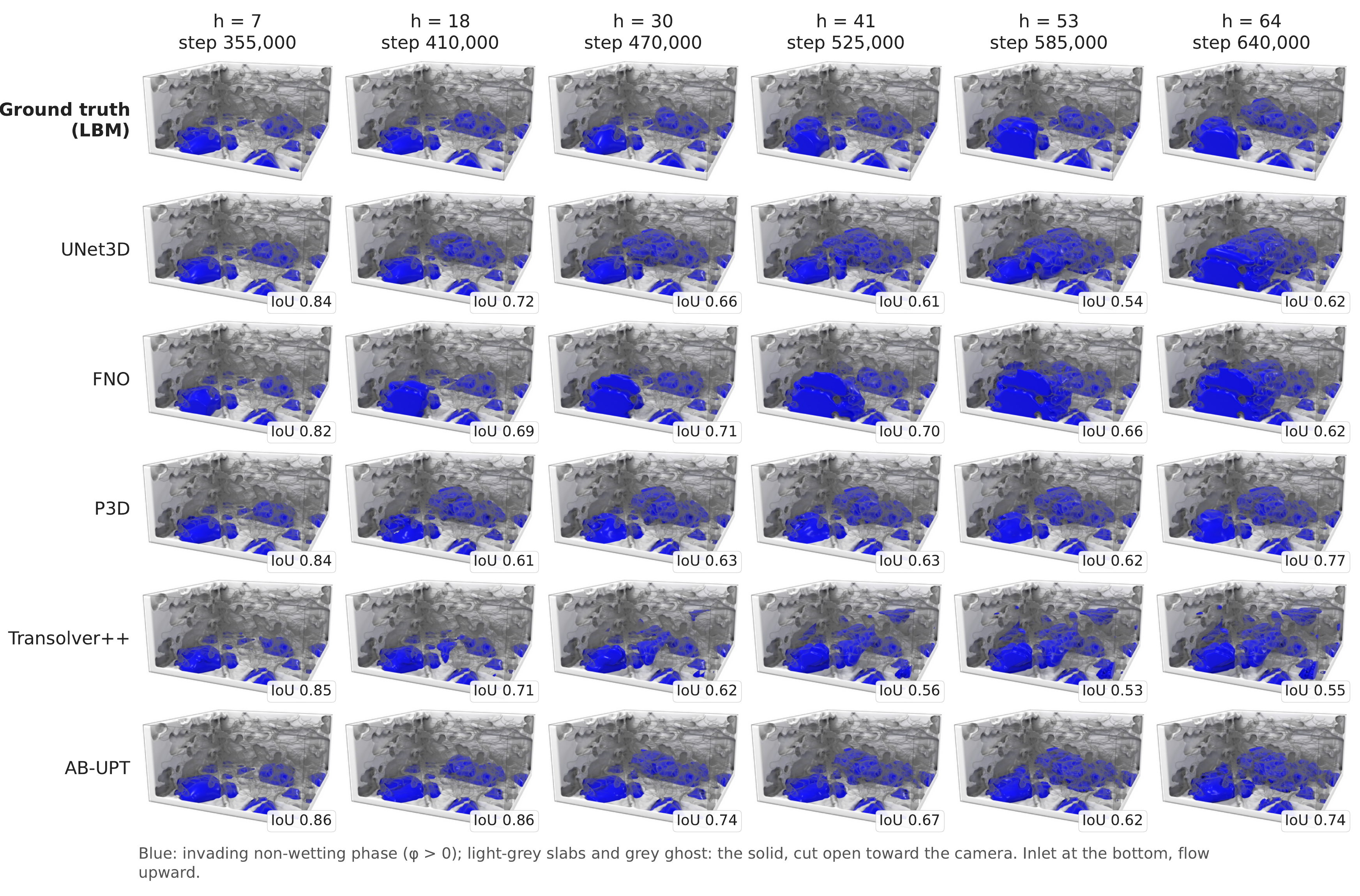}
  \caption{\textbf{GDL intrusion, Geo-Shift: three-dimensional rollout.}
  A $\mu$CT carbon-paper GDL with 40\% PTFE coating at
  $128\times128\times64$ voxels, porosity $0.60$, $M=5$ and
  $\theta=130^{\circ}$; the rollout starts at stored frame 64. Water
  invasion doubles over the window, from $0.10$ to $0.19$ saturation, and
  the baselines bracket that endpoint widely: UNet3D and FNO3D end half
  again too wet, at $0.29$ and $0.27$, P3D lands on the reference volume
  at $0.19$, and Transolver++ stops short at $0.16$. The step-64 overlaps
  are P3D $0.77$, AB-UPT $0.74$, UNet3D and FNO3D $0.62$, Transolver++
  $0.55$.}
  \label{fig:vis-geoshift-gdl-3d}
\end{figure}

\begin{figure}[!htbp]
  \centering
  \includegraphics[width=\linewidth]{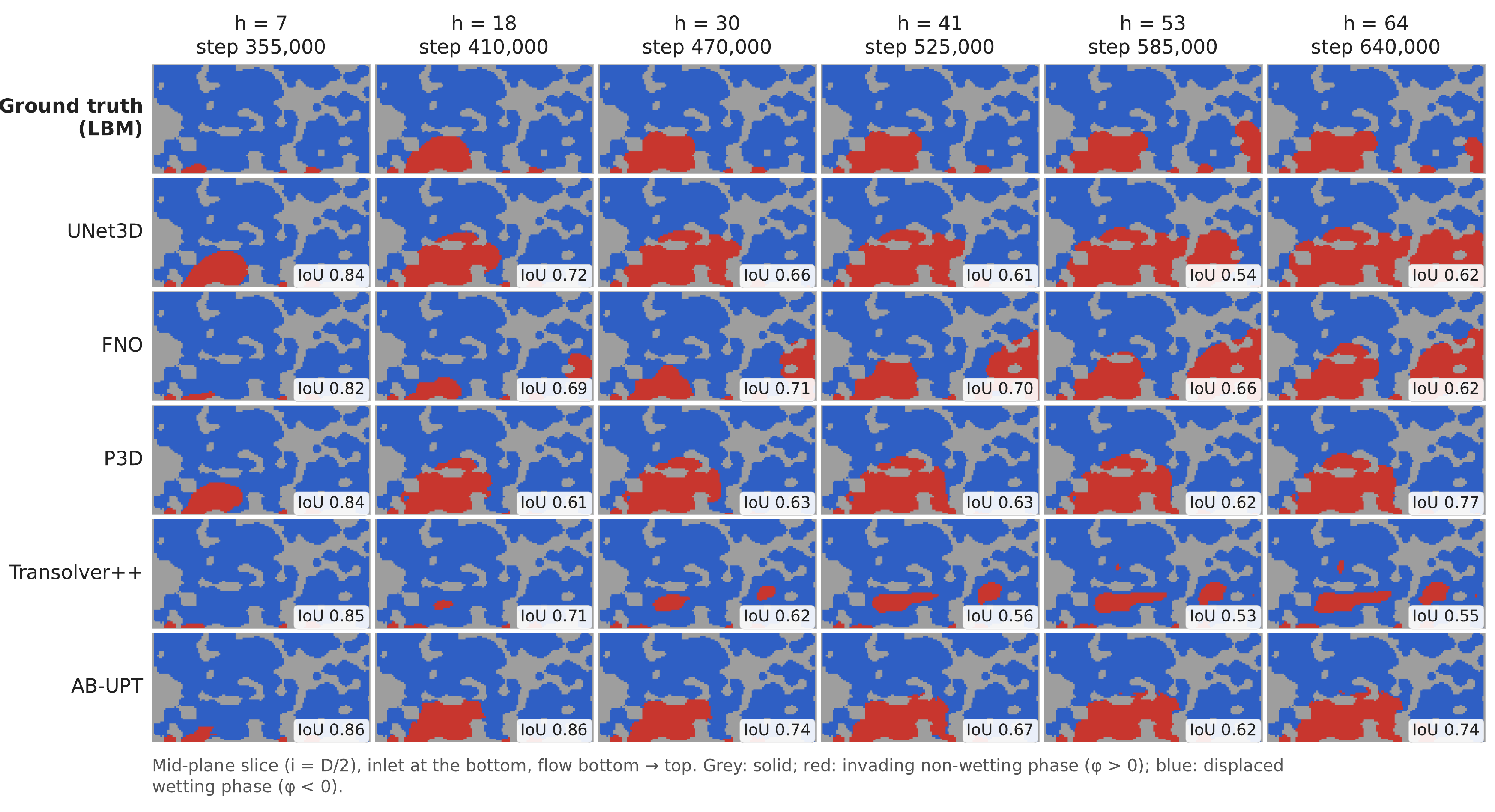}
  \caption{\textbf{GDL intrusion, Geo-Shift: mid-plane slice.}
  The same run and steps as Figure~\ref{fig:vis-geoshift-gdl-3d}. In the
  scanned layer the water advances as a broad body between the fibers
  rather than as a front, and the models disagree about where it starts:
  UNet3D and P3D nucleate a large cluster at the lower left already at
  step 7, FNO3D builds one at the right edge that the reference reaches
  only at step 53, and Transolver++ leaves isolated specks that never
  merge, its overlap decaying to $0.55$ while its saturation stalls at
  $0.16$.}
  \label{fig:vis-geoshift-gdl-2d}
\end{figure}
\FloatBarrier

\subsection{Scale-Up}
\label{app:vis:scaleup}

These runs apply the small-domain weights to the paired large domain at
fixed resolution, so local discretization is unchanged and only the
extent of the pore network grows.

\begin{figure}[!htbp]
  \centering
  \includegraphics[width=0.92\linewidth]{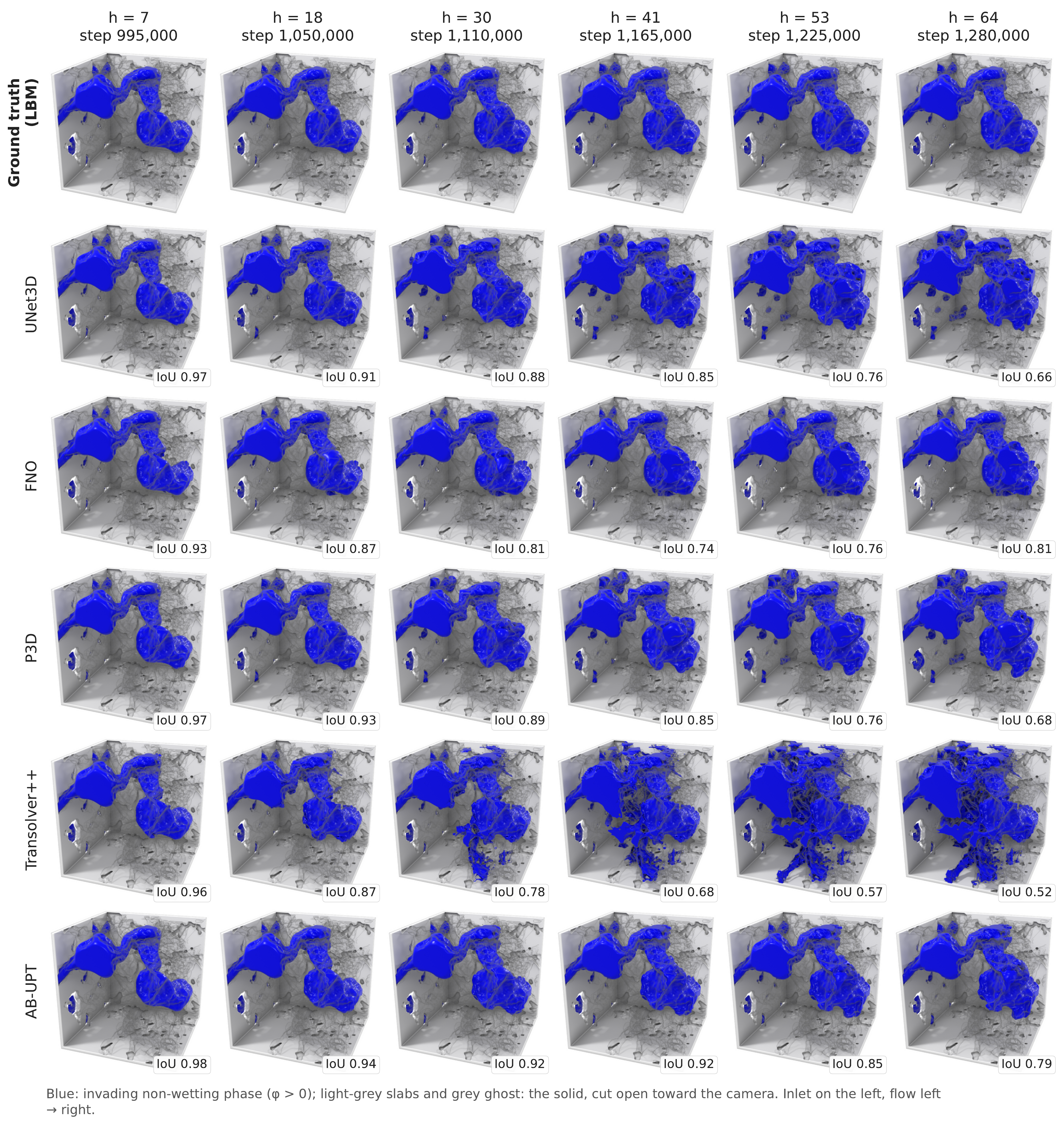}
  \caption{\textbf{Drainage, Scale-Up: three-dimensional rollout.}
  A Buff Berea sandstone sample at $256^3$ voxels with porosity $0.20$,
  $M=0.2$ and $\theta=150^{\circ}$; the rollout starts at stored frame 192.
  In eight times the training volume the
  reference extends its cluster through the pore network and moves from
  $0.22$ to $0.29$ saturation. FNO3D holds the pattern best here at IoU
  $0.81$, just ahead of AB-UPT $0.79$, which also ends closest in volume;
  P3D $0.68$ and UNet3D $0.66$ keep the shape but end too wet, at $0.34$
  and $0.36$; Transolver++ $0.52$ shreds the body into disconnected
  fragments and ends at $0.44$ saturation, half again the reference.}
  \label{fig:vis-scaleup-drainage-3d}
\end{figure}

\begin{figure}[!htbp]
  \centering
  \includegraphics[width=0.92\linewidth]{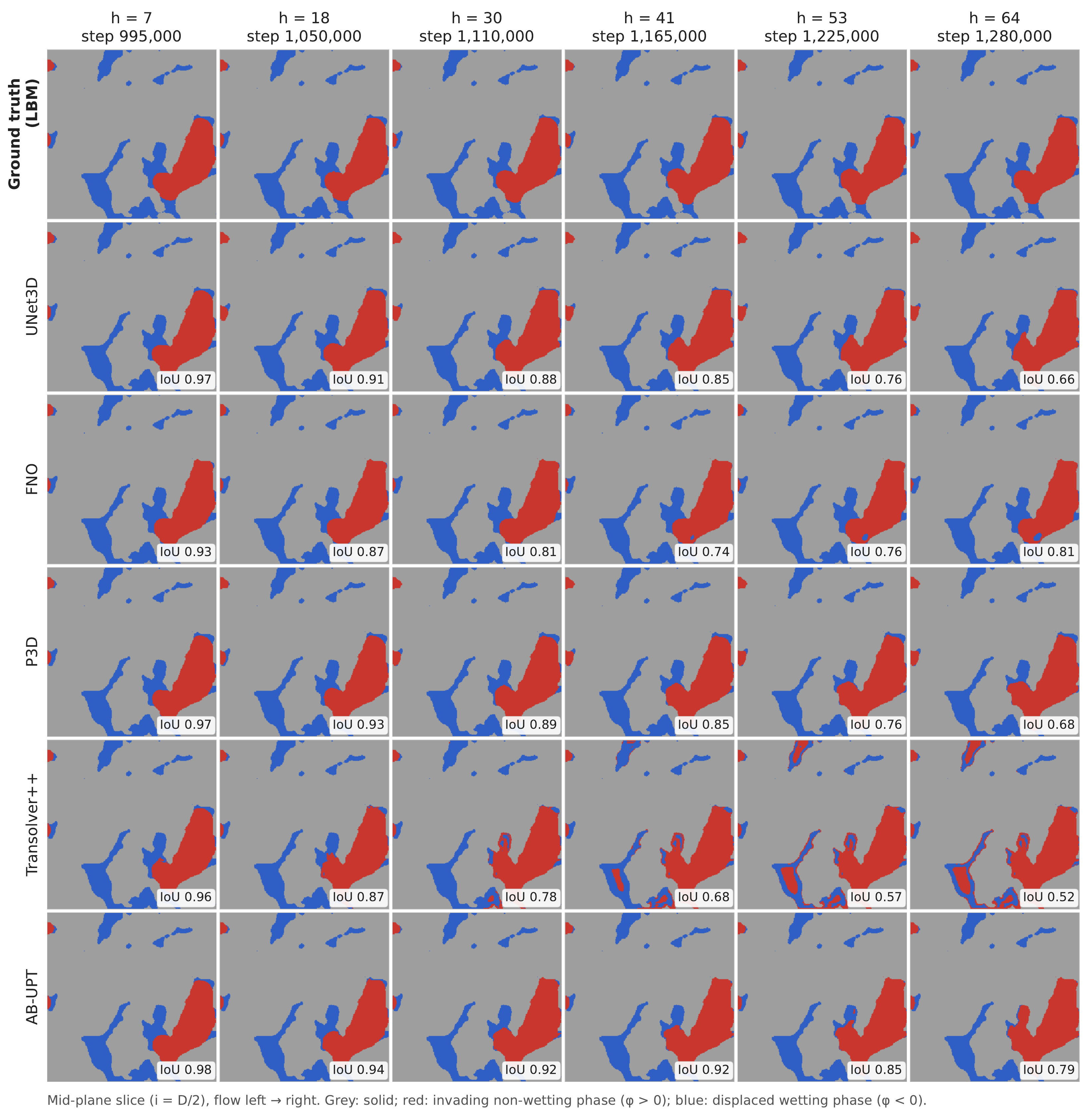}
  \caption{\textbf{Drainage, Scale-Up: mid-plane slice.}
  The same run and steps as Figure~\ref{fig:vis-scaleup-drainage-3d}. The
  slice shows where the excess volume accumulates: UNet3D and P3D thicken
  the cluster where it meets the wetting phase at the top, and
  Transolver++ breaks into the wetting-filled channel at the lower left
  from step 30 onward and holds that intrusion to the end of the window,
  while the reference keeps the channel filled.}
  \label{fig:vis-scaleup-drainage-2d}
\end{figure}
\FloatBarrier

\begin{figure}[!htbp]
  \centering
  \includegraphics[width=0.92\linewidth]{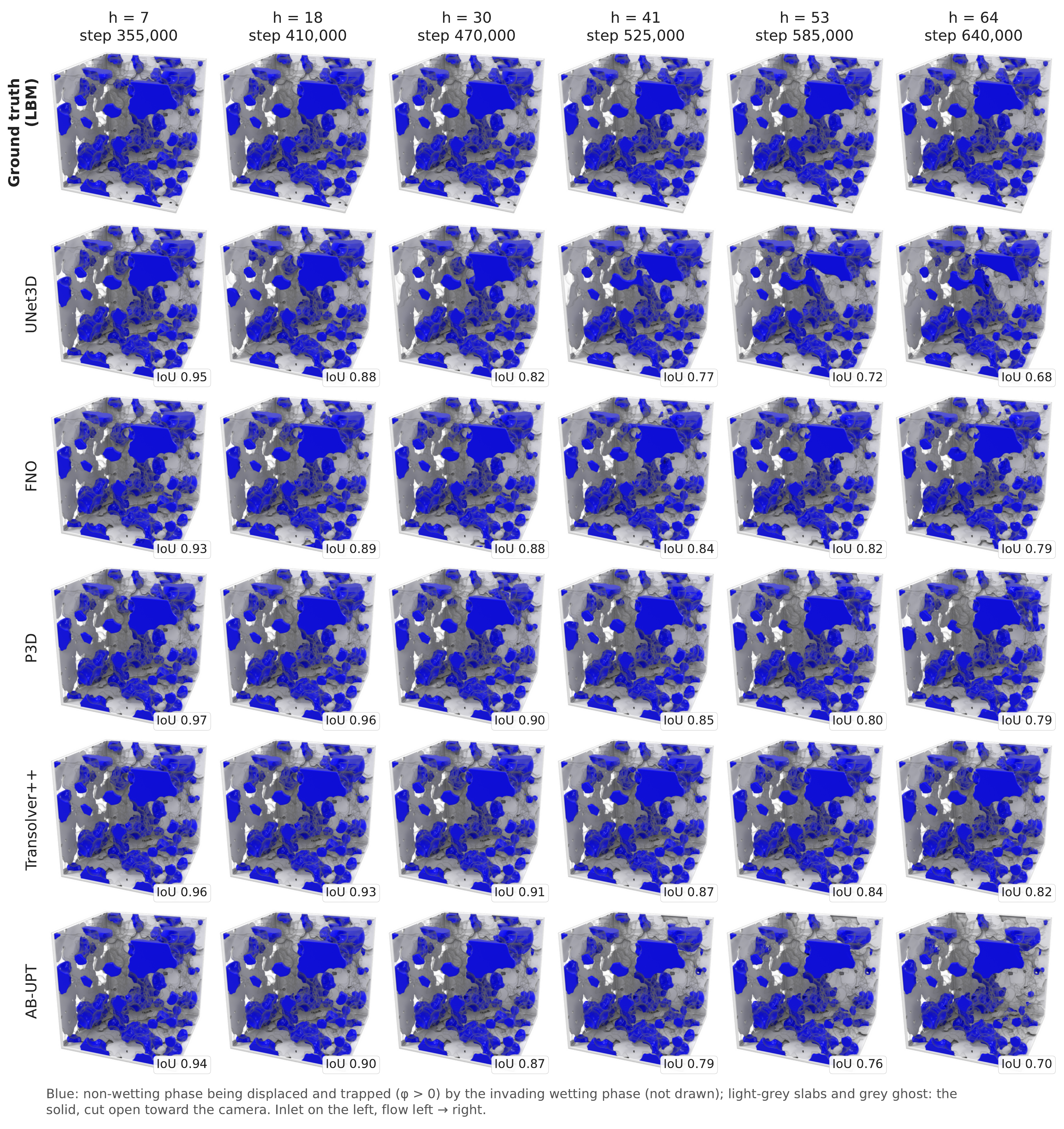}
  \caption{\textbf{Capillary trapping, Scale-Up: three-dimensional
  rollout.} A generated sphere pack at $256^3$ voxels with porosity $0.18$,
  $M=0.1$ and $\theta=120^{\circ}$; the rollout starts at stored frame 64.
  The flood removes a sixth of the trapped
  phase, from $0.45$ to $0.37$ saturation, while the many isolated
  ganglia stay where they are: at this size the window covers a slow,
  distributed erosion rather than a single event. Overlap and volume
  disagree here. Transolver++ holds the highest overlap, $0.82$, while
  ending furthest above the reference saturation at $0.40$, and UNet3D
  ends within $0.01$ of the reference volume at the lowest overlap,
  $0.68$; AB-UPT alone strips too much and ends at $0.30$.}
  \label{fig:vis-scaleup-trapping-3d}
\end{figure}

\begin{figure}[!htbp]
  \centering
  \includegraphics[width=0.92\linewidth]{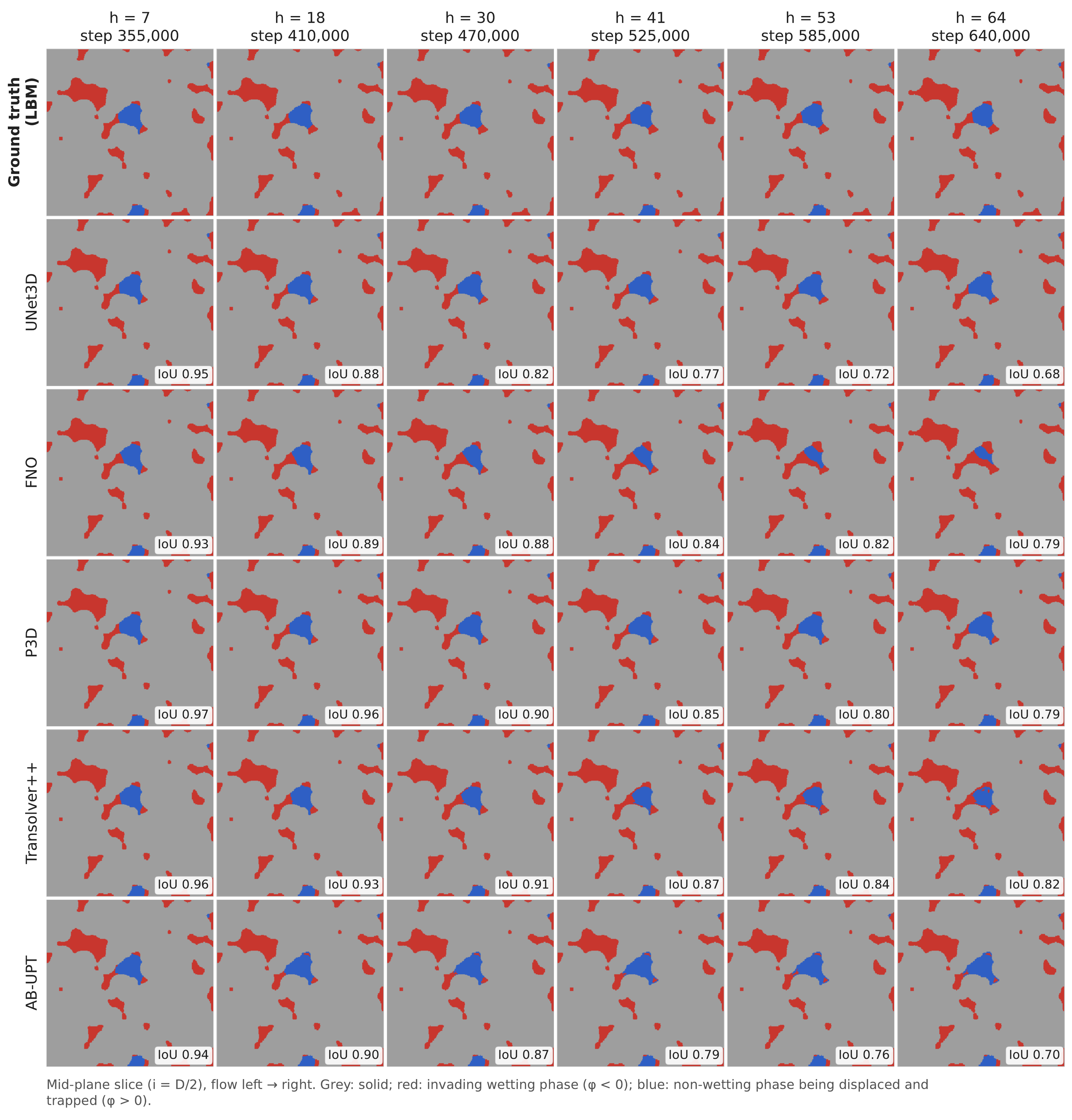}
  \caption{\textbf{Capillary trapping, Scale-Up: mid-plane slice.}
  The same run and steps as Figure~\ref{fig:vis-scaleup-trapping-3d};
  red is the invading wetting phase and blue the non-wetting phase being
  displaced. The slice makes the slowness explicit: the ganglion at the
  center keeps its outline for the whole window and loses volume only
  along its rim, and the models differ by a few voxels around that rim
  --- AB-UPT erodes it furthest, Transolver++ least.}
  \label{fig:vis-scaleup-trapping-2d}
\end{figure}
\FloatBarrier

\begin{figure}[!htbp]
  \centering
  \includegraphics[width=\linewidth]{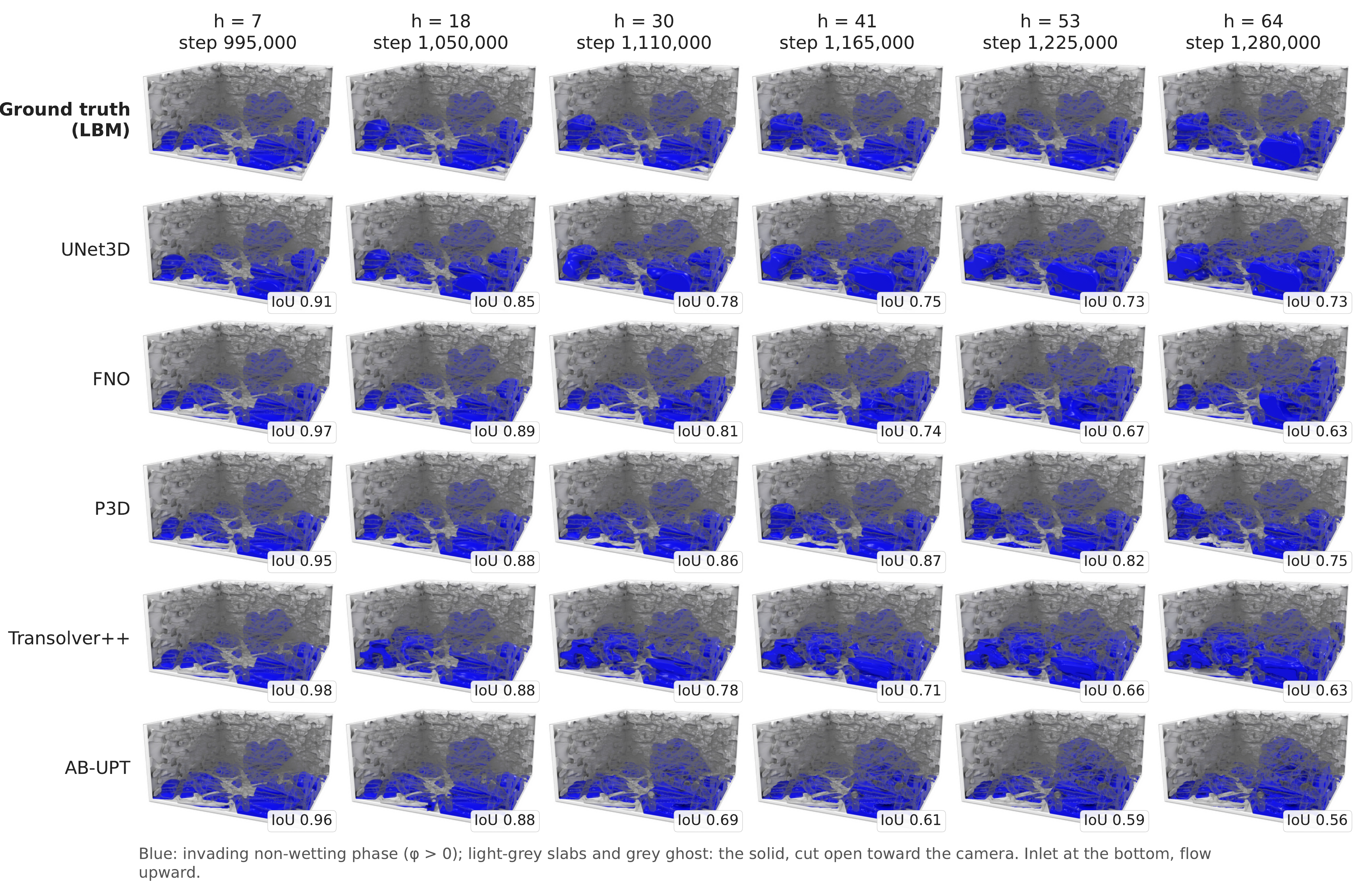}
  \caption{\textbf{GDL intrusion, Scale-Up: three-dimensional rollout.}
  A generated fiber mat at $256\times256\times128$ voxels with porosity
  $0.65$, $M=1$ and $\theta=120^{\circ}$; the rollout starts at stored
  frame 192. The reference
  invades a third more of the layer, from $0.14$ to $0.19$ saturation,
  across four times the in-plane area of the training domain. P3D
  ($0.75$) and UNet3D ($0.73$) hold the intrusion pattern best; FNO3D and
  Transolver++ (both $0.63$) carry too much water and end at $0.23$
  against the reference $0.19$; AB-UPT ends on the right volume, $0.19$,
  at the lowest overlap of the five, $0.56$.}
  \label{fig:vis-scaleup-gdl-3d}
\end{figure}

\begin{figure}[!htbp]
  \centering
  \includegraphics[width=\linewidth]{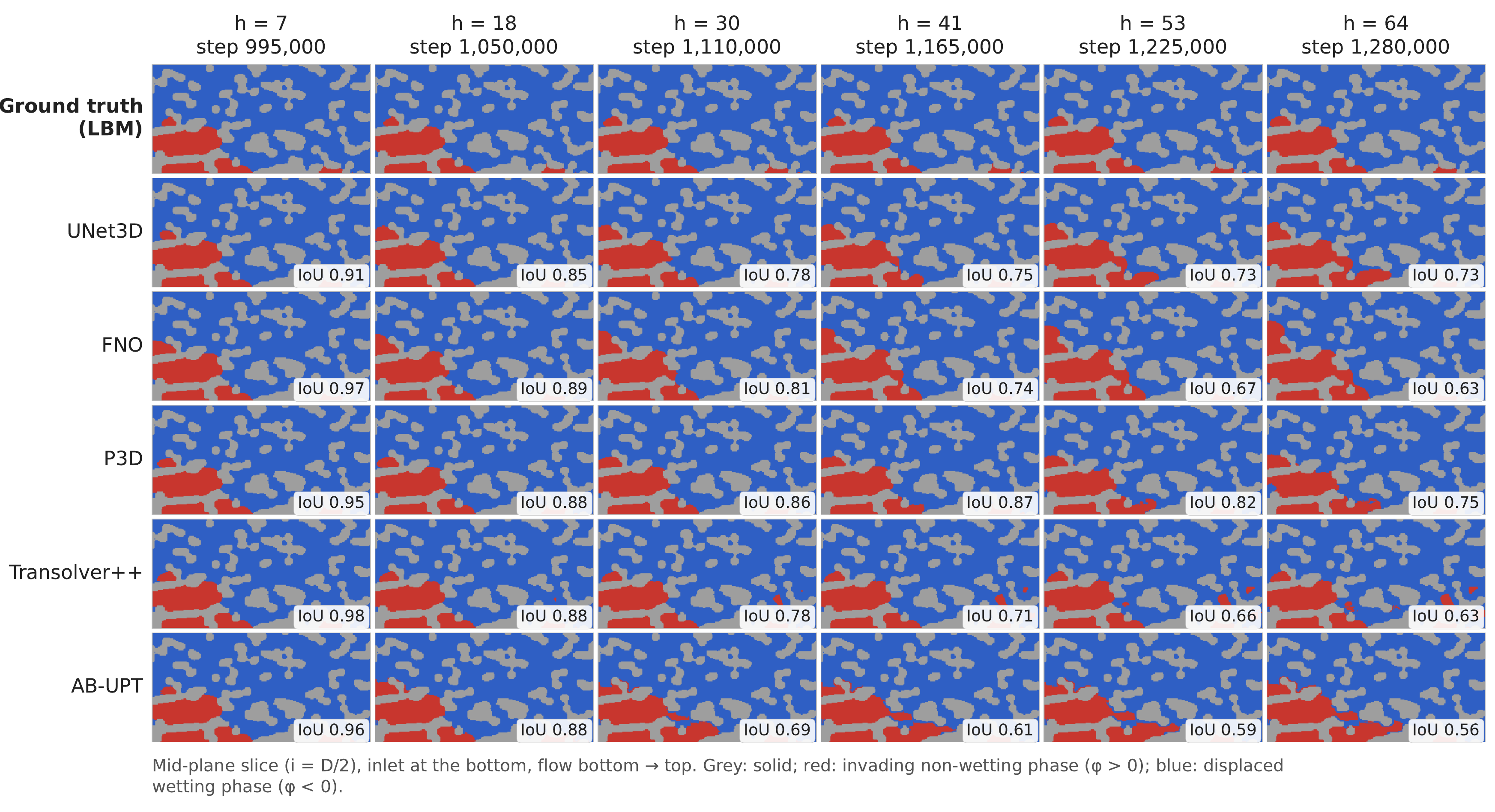}
  \caption{\textbf{GDL intrusion, Scale-Up: mid-plane slice.}
  The same run and steps as Figure~\ref{fig:vis-scaleup-gdl-3d}. In the
  slice the water holds a flat band across the bottom of the layer and
  thickens it instead of fingering upward, and the models differ along
  the right edge of that band: AB-UPT extends it across the foot of the
  slice from step 30 onward, UNet3D and FNO3D follow later, and P3D keeps
  the band closest to the reference width.}
  \label{fig:vis-scaleup-gdl-2d}
\end{figure}
\FloatBarrier

\end{document}